\documentclass[sigconf]{acmart}
\AtBeginDocument{%
  }

\copyrightyear{2025}
\acmYear{2025}
\setcopyright{acmlicensed}
\acmConference[MM '25] {Proceedings of the 33rd ACM International Conference on Multimedia}{October 27--31, 2025}{Dublin, Ireland.}
\acmBooktitle{Proceedings of the 33rd ACM International Conference on Multimedia (MM '25), October 27--31, 2025, Dublin, Ireland}
\acmISBN{979-8-4007-2035-2/2025/10}
\acmDOI{10.1145/3746027.3755184}

\begin{document}

\title[Cross-Domain Attribute Alignment with CLIP]{Cross-Domain Attribute Alignment with CLIP: A Rehearsal-Free Approach for Class-Incremental Unsupervised Domain Adaptation}

\author{Kerun Mi}
\affiliation{
  \department{School of Computer Science and Engineering}
  \institution{Nanjing University of Science and Technology} 
  \city{Nanjing}
  \country{China}}
\email{kerun_mi@njust.edu.cn}
\orcid{0000-0003-2635-2584}

\author{Guoliang Kang}
\authornote{Corresponding authors.}

\affiliation{
  \department{School of Automation Science and Electrical Engineering}
  \institution{Beihang University} 
  \city{Beijing}
  \country{China}
}
\email{kgl.prml@gmail.com}
\orcid{0000-0003-1978-2025}

\author{Guangyu Li}
\affiliation{
  \department{School of Computer Science and Engineering}
  \institution{Nanjing University of Science and Technology} 
  \city{Nanjing}
  \country{China}}
\email{guangyu.li2017@njust.edu.cn}
\orcid{0000-0003-4817-0618}

\author{Lin Zhao}
\affiliation{
  \department{School of Computer Science and Engineering}
  \institution{Nanjing University of Science and Technology}
   \city{Nanjing}
  \country{China}}
\email{linzhao@njust.edu.cn}
\orcid{0000-0002-8756-2027}

\author{Tao Zhou}
\affiliation{
  \department{School of Computer Science and Engineering}
  \institution{Nanjing University of Science and Technology}
   \city{Nanjing}
  \country{China}}
\email{taozhou.ai@gmail.com}
\orcid{0000-0002-3733-7286}

\author{Chen Gong}
\authornotemark[1]
\affiliation{
  \department{School of Automation and Intelligent Sensing}		
  \institution{Shanghai Jiao Tong University}
  \city{Shanghai}
  \country{China}}
\email{chen.gong@sjtu.edu.cn}
\orcid{0000-0002-1154-6194}

\renewcommand{\shortauthors}{Kerun Mi et al.}
\title[Cross-Domain Attribute Alignment with CLIP]{Cross-Domain Attribute Alignment with CLIP: A Rehearsal-Free Approach for Class-Incremental Unsupervised Domain Adaptation}

\begin{abstract}
  Class-Incremental Unsupervised Domain Adaptation (CI-UDA) aims to adapt a model from a labeled source domain to an unlabeled target domain, where the sets of potential target classes appearing at different time steps are disjoint and are subsets of the source classes. The key to solving this problem lies in avoiding catastrophic forgetting of knowledge about previous target classes during continuously mitigating the domain shift. Most previous works cumbersomely combine two technical components. On one hand, they need to store and utilize rehearsal target sample from previous time steps to avoid catastrophic forgetting; on the other hand, they perform alignment only between classes shared across domains at each time step. Consequently, the memory will continuously increase and the asymmetric alignment may inevitably result in knowledge forgetting. In this paper, we propose to mine and preserve domain-invariant and class-agnostic knowledge to facilitate the CI-UDA task. Specifically, via using CLIP, we extract the class-agnostic properties which we name as ``attribute''. In our framework, we learn a ``key-value'' pair to represent an attribute, where the key corresponds to the visual prototype and the value is the textual prompt. We maintain two attribute dictionaries, each corresponding to a different domain. Then we perform attribute alignment across domains to mitigate the domain shift, via encouraging visual attention consistency and prediction consistency. Through attribute modeling and cross-domain alignment, we effectively reduce catastrophic knowledge forgetting while mitigating the domain shift, in a rehearsal-free way. Experiments on three CI-UDA benchmarks demonstrate that our method outperforms previous state-of-the-art methods and effectively alleviates catastrophic forgetting. Code is available at \url{https://github.com/RyunMi/VisTA}.
\end{abstract}


%

\maketitle
\begin{figure}[htbp!]
	\centering
	\includegraphics[width=0.99\linewidth]{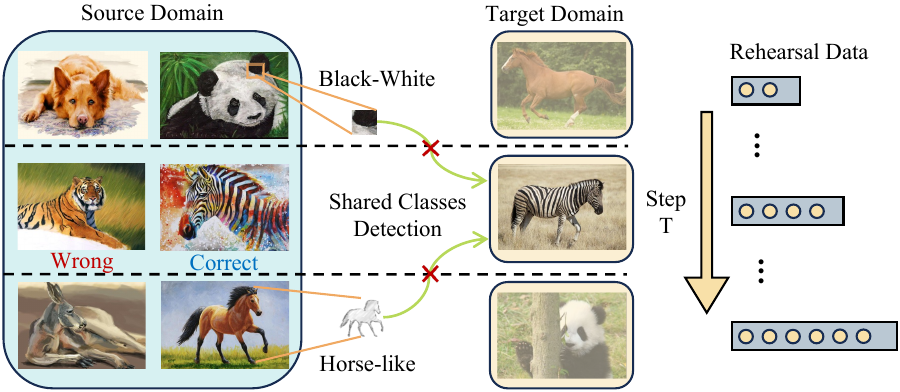}
	\caption{Existing CI-UDA methods retain knowledge by storing rehearsal target data (yellow circles), which will introduce additional computational overhead. These methods also detect the shared classes (\emph{e.g.,} ``zebra'') between domains at each time step, which may result in errors (\emph{e.g.,} ``tiger'') and ignore valuable attributes of source-private classes (\emph{e.g.,} ``Black-White'' of ``panda'' and ``Horse-like'' of ``horse'').}
	\label{CI-UDA}
    \Description{Existing CI-UDA methods retain knowledge by storing rehearsal target data (yellow circles), which will introduce additional computational overhead. These methods also detect the shared classes (\emph{e.g.,} ``zebra'') between domains at each time step, which may result in errors (\emph{e.g.,} ``tiger'') and ignore valuable attributes of source-private classes (\emph{e.g.,} ``Black-White'' of ``panda'' and ``Horse-like'' of ``horse'').}
\end{figure}

\section{Introduction}
Class-Incremental Learning (CIL)~\cite{9349197,9915459,zhou2024class} aims to handle sequentially arriving tasks, where at each time step new classes emerge. The model needs to classify all seen classes during testing without access to the task ID. CIL methods generally rely on labeled data, which is often limited due to the high cost of data annotation in real-world scenarios~\cite{Tao_2020_CVPR,E220396,ijcai2024p144,10840313}. A feasible approach is to leverage an off-the-shelf labeled dataset (\emph{i.e.,} source domain) to transfer a model to a class-incremental unlabeled dataset (\emph{i.e.,} target domain), with the source domain containing all classes. 

However, the distribution shift between domains poses significant challenges to the transferability of a model. Conventional unsupervised domain adaptation (UDA) or partial domain adaptation (PDA) methods can be utilized to mitigate the distribution shift by feature alignment~\cite{pmlr-v37-ganin15,MMwjd,NEURIPS2018_ab88b157,mdd,Kang_2019_CVPR,MMshuang,2018-6-1258} and domain-invariant knowledge transfer~\cite{Saito_2019_ICCV,9108582,zhang-etal-2021-matching,NEURIPS2023_5555cc3f,tang2025direct}.  Nevertheless, existing domain adaptation methods may suffer from catastrophic knowledge forgetting~\cite{mccloskey1989catastrophic} in class-incremental target domain, inspiring methods specifically designed for Class-Incremental Unsupervised Domain Adaptation (CI-UDA)~\cite{lin2022prototype,wei2024class,CISFUDA}.

Recently, several CI-UDA methods have been proposed~\cite{lin2022prototype,wei2024class,CISFUDA}. 
They usually consist of two technical components. 
On one hand, they typically store and utilize rehearsal data from previous target classes (as illustrated by the yellow circles in Figure~\ref{CI-UDA}) to retain historical knowledge. However, rehearsal data may not be available due to constraints such as data privacy or memory limitations (\emph{e.g.,} the memory will increase as the number of tasks increases).
On the other hand, to avoid negative knowledge transfer~\cite{9108582, MMpartial}, CI-UDA methods perform alignment only between classes shared across domains.  
However, they may still suffer from knowledge forgetting. 
Specifically, as shown in Figure~\ref{CI-UDA}, suppose the target class is ``zebra'' (a shared class) at step $T$. On one hand, the shared-class discovery process is imperfect. It may mistakenly treat private classes in source domain (\emph{i.e.,} source-private classes), such as ``tiger,'' as a shared class, leading to misalignment. On the other hand, even in source-private classes, valuable knowledge exists but may be ignored during cross-domain alignment, such as ``Black-White'' of ``panda,'' and ``Horse-like'' of ``horse,'' which are also typical properties of ``zebra''~\cite{9786767}. Recent works utilize CLIP-based~\cite{pmlr-v139-radford21a} prompt learning~\cite{zhou2022coop,10313995,Singha_2023_ICCV,Du_2024_CVPR,phan2024enhancing,MMVLM} to deal with domain adaptation problem. Technically, some methods can be directly applied to the CI-UDA setting, but as they typically do not consider the catastrophic forgetting issue, their performance is still far from satisfactory.

In this paper, we propose cross-domain \textbf{Vis}ion-\textbf{T}ext \textbf{A}ttribute \textbf{A}lignment (VisTA), a novel CI-UDA framework based on CLIP~\cite{pmlr-v139-radford21a}. 
In our framework, we aim to mine and preserve domain-invariant and class-agnostic knowledge.
Firstly, inspired by~\cite{Wang_2023_CVPR}, we employ an \textbf{Attribute Modeling} module to utilize CLIP to extract class-agnostic properties which we refer to as ``attribute''. 
We freeze the encoders of CLIP and construct a dictionary for source domain and target domain, respectively. The dictionaries store attributes in the form of ``key-value'' pairs, where the key and value bridge the visual and textual modalities. For each input image, several textual attributes are selected from the dictionary based on its visual attributes. These textual attributes, serving as prompts, are sent into CLIP to compute the class probability of the image.
Then we perform attribute alignment across domains to mitigate the domain shift, via encouraging visual attention consistency and prediction consistency. Specifically, for each image in both domains, prompts are selected from the two dictionaries to compute paired class probabilities (one from each domain). However, since the two dictionaries are learned independently, the attributes selected from the other domain are domain-specific and may not effectively contribute to the current prediction due to the domain shift.
Therefore, we introduce a \textbf{Visual Attention Consistency} (VAC) module to ensure the semantically similar attributes across domains are selected for paired prediction. VisTA then encourages \textbf{Prediction Consistency} by minimizing the Jensen–Shannon divergence between these paired probability distributions, enabling the learning of domain-invariant attributes.
Benefiting from the modeling of domain-invariant and class-agnostic attributes, we are able to deal with the CI-UDA task in a rehearsal-free manner.

In a nutshell, the contributions of our work are summarized as follows:
\begin{itemize}
	\item We propose a CI-UDA framework named VisTA, which leverages CLIP to learn class-agnostic attributes that act as prompts, achieving rehearsal-free training.
	\item VisTA learns domain-invariant attributes through attribute alignment, guided by a Visual Attention Consistency module and a Prediction Consistency loss.
	\item Extensive experiments firmly demonstrate the effectiveness of VisTA, as it achieves state-of-the-art performance on Office-31, Office-Home, and Mini-DomainNet.
\end{itemize}

\begin{figure*}[t]
	\centering
	\includegraphics[width=0.786\linewidth]{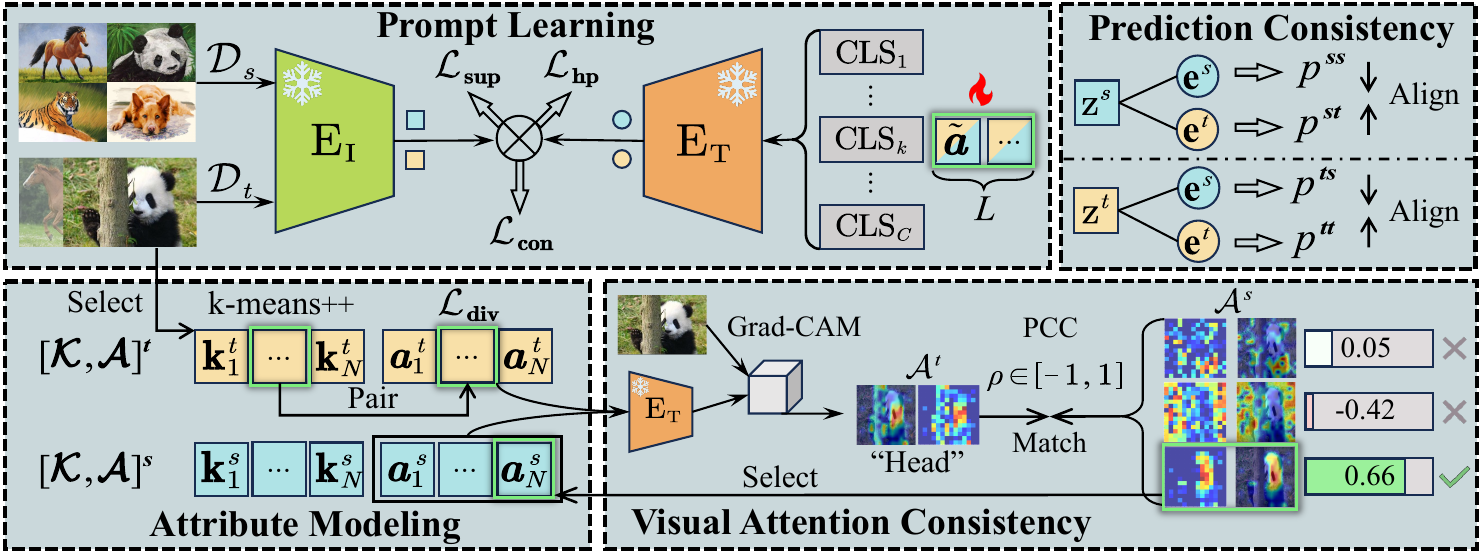}
	\caption{Framework of VisTA. Parameters of $\mathrm{E_I}$ and $\mathrm{E_T}$ in CLIP are frozen and only prompts are tunable during training. The classification loss $\mathcal{L}_{\mathrm{sup}}$ is applied separately to $\mathcal{D}_s$ (visual features $\boldsymbol{\mathrm{z}}^s$ and text embeddings $\boldsymbol{\mathrm{e}}^s$ in blue) and $\mathcal{D}_t$ ($\boldsymbol{\mathrm{z}}^t$ and $\boldsymbol{\mathrm{e}}^t$ in yellow). VisTA integrates three core modules. Attribute Modeling module for constructing domain-specific dictionaries $[\mathcal{K}, \mathcal{A}]$ of visual attributes $\mathbf{k}\in\mathcal{K}$ and textual attributes $\boldsymbol{a}\in\mathcal{A}$. Given an image from $\mathcal{D}_t$, VisTA selects target attributes $\tilde{\boldsymbol{a}}^t$ based on the cosine similarity between $\mathbf{k}^t$ and $\boldsymbol{\mathrm{z}}^t$, while selecting source attributes $\tilde{\boldsymbol{a}}^s$ via a Visual Attention Consistency module. VisTA then aligns paired class probabilities (\emph{e.g.,} $p^{ts}$ and $p^{tt}$) via a Prediction Consistency loss $\mathcal{L}_{\mathrm{con}}$. We use $p^{tt}$ for inference on $\mathcal{D}_t$.}
	\label{VisTA}
\Description{Framework of VisTA. Parameters of $\mathrm{E_I}$ and $\mathrm{E_T}$ in CLIP are frozen and only prompts are tunable during training. The classification loss $\mathcal{L}_{\mathrm{sup}}$ is applied separately to $\mathcal{D}_s$ (visual features $\boldsymbol{\mathrm{z}}^s$ and text embeddings $\boldsymbol{\mathrm{e}}^s$ in blue) and $\mathcal{D}_t$ ($\boldsymbol{\mathrm{z}}^t$ and $\boldsymbol{\mathrm{e}}^t$ in yellow). VisTA integrates three core modules. Attribute Modeling module for constructing domain-specific dictionaries $[\mathcal{K}, \mathcal{A}]$ of visual attributes $\mathbf{k}\in\mathcal{K}$ and textual attributes $\boldsymbol{a}\in\mathcal{A}$. Given an image from $\mathcal{D}_t$, VisTA selects target attributes $\tilde{\boldsymbol{a}}^t$ based on the cosine similarity between $\mathbf{k}^t$ and $\boldsymbol{\mathrm{z}}^t$, while selecting source attributes $\tilde{\boldsymbol{a}}^s$ via a Visual Attention Consistency module. VisTA then aligns paired class probabilities (\emph{e.g.,} $p^{ts}$ and $p^{tt}$) via a Prediction Consistency loss $\mathcal{L}_{\mathrm{con}}$. We use $p^{tt}$ for inference on $\mathcal{D}_t$.}
\end{figure*}

\section{Related Work}
In this section, we review some relevant works, including unsupervised domain adaptation and vision language models.

\noindent\textbf{Unsupervised Domain Adaptation.}
To mitigate distribution shift, conventional UDA methods or PDA methods typically fall into two main directions. The first direction of work aims to align the feature distributions across different domains. Common techniques include minimizing the statistical distribution metrics in the feature space directly~\cite{MMwjd,mdd,Kang_2019_CVPR} and applying adversarial learning to obtain domain-agnostic features~\cite{pmlr-v37-ganin15,NEURIPS2018_ab88b157,MMshuang}. The other direction of work seeks to transfer domain-invariant knowledge between models in source domain and target domain. For example, the works of~\cite{Saito_2019_ICCV,9108582,NEURIPS2023_5555cc3f} propose to learn an invariant classifier with consistent predictions, while~\cite{zhang-etal-2021-matching,tang2025direct} propose to improve the performance of the target domain by knowledge distillation.

However, in practice, UDA is often integrated with continual learning problems~\cite{CoTTA,MMCDA,MMDIL}. One scenario is where target data arrives in a streaming manner with different classes. In this scenario, conventional UDA methods suffer from the catastrophic forgetting problem~\cite{mccloskey1989catastrophic}.
Therefore, some recent CI-UDA methods~\cite{lin2022prototype,wei2024class,CISFUDA} have been developed to mitigate the domain shift while learning class-incremental target classes. For instance, ProCA~\cite{lin2022prototype} detects the shared classes by computing cumulative prediction probabilities of target examples and achieves adaptation through prototype alignment. PLDCA~\cite{wei2024class} builds upon ProCA and further alleviates negative transfer through domain-level and instance-level contrastive alignment. 
 Besides, GROTO~\cite{CISFUDA} designs a multi-granularity class prototype self-organization module and a prototype topology distillation module to handle CI-UDA in a source-free scenario (\emph{i.e.,} CI-SFUDA). It is worth noting that existing CI-UDA methods are suboptimal owing to continuously increasing memory and asymmetric alignment.

\noindent\textbf{Vision Language Models.}
Recent Vision Language Models (VLMs) like CLIP~\cite{pmlr-v139-radford21a} have demonstrated impressive performance on various downstream vision tasks by pretraining on large-scale image-text pairs~\cite{10445007}. VLMs typically use hand-crafted text like ``a photo of a [class\_name]'' for zero-shot prediction on downstream tasks, which preserves generalization knowledge while maintaining low computational cost. However, hand-crafted text is not always effective, and thus prompt learning has gained increasing attention. For instance, CoOp~\cite{zhou2022coop} and CoCoOp~\cite{Zhou_2022_CVPR} use learnable continuous prompts to improve the generalization performance of CLIP. MaPLe~\cite{Khattak_2023_CVPR} proposes multi-modal prompt learning to align text and image representations. Moreover, since only the prompts should be stored, certain prompt learning methods serve as rehearsal-free learners, which can be effectively utilized to address class-incremental problem~\cite{Wang_2023_CVPR}. 

However, these prompt learning methods often suffer from performance degradation when encountering domain shift problems. To address this, DAPL~\cite{10313995} introduces domain-specific and domain-agnostic prompts to learn the label distribution of target domain. AD-CLIP~\cite{Singha_2023_ICCV} learns domain-invariant prompts by combining domain style information with image content information. DAMP~\cite{Du_2024_CVPR} mutually aligns visual and textual embeddings to learn domain-agnostic prompts. PGA~\cite{phan2024enhancing} frames UDA as a multi-objective optimization problem and promotes consensus among per-objective gradients. Although existing prompt learning methods have shown quite promising performance, they cannot deal with CI-UDA, as the historical knowledge encoded in prompts may be overwritten by new class information, leading to catastrophic forgetting. To this end, in this paper, we propose a rehearsal-free method based on prompt learning for CI-UDA, which effectively reduces catastrophic forgetting while mitigating distribution shift.

\section{Preliminaries}
\subsection{CI-UDA Problem Formulation}
In CI-UDA, we consider two domains, including the labeled source domain $\mathcal{D}_s=\{(\boldsymbol{\mathrm{x}}^s_i, \boldsymbol{\mathrm{y}}^s_i)\}^{n_s}$ (where $\boldsymbol{\mathrm{x}}^s_i$ denotes the $i$-th source image, $\boldsymbol{\mathrm{y}}^s_i\in\{1,2\cdots, C\}$ denotes the label of $i$-th image and $n_s$ means the number of source examples),
and unlabeled target domain $\mathcal{D}_t=\{\boldsymbol{\mathrm{x}}^t_j\}^{n_t}$. 
The sample in $\mathcal{D}_s$ is available at all time steps and covers all the considered $C$ classes, and the sample in $\mathcal{D}_t$ comes incrementally. For each time step, 
the underlying class set of $\mathcal{D}_t$ is only a subset of $\{1, \cdots, C\}$ and the class sets of different time steps are disjoint.

The goal of CI-UDA is to learn a model by leveraging data from $\mathcal{D}_s$ and class-incremental $\mathcal{D}_t$, such that it performs well on all seen classes of $\mathcal{D}_t$ during testing. 

\subsection{Prompt Learning in CLIP}
CLIP~\cite{pmlr-v139-radford21a}, a prominent VLM, pretrains an image encoder $\mathrm{E_I}$ and a text encoder $\mathrm{E_T}$ on large-scale image-text pairs to learn well-aligned visual and textual representations. In downstream tasks, class-specific textual prompts $\boldsymbol{\mathrm{P}}_k$ may be utilized for each class $k$ (\emph{e.g.,} ``a photo of $[\text{CLS}_k]$,'' where $[\text{CLS}_k]$ is the $k$-th class name). CLIP predicts the probability that an input image $\boldsymbol{\mathrm{x}}$ belongs to each class by computing the cosine similarity between the visual feature $\boldsymbol{\mathrm{z}}=\mathrm{E_I}(\boldsymbol{\mathrm{x}})\in\mathbb{R}^D$ (where $D$ is the feature dimension) and the class-wise text embeddings $\boldsymbol{\mathrm{w}}_k=\mathrm{E_T}(\boldsymbol{\mathrm{P}}_k)\in\mathbb{R}^D, k=1, \cdots, C$:
\begin{equation}\label{clip}
	p(\boldsymbol{\mathrm{y}}=k|\boldsymbol{\mathrm{x}})=\frac
	{\exp \left(\cos\langle \boldsymbol{\mathrm{w}}_k, \boldsymbol{\mathrm{z}} \rangle  / \tau \right)}
	{\sum^C_{c=1} \exp \left(\cos\langle \boldsymbol{\mathrm{w}}_c, \boldsymbol{\mathrm{z}} \rangle / \tau \right)},
\end{equation}
where $\cos\langle \cdot, \cdot \rangle$ is cosine similarity, and $\tau$ is temperature parameter.

However, such hand-crafted prompts may be suboptimal. To further enhance the performance of CLIP in downstream tasks, CoOp~\cite{zhou2022coop} introduces learnable continuous vectors $\boldsymbol{V}$ with length $M$ to replace hand-crafted prompt templates. The learnable prompt for class $k$ is then defined as $\boldsymbol{\mathrm{P}}_k = [\boldsymbol{V}_1,\boldsymbol{V}_2,\cdots,\boldsymbol{V}_{M},\text{CLS}_k]$.
During training, the prompt $\mathbf{P}_k$ is updated to minimize the cross-entropy loss on sample from downstream tasks.
For the inference, they follow the same way which utilizes Eq.~(\ref{clip}) to predict the label for each example. The only difference is that they use the learned prompts instead of manually designed prompts for prediction.
\section{Method}
As shown in Figure~\ref{VisTA}, at time step $T$, VisTA processes $\mathcal{D}_s$ and the current $\mathcal{D}_t$ through $\mathrm{E_I}$, generating visual features $\boldsymbol{\mathrm{z}}^s$ and $\boldsymbol{\mathrm{z}}^t$. Taking $\boldsymbol{\mathrm{z}}^t$ as an example, we retrieve $L$ textual attributes from target dictionary based on the cosine similarity between $\boldsymbol{\mathrm{z}}^t$ and visual attributes (Section~\ref{Learn}). These attributes serve as textual prompts for $\mathrm{E_T}$.
Concurrently, we propose a Visual Attention Consistency module (Section~\ref{vac}), 
which applies a Grad-CAM-based attention heatmap matching mechanism to select $L$ source attributes with similar semantic concepts as prompts.
This process yields two class probability distributions for $\boldsymbol{\mathrm{z}}^t$. 
The target attributes then are optimized through a Prediction Consistency loss (Section~\ref{pcl}) to enable the learning of class-agnostic and domain-invariant knowledge.
A similar procedure is adopted for $\boldsymbol{\mathrm{z}}^s$. 
Finally, Section~\ref{obj} discusses two regularization terms and the training objective of VisTA.

\subsection{Attribute Modeling}\label{Learn}
In CI-UDA setting, the underlying class set of target domain at each time step is only a subset of that of source domain and is disjoint from that of previous time steps. 
If mitigating knowledge forgetting at the class-level, we may inevitably need to store previous target sample and discover shared classes between domains at each time step to perform alignment\textemdash a cumbersome process that may introduce too much noise during training.
In this paper, we aim to mitigate the knowledge forgetting in CI-UDA at the ``attribute'' level. 
The ``attribute'' refers to the basic components which combine to support correct predictions.

Specifically, with CLIP extracting visual and textual features, we represent each attribute as a ``key-value'' pair, where the value refers to the textual representation of attribute and the key refers to the visual representation of attribute, in other words, the visual prototype. Formally, attributes are denoted as:
\begin{equation}\label{dict}
	[\mathcal{K},\mathcal{A}]\coloneq[\{\boldsymbol{\mathrm{k}}_1,\boldsymbol{a}_1\},\{\boldsymbol{\mathrm{k}}_2,\boldsymbol{a}_2\},\cdots,\{\boldsymbol{\mathrm{k}}_{N},\boldsymbol{a}_{N}\}],
\end{equation}
where each key $\boldsymbol{\mathrm{k}}_i\in\mathbb{R}^D~(i=1,2,\dots,N)$ is designed to capture the visual attributes of an image $\boldsymbol{\mathrm{x}}$, and each value $\boldsymbol{a}_i \in \mathbb{R}^{M \times D}~(i=1,2,\dots,N)$ encodes the textual description of a specific attribute.

VisTA maintains a source attribute dictionary $[\mathcal{K}^s, \mathcal{A}^s]$ and a target attribute dictionary $ [\mathcal{K}^t, \mathcal{A}^t]$.  
We design specific update strategies for attributes to learn class-agnostic knowledge.

\noindent\textbf{Visual Attributes.} 
We perform k-means++ clustering~\cite{k-means++} on all source features to obtain source visual attributes $\mathcal{K}^s$ before training and keep $\mathcal{K}^s$ fixed during training. As the target classes appearing at different time steps are disjoint in CI-UDA, target visual attributes $\mathcal{K}^t$ are initialized via k-means++ clustering on the data from the first time step in the class-incremental training sequences. During each subsequent time step in class-incremental learning process, we apply a moving average strategy to update $\mathcal{K}^t$.

\noindent\textbf{Textual Attributes.} VisTA randomly initializes $\mathcal{A}^s$ and $\mathcal{A}^t$. These textual attributes are then modeled through supervised training or self-training to mine and preserve class-agnostic knowledge.

Given an image $\boldsymbol{\mathrm{x}}$ from both domains, we select the top-$L$ visual attributes $\widetilde{\mathcal{K}}\subseteq \mathcal{K}$ based on their cosine similarity to $\boldsymbol{\mathrm{x}}$. The paired textual attributes are then indexed from the dictionary as $\widetilde{\mathcal{A}}=\widetilde{\boldsymbol{a}}_{{1:L}}$. These textual attributes are concatenated with the class name to form the prompt:
\begin{equation}\label{prompt}
	\boldsymbol{\mathrm{P}}_k(\widetilde{\mathcal{A}}) = [\widetilde{\boldsymbol{a}}_1, \widetilde{\boldsymbol{a}}_2,\cdots,\widetilde{\boldsymbol{a}}_L,\text{CLS}_k], k=1, \cdots, C.
\end{equation}

The source textual attributes $\mathcal{A}^s$ are optimized by minimizing cross-entropy loss on labeled $\mathcal{D}_s$:
\begin{equation}\label{ce}
	\mathcal{L}_{\mathrm{sup}}^s=-\log p\left(\boldsymbol{\mathrm{y}}=\boldsymbol{\mathrm{y}}^s|\boldsymbol{\mathrm{x}}^s\right).
\end{equation}

Moving to unlabeled $\mathcal{D}_t$, target textual attributes $\mathcal{A}^t$ are optimized by minimizing self-training loss:
\begin{equation}\label{sl}
	\mathcal{L}_{\mathrm{sup}}^t=-\mathbb{I}(\max(\hat{p})\geq\gamma)\log p\left(\boldsymbol{\mathrm{y}}=\hat{\boldsymbol{\mathrm{y}}}^{t}|\boldsymbol{\mathrm{x}}^{t}\right),
\end{equation}
where $\hat{\boldsymbol{\mathrm{y}}}^{t}$ represents the pseudo-label, $\mathbb{I}(\cdot)$ is the indicator function, $\gamma$ is a threshold to select high-confidence pseudo-label of target example, and $\hat{p}$ denotes the debiased soft pseudo-label. 
This debiasing follows DebiasPL~\cite{Wang_2022_CVPR}, which aims to enhance the reliability of pseudo-labels and has recently been adopted in several CLIP-based UDA methods~\cite{Lai_2023_ICCV,Li_2024_CVPR}.
In detail, $\hat{p}$ is computed as:
\begin{equation}\label{debias}
	\hat{p} = p - \tau \log q,~q\leftarrow mq + (1-m) \frac{1}{B}\sum_{j=1}^B p_j,
\end{equation} 
where $m$ is a momentum coefficient, $\tau$ is a debias factor, $B$ denotes the batch size, and $q$ is initialized before training as a uniform probability vector over $C$ classes.

\subsection{Cross-Domain Attribute Alignment}
To mitigate the domain shift in CI-UDA, VisTA performs cross-domain attribute alignment through a Visual Attention Consistency module and a Prediction Consistency loss.
We want to mention that previous work AttriCLIP~\cite{Wang_2023_CVPR} can also extract class-agnostic attributes to benefit general continual learning, but cannot guarantee the domain-invariant property, which is crucial for CI-UDA. Hence, we design two novel modules to address this limitation.

\begin{table*}[t]
	\centering
	\caption{Final Accuracy (\%) on Office-31. DA, CI, and RF respectively represent domain adaptation, class-incremental, and rehearsal-free. The $\ast$ indicates the result is cited from GROTO~\cite{CISFUDA}.}
	\scalebox{0.81}{  
		\begin{tabular}{lccc|c|cccccccc}
			\toprule
			Method & DA & CI & RF & Backbone & A$\rightarrow$D & A$\rightarrow$W & D$\rightarrow$A & D$\rightarrow$W & W$\rightarrow$A & W$\rightarrow$D & Avg.\\
			\midrule
			Vanilla$^\ast$~(ICLR21)& --  & -- & -- & \multirow{3}{*}{ViT} & 82.4 & 80.0 & 70.8 & 83.1 & 75.1 & 87.4 & 79.8  \\
			ProCA~(ECCV22)& \cmark  & \cmark & \xmark & & 91.1 & 86.3 & 75.6 & \textbf{98.3} & 77.2 & \textbf{99.4} &  88.0 \\
			PLDCA~(TIP24)& \cmark  & \cmark & \xmark & & \textbf{91.5} & \textbf{90.7} & 75.6 & 96.9 & 77.8 & 99.4 & 88.7 \\
			\midrule
			Vanilla~(ICML21)& --  & -- & -- & \multirow{8}{*}{CLIP} & 77.2 & 75.2 & 79.7 & 75.2 & 79.7 & 77.2 & 77.4 \\
			CoOp~(IJCV22)& --  & -- & -- & & 89.0 & 87.0 & 83.0 & 96.1 & 81.5 & 98.1 & 89.1 \\
			AttriCLIP~(CVPR23)& \xmark  & \cmark & \cmark & & 82.8 & 85.7 & 79.5 & 94.4 & 79.8 & 95.4 & 86.3\\
			AD-CLIP~(ICCVW23)& \cmark  & \xmark & \cmark & & 85.1 & 85.9 & 83.0 & 95.1 & 81.4 & 93.6 & 87.4 \\
			DAMP~(CVPR24)& \cmark  & \xmark & \cmark & & 82.8 & 79.3 & 81.3 & 90.3 & 80.1 & 91.7 & 84.3 \\
			PGA~(NeurIPS24)& \cmark  & \xmark & \cmark & & 49.0 & 68.4 & 81.2 & 88.7 & 81.3 & 81.8 & 75.1\\
			DAPL~(TNNLS25)& \cmark  & \xmark & \cmark  & & 79.1 & 76.7 & 79.5 & 79.2 & 81.1 & 79.3 & 79.2 \\
			\rowcolor[HTML]{d1cfcf}
			VisTA~(Ours)& \cmark  & \cmark & \cmark & & 87.8 & 88.2 & \textbf{84.5} & 95.2 & \textbf{84.0} & 96.5 & \textbf{89.4} \\
			\bottomrule
		\end{tabular}
	}
	\label{Final-31}
\end{table*}

\noindent\textbf{Visual Attention Consistency (VAC).}\label{vac}
Taking $\boldsymbol{\mathrm{x}}^t$ as an example, we select $L$ visual attributes $\widetilde{\mathcal{A}}^t$ from target dictionary through cosine similarity between $\boldsymbol{\mathrm{x}}^t$ and $\mathcal{K}^t$. Then we select $L$ source visual attributes $\widetilde{\mathcal{A}}^s$ for alignment.
Note that the update of visual attributes mentioned previously (Section 4.1) is domain-specific, so selecting $\widetilde{\mathcal{A}}^s$ for $\boldsymbol{\mathrm{x}}^t$ by cosine similarity may introduce bias due to domain shift. Therefore, based on Grad-CAM~\cite{selvaraju2017grad}, VisTA proposes an attention heatmap matching mechanism for cross-domain attribute selection.

Specifically, we compute $L$ target CAM scores for $\boldsymbol{\mathrm{x}}^t$ as:
\begin{equation}
	S_{\mathrm{CAM}}^t=\frac
	{\exp \left(\cos\langle \mathrm{E_T}(\boldsymbol{a}_m^t), \boldsymbol{\mathrm{z}^t} \rangle  / \tau \right)}
	{\sum^{N}_{n=1} \exp \left(\cos\langle \mathrm{E_T}(\boldsymbol{a}_n^t), \boldsymbol{\mathrm{z}^t} \rangle / \tau \right)},
	\label{cam}
\end{equation} 
where $\boldsymbol{a}_m^t\in\widetilde{\mathcal{A}}^t,(m=1,\cdots,L)$ are individual textual attributes.

Then we use the gradients of $S_{\mathrm{CAM}}^t$ with respect to the features from a layer as weights, and perform a weighted aggregation to highlight attribute-level discriminative regions. This procedure follows Grad-CAM to generate $L$ heatmaps $\boldsymbol{\mathrm{H}}^t$ for $\boldsymbol{\mathrm{x}}^t$.  
In the same way, we also compute $N$ source CAM scores $S_{\mathrm{CAM}}^s$ for all source attributes $\mathcal{A}^s$, thereby obtaining $N$ candidate heatmaps $\boldsymbol{\mathrm{H}}^s$ for $\boldsymbol{\mathrm{x}}^t$. 

To select $\widetilde{\mathcal{A}}^s$ from $N$ candidates, VisTA quantify the visual attention consistency between flattened $\boldsymbol{\mathrm{H}}^t$ and $\boldsymbol{\mathrm{H}}^s$ using the Pearson Correlation Coefficient (PCC): $\rho = \frac{\mathrm{cov}(\boldsymbol{\mathrm{H}}^s, \boldsymbol{\mathrm{H}}^t)}{\sigma_{\boldsymbol{\mathrm{H}}^s}\sigma_{\boldsymbol{\mathrm{H}}^t}}\in[-1,1]$, where $\mathrm{cov}$ represents the covariance, and $\sigma$ denotes the standard deviation. A higher value of $\rho$ highlights attributes for prioritized selection.  

The visual attention consistency in images can be interpreted as the similarity of semantic concepts. As illustrated in the bottom-right panel of Figure~\ref{VisTA}, we analyze a ``panda'' image from $\mathcal{D}_t$. The attention heatmap $\boldsymbol{\mathrm{H}}^t$ of the selected target attribute $\widetilde{\mathcal{A}}^t$ exhibits concentrated activations in the central-right region of the image, likely corresponding to the semantic concept of ``Head''. Notably, the selected source attribute $\widetilde{\mathcal{A}}^s$ with the highest $\rho=0.66$ explicitly aligns with the same semantic concept of ``Head''.

Therefore, for $L$ selected target attributes $\widetilde{\mathcal{A}}^t$ and total $N$ source attributes $\mathcal{A}^s$, we compute $L \times N$ PCC $\rho$. Leveraging these values, we match and identify the top-$L$ attributes $\widetilde{\mathcal{A}}^s$ as the cross-domain attribute selection result for $\boldsymbol{\mathrm{x}}^t$. The procedure to employ VAC module is identical for $\boldsymbol{\mathrm{x}}^s$.
Importantly, PCC, defined as cosine similarity on normalized vectors, reduces the influence of global style and is thus suitable for quantifying the visual attention consistency.

\noindent\textbf{Prediction Consistency.}\label{pcl}
To achieve attribute alignment, we introduce a Prediction Consistency loss applied to the attributes selected by the VAC module, which enforces domain invariance for these attributes exhibiting semantic similarity. 

As illustrated in the top-right panel of Figure~\ref{VisTA}, the selected attributes $\widetilde{\mathcal{A}}^s$ and $\widetilde{\mathcal{A}}^t$ are served as textual prompts for $\mathrm{E_T}$ to generate class-wise text embeddings $\boldsymbol{\mathrm{e}}_k^{s,t}=\mathrm{E_T}(\boldsymbol{\mathrm{P}}_k(\widetilde{\mathcal{A}}^{s,t})), k=1, \dots, C$. This enables the computation of paired class probabilities for $\boldsymbol{\mathrm{x}}$ using Equation~(\ref{clip}). 
For example, given an image $\boldsymbol{\mathrm{x}}^t$ is sampled from $\mathcal{D}_t$, we utilize prompts $\boldsymbol{\mathrm{P}}(\widetilde{\mathcal{A}}^s)$ and $\boldsymbol{\mathrm{P}}(\widetilde{\mathcal{A}}^{t})$ to obtain class probability vectors $p^{ts}$ and $p^{tt}$, respectively. For an image $\boldsymbol{\mathrm{x}}^s$ is sampled from $\mathcal{D}_s$, analogous terms $p^{ss}$ and $p^{st}$ are computed. Here, the superscripts of $p$ denote the domain of the $\boldsymbol{\mathrm{x}}$ (first symbol) and the dictionary from which the prompt is selected (second symbol).

The Prediction Consistency loss is achieved by minimizing the Jensen–Shannon divergence $\mathrm{D_{JS}}$ between each pair of class probabilities:
\begin{gather}\label{con}
	\mathcal{L}_{\mathrm{con}}=\mathrm{D_{JS}}(p^{ss},p^{st})+\mathrm{D_{JS}}(p^{tt},p^{ts}).
\end{gather}

In this way, VisTA effectively reduces catastrophic knowledge forgetting while mitigating the domain shift by learning class-agnostic and domain-invariant attributes. During inference, we use $p^{tt}$ as the prediction score of target example.

\begin{table*}[t]
	\centering
	\caption{Final Accuracy (\%) on Office-Home. DA, CI, and RF respectively represent domain adaptation, class-incremental, and rehearsal-free. The $\diamondsuit$ indicates the result is cited from DAPL~\cite{10313995}, and the $\ast$ indicates the result is cited from GROTO~\cite{CISFUDA}.}
	\scalebox{0.8}{  
		\begin{tabular}{lccc|c|cccccccccccccc}
			\toprule
			Method & DA & CI & RF & Backbone &A$\rightarrow$C & A$\rightarrow$P & A$\rightarrow$R & C$\rightarrow$A & C$\rightarrow$P & C$\rightarrow$R & P$\rightarrow$A & P$\rightarrow$C & P$\rightarrow$R & R$\rightarrow$A & R$\rightarrow$C & R$\rightarrow$P & Avg.\\
			\midrule
			Vanilla$^\ast$~(ICLR21)& --  & -- & -- &  \multirow{3}{*}{ViT}  & 53.2 & 77.7 & 82.1 & 69.1 & 76.6 & 78.7 & 67.8 & 50.8 & 82.1 & 73.0 & 50.2 & 81.8 & 70.3\\
			ProCA~(ECCV22)& \cmark  & \cmark & \xmark & & 56.8 & 81.9 & 89.9 & 68.4 & 80.9 & 83.9 & 70.1 & 51.3  & 89.9 & 77.4 & 50.3 & 87.9 & 74.1 \\
			PLDCA~(TIP24)& \cmark  & \cmark & \xmark & & 58.2 & 83.2 & 89.5 & 76.5 & 83.3 & 85.3 & 74.2 & 54.5 & 87.6 & 80.4 & 55.8  & 90.0 & 76.6 \\
			\midrule
			Vanilla$^\diamondsuit$~(ICML21)& --  & -- & -- & \multirow{8}{*}{CLIP} & 67.8 & 89.0 & 89.8 & 82.9 & 89.0 & 89.8 & 82.9 & 67.8 & 89.8 & 82.9 & 67.8 & 89.0 & 82.4\\
			CoOp~(IJCV22)& --  & -- & -- & & 70.6 & 88.9 & 89.8 & 81.4 & 88.4 & 87.9 & 79.8 & 69.6 & 89.6 & 83.3 & 71.6 & 91.9 & 82.7\\
			AttriCLIP~(CVPR23)& \xmark  & \cmark & \cmark & & 67.4 & 87.9 & 89.4 & 84.0 & 90.2 & 87.3 & 82.3 & 67.3 & 88.7 & 83.9 & 69.0 & 90.4 & 82.3\\
			AD-CLIP~(ICCVW23)& \cmark  & \xmark & \cmark & & 70.9 & 91.3 & 89.3 & 82.2 & 91.3 & 90.4 & 80.4 & 71.9 & 90.5 & 83.3 & 71.4 & 92.7 & 83.8\\
			DAMP~(CVPR24)& \cmark  & \xmark & \cmark & & 68.3 & 86.7 & 88.7 & 81.3 & 90.1 & 88.2 & 81.5 & 68.3 & 89.1 & 80.3 & 69.0   & 90.7 & 81.9\\
			PGA~(NeurIPS24)& \cmark  & \xmark & \cmark & & 61.6 & 70.9 & 85.3 & 55.7 & 83.7 & 84.3 & 58.5 & 59.7 & 81.3 & 59.4 & 63.1 & 82.1 & 70.5\\
			DAPL~(TNNLS25)& \cmark  & \xmark & \cmark & & 69.3 & 91.2 & 90.5 & 83.3 & 90.1 & 90.1 & 83.3 & 68.9 & 90.2 & 82.2 & 70.4 & 90.8 & 83.4 \\
			\rowcolor[HTML]{d1cfcf} 
			VisTA~(Ours)& \cmark  & \cmark & \cmark & & \textbf{71.8} & \textbf{92.8} & \textbf{91.3} & \textbf{84.4} & \textbf{92.9} & \textbf{91.1} & \textbf{84.7} & \textbf{72.8} & \textbf{91.5} & \textbf{84.8} & \textbf{72.2} & \textbf{92.8} & \textbf{85.3}\\
			\bottomrule
		\end{tabular}
	}
	\label{Final-Home}
\end{table*}

\begin{table*}[t]
	\centering
	\caption{Final Accuracy (\%) on Mini-DomainNet. DA, CI, and RF respectively represent domain adaptation, class-incremental, and rehearsal-free. The $\diamondsuit$ indicates the result is cited from DAPL~\cite{10313995}.}
	\scalebox{0.8}{  
		\begin{tabular}{lccc|c|cccccccccccccc}
			\toprule
			Method & DA & CI & RF & Backbone & C$\rightarrow$P     & C$\rightarrow$R    & C$\rightarrow$S       & P$\rightarrow$C         & P$\rightarrow$R      & P$\rightarrow$S         & R$\rightarrow$C         & R$\rightarrow$P        & R$\rightarrow$S         & S$\rightarrow$C         & S$\rightarrow$P         & S$\rightarrow$R         & Avg.\\
			\midrule
			Vanilla~(ICLR21)& --  & -- & -- & \multirow{3}{*}{ViT} & 50.2 & 61.9 & 43.8 & 50.8 & 76.8 & 51.6 & 58.9 & 60.0 & 44.9 & 60.8 & 48.6  & 66.4 & 56.2\\
			ProCA~(ECCV22)& \cmark  & \cmark & \xmark & & 50.7 & 73.3 & 51.0 & 56.2 & 83.0 & 37.0 & 56.8 & 55.2 & 39.7 & 57.7 & 44.3 & 70.8 & 56.3 \\
			PLDCA~(TIP24)& \cmark  & \cmark & \xmark & & 56.2 & 71.5 & 51.9 & 61.5 & 78.6 & 52.5 & 60.5 & 66.2 & 42.9 &  62.1 & 58.8 & 71.9 & 61.2 \\
			\midrule
			Vanilla$^\diamondsuit$~(ICML21)& --  & -- & -- & \multirow{8}{*}{CLIP} & 80.3 & 90.5 & 77.8 & 82.7 & 90.5 & 77.8  & 82.7 & 80.3 & 77.8 & 82.7  & 80.3 &  90.5 &  82.8\\
			CoOp~(IJCV22)& --  & -- & -- &  & 78.5 & 88.5 & 77.8 & 82.5 & 88.8 & 73.3  & 83.0 & 77.8 & 79.2 & 82.8  & 76.5 &  86.2 &  81.2\\
			AttriCLIP~(CVPR23)& \xmark  & \cmark & \cmark & & 73.8 & 74.7 & 76.7 & 81.8 & 84.0 & 78.2 & 83.0 & 70.5 & 73.3 & 83.8 & 78.0 & 78.2 & 78.0 \\
			AD-CLIP~(ICCVW23)& \cmark  & \xmark & \cmark & & 77.5 & 89.5 & 75.7 & 85.7 & 90.8 &  77.5 & 83.8 & 78.7 & 79.8 & 85.2 & 79.5 & 89.8 & 82.8 \\
			DAMP~(CVPR24)& \cmark  & \xmark & \cmark & & 77.8 & 85.0 & 76.5 & 82.7   & 87.3   & 79.2  & 79.7  & 74.0 & 78.5  & 84.3   & 74.7  & 82.3 & 80.2\\
			PGA~(NeurIPS24)& \cmark  & \xmark & \cmark & & 80.9 & 90.2 & 80.3 & 80.7 & 89.3 & 76.2 & 65.8 & 75.6 & 76.0 & 84.4 & 81.8 & 89.6 & 80.9\\
			DAPL~(TNNLS25)& \cmark  & \xmark & \cmark & & 81.5 & 90.3 & 78.3 & 85.5 & 91.0 & 78.0  & 85.0  & 81.0            & 77.3 & 85.7  & 80.2   & 90.8 & 83.7 \\
			\rowcolor[HTML]{d1cfcf} 
			VisTA~(Ours)& \cmark  & \cmark & \cmark & & \textbf{84.0} &\textbf{91.5} & \textbf{81.5} & \textbf{85.7} & \textbf{91.7} & \textbf{81.7} & \textbf{85.0} & \textbf{83.0} & \textbf{81.2} & \textbf{86.0} & \textbf{82.3} & \textbf{91.2} & \textbf{85.4}\\
			\bottomrule
		\end{tabular}
	}
	\label{Final-Domain}
\end{table*}

\begin{table*}[t]
	\centering
	\caption{Step-level Accuracy (\%) on Office-Home and Mini-DomainNet. DA, CI, and RF respectively represent domain adaptation, class-incremental, and rehearsal-free.}
	\scalebox{0.8}{  
		\begin{tabular}{lccc|ccccccc|ccccccc}
			\toprule
			\multirow{2}{*}{Method} & \multirow{2}{*}{DA} & \multirow{2}{*}{CI} & \multirow{2}{*}{RF} & \multicolumn{7}{c|}{Office-Home} & \multicolumn{7}{c}{Mini-DomainNet} \\
			\cmidrule(lr){5-11} \cmidrule(lr){12-18} & & & & Step 1 & Step 2 & Step 3 & Step 4 & Step 5 & Step 6 & Avg. & Step 1 & Step 2 & Step 3 & Step 4 & Step 5 & Step 6 & Avg.\\
			\midrule
			ProCA~(ECCV22)& \cmark  & \cmark & \xmark & 73.2          & 74.0          & 72.3          & 73.3          & 73.8         & 74.1          & 73.5 & 59.8 & 55.3 & 57.8 & 57.1 & 55.1 & 56.3 & 56.9 \\
			PLDCA~(TIP24)& \cmark  & \cmark & \xmark & 70.8          & 74.4          & 73.7          & 74.8          & 76.1        & 76.6          & 74.4 &  61.0 & 60.9 & 63.7 & 61.4 & 60.4 & 61.2 & 61.4 \\
			AttriCLIP~(CVPR23)& \xmark  & \cmark & \cmark & \textbf{82.7}          & 84.5          & 81.8          & 82.7          & 83.1        & 82.3          & 82.9 & 74.1 & 77.9 & 80.2 & 79.9 & 80.0 & 78.0 & 78.4\\ 
			AD-CLIP~(ICCVW23)& \cmark  & \xmark & \cmark & 79.5          & 83.5          & 81.8          & 83.5          & 83.3        & 83.8          & 82.6 & 81.5 & 83.2 & 84.4 & 83.9 & 83.6 & 82.1 & 83.1\\
			DAMP~(CVPR23)& \cmark  & \xmark & \cmark & 81.0            & 83.3          & 82.1          & 82.4          & 82.0          & 81.9          & 82.1  &  82.5 & 83.5 & 84.2 & 83.2 & 82.3 & 80.2 & 82.6\\
			DAPL~(TNNLS25)& \cmark  & \xmark & \cmark & 81.6          & 83.7          & 81.9          & 82.9          & 83.4        & 83.4          & 82.8 & \textbf{83.7} & \textbf{85.6} & 86.3 & 85.5 & 84.3 & 83.7 & 84.9\\
			\rowcolor[HTML]{d1cfcf}
			VisTA~(Ours)& \cmark  & \cmark & \cmark & 80.3   & \textbf{84.7}   & \textbf{84.2} & \textbf{85.1} & \textbf{85.1} & \textbf{85.3} & \textbf{84.1} & 83.5 & 85.5 & \textbf{87.2} & \textbf{86.6} & \textbf{86.1} & \textbf{85.4} & \textbf{85.7}\\
			\bottomrule
		\end{tabular}
	}
	\label{Step-Home_Domain}
\end{table*}

\subsection{Training Objective}\label{obj}
Notably, we aim to enhance the generalization capacity of $\mathcal{A}^s$, enabling it to effectively guide predictions in $\mathcal{D}_t$.
Inspired by~\cite{yao2023visual}, VisTA proposes a regularization loss which minimizes the distance between class-wise embeddings $\boldsymbol{\mathrm{e}}^s_k$ generated by $\mathcal{A}^s$ and those derived from hand-crafted prompts ($\boldsymbol{\mathrm{w}}_k$):
\begin{equation}\label{hp}
	\mathcal{L}_{\mathrm{hp}}=\sum_{k\operatorname{=}1}^C|\boldsymbol{\mathrm{e}}^s_k-\boldsymbol{\mathrm{w}}_k|.
\end{equation}

Finally, to promote the diversity of textual attributes, a regularization loss is applied separately to both $\mathcal{D}_s$ and $\mathcal{D}_t$ to enforce orthogonality among the attributes within $\mathcal{A}$:
\begin{equation}\label{div}
	\mathcal{L}_{\mathrm{div}}=\frac{1}{N(N-1)}\sum_{m=1}^{N}\sum_{n=m+1}^{N}|\cos\langle \mathrm{E_T}(\boldsymbol{a}_m),\mathrm{E_T}(\boldsymbol{a}_n)\rangle|.
\end{equation}

As a result, the final optimization objective of VisTA is:
\begin{equation}\label{final}
	\mathcal{L} = \mathcal{L}_{\mathrm{sup}}+\lambda_1\mathcal{L}_{\mathrm{con}}+\lambda_2 \mathcal{L}_{\mathrm{hp}}+\lambda_3 \mathcal{L}_{\mathrm{div}},
\end{equation}
where $\lambda_1$, $\lambda_2$, and $\lambda_3$ are non-negative trade-off weights, and $\mathcal{L}_{\mathrm{sup}}=\mathcal{L}_{\mathrm{sup}}^s+\mathcal{L}_{\mathrm{sup}}^t$ is the classification loss.

\section{Experiment}
\subsection{Experimental Setup}
\noindent\textbf{Datasets.}
\textbf{Office-31}~\cite{office31} includes 31 categories from three domains: Amazon (A), DSLR (D), Webcam (W), totaling 4,600 images. \textbf{Office-Home}~\cite{Venkateswara_2017_CVPR} comprises 65 categories across four distinct domains: Art (A), Clipart (C), Product (P), and Real World (R), totaling 15,500 images. \textbf{Mini-DomainNet} is a subset of DomainNet~\cite{Peng_2019_ICCV} and includes 126 categories across four domains: Clipart (C), Painting (P), Real World (R), and Sketch (S).

Following ProCA~\cite{lin2022prototype}, we divide each domain of Office-31 into three disjoint subsets, each containing 10 classes in alphabetical order, and divide each domain of Office-Home into six disjoint subsets, each containing 10 classes in an order consistent with ProCA~\cite{lin2022prototype}. Additionally, as the first CI-UDA method to handle Mini-DomainNet, we divide each domain into six disjoint subsets, each containing 20 classes in alphabetical order. More details of datasets construction are in \textbf{Appendix~\ref{sec:appendix_Data}}.

\noindent\textbf{Baseline Methods.}
We compare VisTA with: (1) source-only: ViT-B/16~\cite{dosovitskiy2021an} and CoOp~\cite{zhou2022coop}; (2) zero-shot: CLIP; (3) existing CI-UDA methods: ProCA~\cite{lin2022prototype} and PLDCA~\cite{wei2024class}; (4) prompt learning method for CIL: AttriCLIP~\cite{Singha_2023_ICCV}; (5) prompt learning methods for UDA: AD-CLIP~\cite{Singha_2023_ICCV}, DAMP~\cite{Du_2024_CVPR}, PGA~\cite{phan2024enhancing}, and  DAPL~\cite{10313995}.

\noindent\textbf{Implementation Details.} We use ViT-B/16 as the image encoder for VisTA and all baseline methods. Details on the hyperparameters of VisTA, as well as the training procedures for VisTA and several baseline methods, are provided in \textbf{Appendix~\ref{sec:hyper}}. We analyze the sensitivity of our method to hyperparameters in Section~\ref{hyp}.

\noindent\textbf{Evaluation Metrics.}
To comprehensively evaluate the performance of VisTA, we employ three metrics for CI-UDA:

\noindent 1) \textbf{Final Accuracy}: the classification accuracy across all classes at the final time step for each adaptation task; 

\noindent 2) \textbf{Step-level Accuracy}: the average classification accuracy over all adaptation tasks at each time step; 

\noindent 3) \textbf{S-1 Accuracy}: the average classification accuracy of all adaptation tasks at each time step for classes in Step-1.

\subsection{Comparisons with previous state-of-the-arts}\label{soat}
\begin{figure}[t]
  \centering
  \subfloat[S-1 Acc. (\%)]{
      \includegraphics[width=0.235\textwidth]{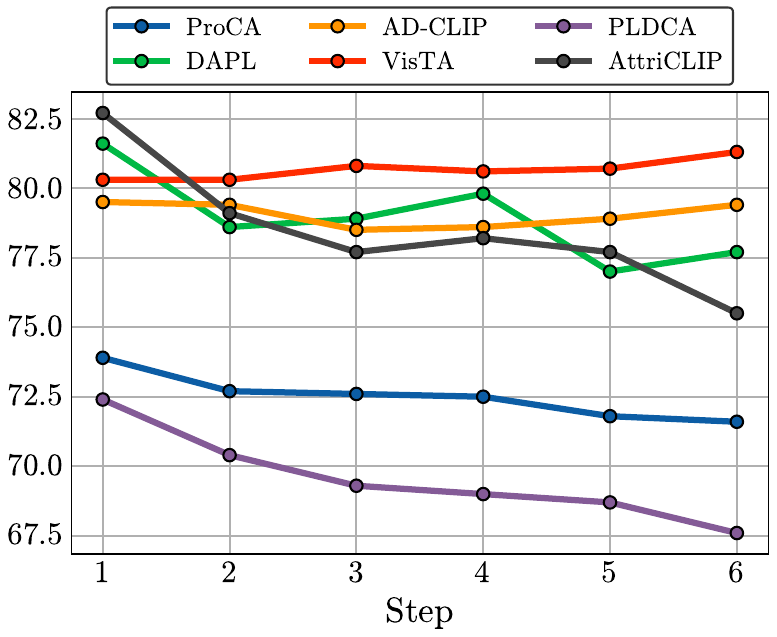}
  }
  \subfloat[$\Delta\%$~(Step-1)]{
      \includegraphics[width=0.23\textwidth]{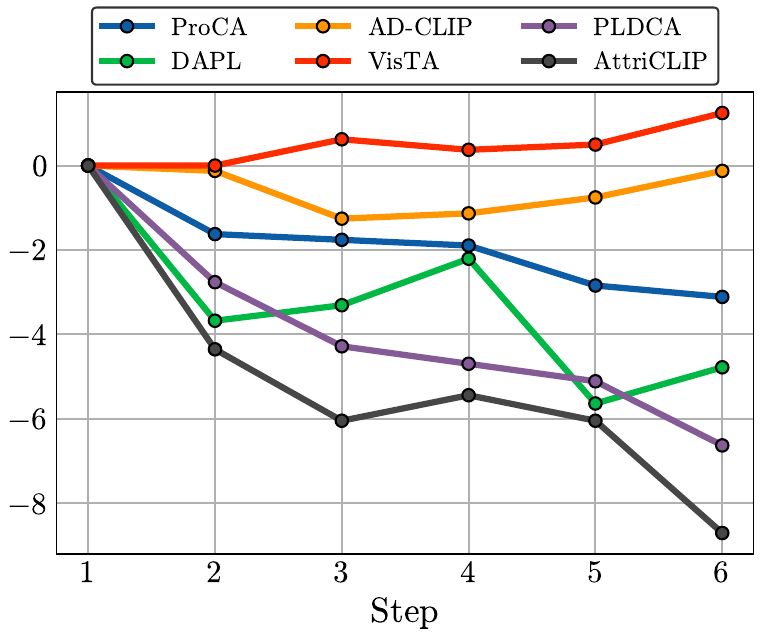}
  }
  \caption{S-1 Accuracy at each step and its percentage change ($\Delta \%$) compared with Step-1 on Office-Home.}
  \label{S-1}
  \Description{S-1 Accuracy at each step and its percentage change ($\Delta \%$) compared with Step-1 on Office-Home.}
\end{figure}

The Final Accuracy results are summarized in Tables \ref{Final-31}, \ref{Final-Home}, \ref{Final-Domain}, while the Step-level Accuracy results are detailed in Table \ref{Step-Home_Domain}.  
The numbers reported in the tables are reproduced by us using the officially released code, unless otherwise specified. 
Additionally, the results of S-1 Accuracy are visualized in Figure \ref{S-1}.
Due to page limitations, complete results with extended details for all metrics across three benchmarks are reported in \textbf{Appendix~\ref{sec:more_results}}.

\noindent\textbf{Comprehensive learning ability of VisTA.}
In Tables~\ref{Final-31}, \ref{Final-Home}, \ref{Final-Domain}, each column corresponds to one specific CI-UDA task, \emph{i.e.,} Source $\rightarrow$ Target. The column ``Avg.'' means the average of Final Accuracy for all CI-UDA tasks.

The results of Final Accuracy demonstrate that VisTA performs favorably against other methods.
VisTA achieves improvements of 0.7\%, 8.7\%, and 24.2\% over the leading CI-UDA method PLDCA on Office-31, Office-Home, and Mini-DomainNet, respectively. It also outperforms other top competitors, surpassing CoOp by 0.3\% on Office-31, AD-CLIP by 1.5\% on Office-Home, and DAPL by 1.7\% on Mini-DomainNet. We find that the performance advantages of VisTA scale with benchmark complexity (Office-31$\rightarrow$Office-Home$\rightarrow$Mini-DomainNet), as quantified by the number of time steps and underlying classes. 
It substantiates that the class-agnostic and domain-invariant attributes learned by VisTA effectively alleviate catastrophic forgetting and domain shift, especially in scenarios with numerous classes and long-sequence tasks.

The results further indicate that existing CI-UDA methods with ViT-B/16 underperform against source-only CoOp (on all benchmarks) and zero-shot CLIP (except on Office-31). This highlights the exceptional generalization ability of pretrained CLIP, which encodes comprehensive category knowledge via prompt learning.

Additionally, some UDA methods (\emph{i.e.,} DAMP, PGA) and CIL method (\emph{i.e.,} AttriCLIP) based on prompt learning obtain worse results than source-only CoOp across all three benchmarks. This suggests that an exclusive focus on mitigating either domain shift or catastrophic forgetting may reduce the generalization capability of prompt learning methods in handling CI-UDA.

\begin{figure*}[t]
  \centering
  \subfloat[\textbf{Sensitivity to loss weight $\lambda_1$ and $\lambda_2$.}]{
    \begin{minipage}{0.48\textwidth}
      \centering
      \includegraphics[width=0.48\textwidth]{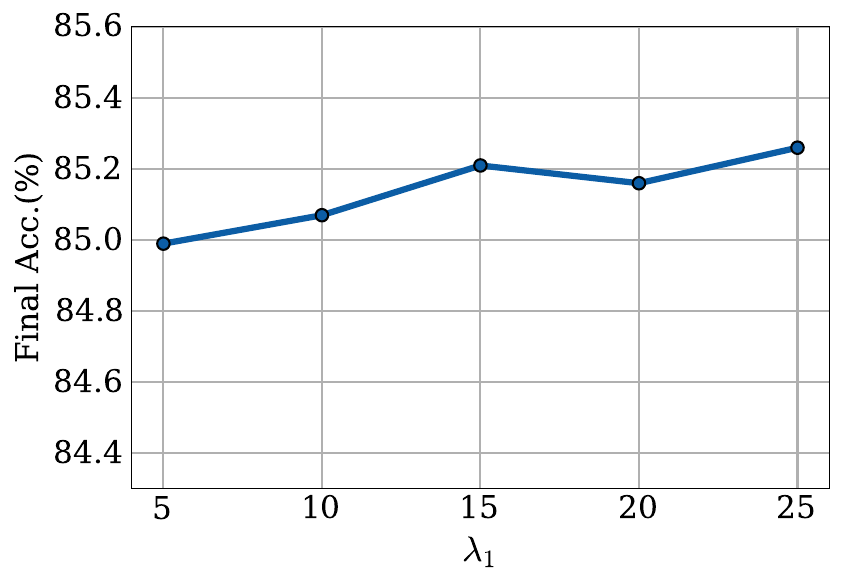}
      \includegraphics[width=0.48\textwidth]{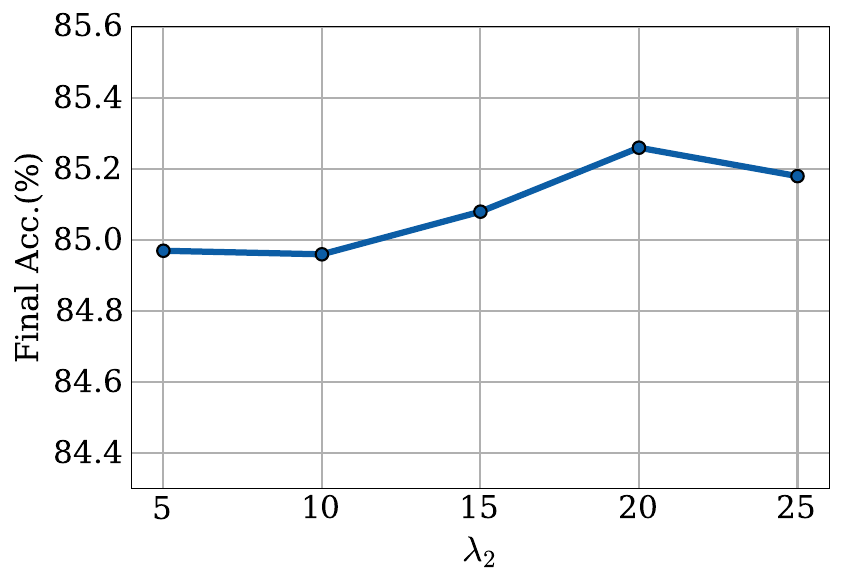}
    \end{minipage}
  }
  \subfloat[\textbf{Sensitivity to hyperparameters $M$ and $L$ of attribute dictionary.}]{
    \begin{minipage}{0.48\textwidth}
      \centering
      \includegraphics[width=0.48\textwidth]{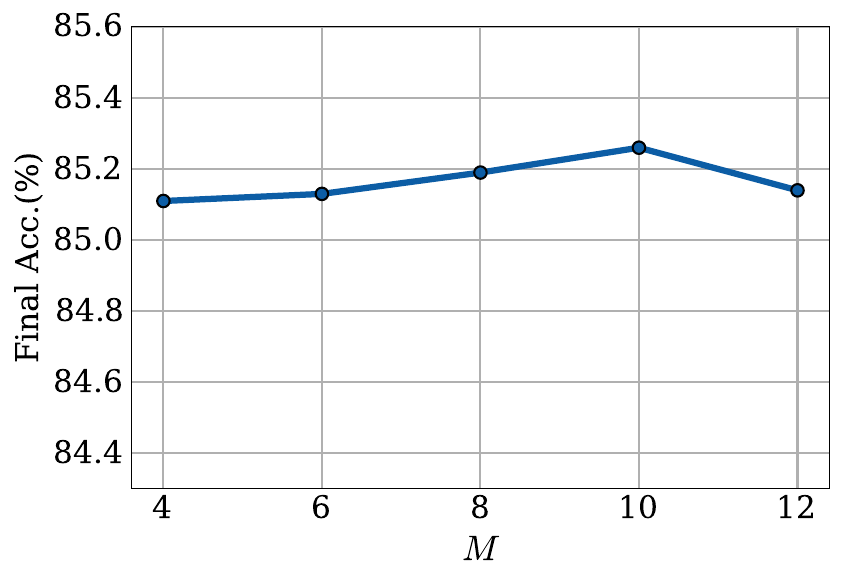}
      \includegraphics[width=0.465\textwidth]{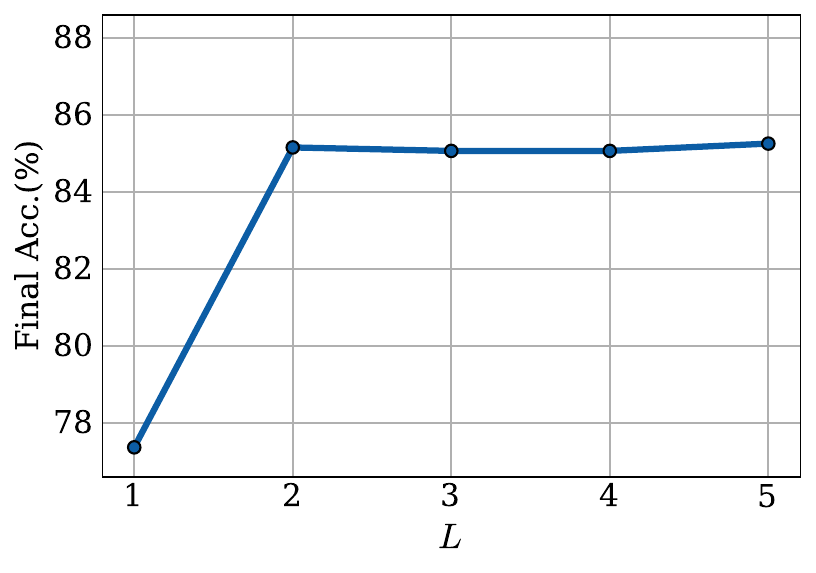}
    \end{minipage}
  }
  \caption{Hyperparameter sensitivity analysis with respect to Final Accuracy on Office-Home.}
  \label{sens}
  \Description{Hyperparameter sensitivity analysis with respect to Final Accuracy on Office-Home.}
\end{figure*}

\noindent\textbf{Incremental learning ability of VisTA.}
The Step-level Accuracy and S-1 Accuracy are used to evaluate whether a method can retain knowledge of previous target classes while learning new ones. 
Each column in Table~\ref{Step-Home_Domain} represents the average accuracy of all adaptation tasks at a specific time step, and the column ``Avg.'' denotes the average Step-level Accuracy. 
The x-axis indicates time steps and the y-axis shows the S-1 Accuracy in Figure~\ref{S-1}(a).

It is observed that all comparison methods are affected by catastrophic forgetting, resulting in lower S-1 Accuracy at the final step compared with the first step. Some methods also exhibit a decline in Step-level Accuracy over time, indicating that they not only forget previously learned target classes but also struggle to learn new ones. In contrast, VisTA shows an upward trend in both Step-level Accuracy and S-1 Accuracy over time, achieving the best performance at both the final step and on average. The results of both metrics also show that the performance in the early adaptation phase is not state-of-the-art. We analyze that although VisTA aligns attributes across domains as much as possible, the learned $\mathcal{A}^t$ is not sufficiently refined in the early stage, while $\mathcal{A}^s$ is trained on all classes. Aligning $\mathcal{A}^s$ and $\mathcal{A}^t$ with unequal learning progress may harm performance. Only after $\mathcal{A}^t$ fully learns the attributes over time steps can the performance advantages manifest.

Moreover, Figure~\ref{S-1} (b) illustrates the percentage change in S-1 Accuracy at each step compared with the first step. VisTA is the only method that consistently achieves positive gains and demonstrates continuous improvement. This indicates that VisTA effectively preserves and progressively reinforces knowledge of class-incremental $\mathcal{D}_t$ when addressing CI-UDA on Office-Home.

\noindent\textbf{Extension to source-free scenario.}
In \textbf{Appendix~\ref{sec:sf}}, we compare VisTA with the CI-SFUDA method GROTO~\cite{CISFUDA}. 
These results demonstrate the capabilities of VisTA under source-free scenarios.

\begin{table}[!t]
	\caption{Ablation analysis on Office-Home.}
	\centering
	\label{ablation}
	\scalebox{0.8}{
		\begin{tabular}{c|cccc>{\columncolor[HTML]{d1cfcf}}c}
			\toprule
			Method &  w/o. VAC & w/o. $\mathcal{L}_{\mathrm{con}}$ & w/o. $\mathcal{L}_{\mathrm{hp}}$ & w/o. $\mathcal{L}_{\mathrm{div}}$ & VisTA\\ \midrule
			Final Acc. (\%) & 84.9  & 83.5  & 85.0 & 85.1 & \textbf{85.3} \\
			Final S-1 Acc. (\%) & 80.5 & 80.3 & 80.9 & 81.0 &\textbf{81.3}\\ \bottomrule
	\end{tabular}
	}
\end{table}

\subsection{Ablation Analysis}
Table~\ref{ablation} presents Final Accuracy and S-1 Accuracy at the final step (\emph{i.e.,} Final S-1 Accuracy) on Office-Home, obtained by removing specific modules (\emph{i.e.,} ``w/o.'') while keeping other settings identical. 
More studies about various VLMs and computational overhead are reported in \textbf{Appendix~\ref{sec:ablation}}.

\noindent\textbf{Effect of VAC.} 
The ``w/o. VAC'' is achieved by removing the VAC module, with attributes selected directly from the dictionary corresponding to the other domain using cosine similarity. This leads to a noticeable performance degradation, demonstrating that the selection enabled by the VAC module effectively mitigates bias caused by the domain shift.

\noindent\textbf{\label{ablat}Effect of $\mathcal{L}_{\mathrm{con}}$.} The ``w/o. $\mathcal{L}_{\mathrm{con}}$'' indicates that $\mathcal{D}_s$ and $\mathcal{D}_t$ share the same attributes dictionary without $\mathcal{L}_{\mathrm{con}}$. 
Notably, the ``w/o. $\mathcal{L}_{\mathrm{con}}$'' leads to the sharpest performance decline, demonstrating that separate attribute modeling in $\mathcal{D}_s, \mathcal{D}_t$ with alignment effectively prevents conflicting knowledge acquisition in entangled attributes.

\noindent\textbf{Effect of regularization terms.} The ``w/o. $\mathcal{L}_{\mathrm{hp}}$,'' and ``w/o. $\mathcal{L}_{\mathrm{div}}$'' denote the $\mathcal{L}_{\mathrm{hp}}$ and $\mathcal{L}_{\mathrm{div}}$ are removed from the objective (\ref{final}), respectively. The observed performance decline confirms the effectiveness of both regularization terms in attribute learning. 

\begin{figure}[t]
	\centering
	\includegraphics[width=0.967\linewidth]{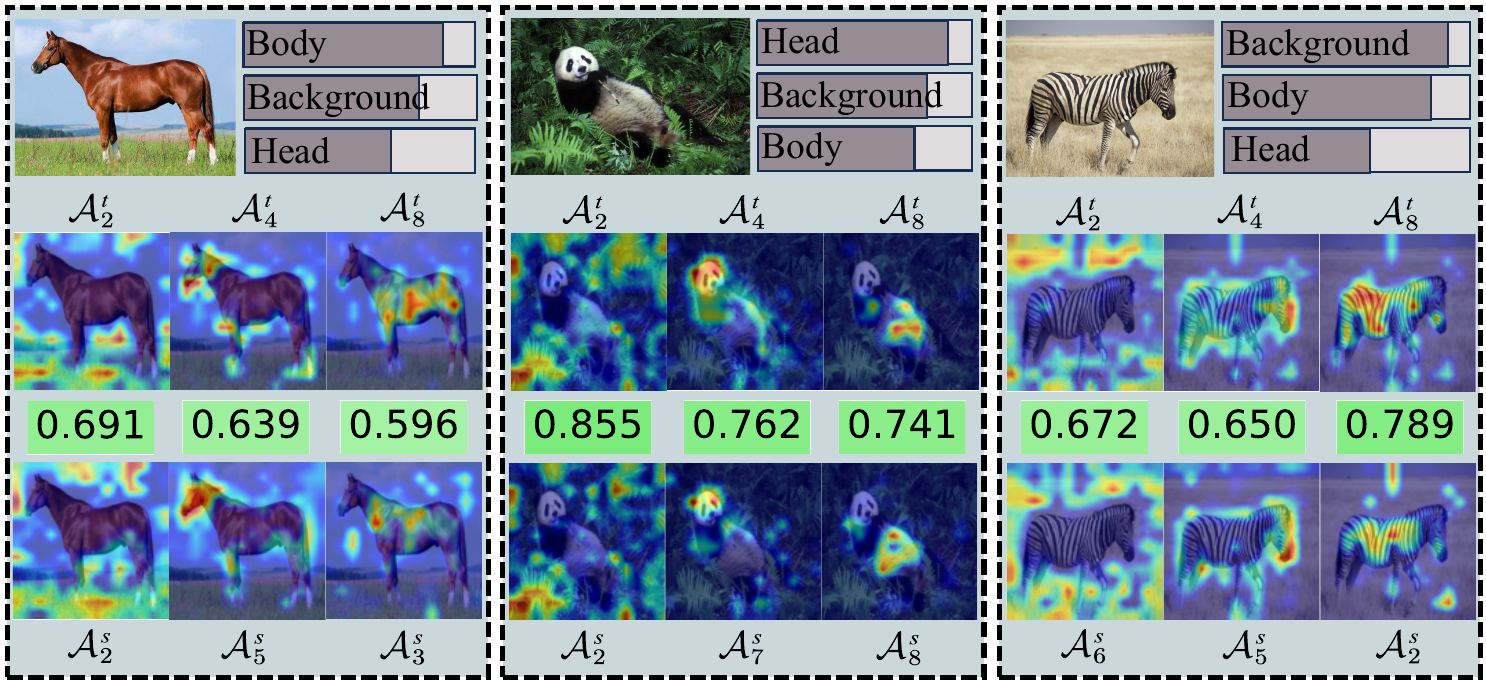}
	\caption{Grad-CAM visualization of C $\rightarrow$ R task (Mini-DomainNet) for different classes in $\mathcal{D}_t$. Attributes selected directly from $[\mathcal{K},\mathcal{A}]^t$ are displayed in a ranked order beside the first-row images. The second and third rows show some attention heatmaps for $\mathcal{A}^t$ and $\mathcal{A}^s$, respectively. Heatmaps with the same semantic concept are grouped in columns, and their matching is guided by $\rho$ (highlighted in green).}
	\label{attn}
\Description{Grad-CAM visualization of C $\rightarrow$ R task (Mini-DomainNet) for different classes in $\mathcal{D}_t$. Attributes selected directly from $[\mathcal{K},\mathcal{A}]^t$ are displayed in a ranked order beside the first-row images. The second and third rows show some attention heatmaps for $\mathcal{A}^t$ and $\mathcal{A}^s$, respectively. Heatmaps with the same semantic concept are grouped in columns, and their matching is guided by $\rho$ (highlighted in green).}
\end{figure}

\noindent\textbf{\label{hyp}Sensitivity to loss weights.} 
We need to determine three loss weights of the objective~(\ref{final}). 
Empirical observations reveal $\lambda_3$ has negligible impact, so we fix it at 1.0 and vary $\lambda_1$ and $\lambda_2$.
As shown in Figure~\ref{sens}(a), 
the performance of VisTA is generally insensitive to $\lambda_1, \lambda_2\in[5, 10, 15, 20, 25]$, with best performance at $\lambda_1=25, \lambda_2=20$.

\noindent\textbf{Sensitivity to hyperparameters of attribute dictionary.} 
We consider the following hyperparameters, such as the prompt length $M$, the number of attributes in the dictionary $N$, and the number of selected attributes $L$.
To cap training and computational costs, we fix $N = 8$ and explore variations in $M$ and $L$. Figure~\ref{sens}(b) shows that the performance of VisTA is robust to $M$. Moreover, when a sufficient number of attributes are selected ($L \geq 2$), VisTA also exhibits robustness to $L$.

\noindent\textbf{Visualization of textual attributes.} 
To verify whether the learned attributes reflect the semantic concept of images, we visualize the image contents of distinct classes corresponding to different attributes using Grad-CAM~\cite{selvaraju2017grad}. 
To further demonstrate the attribute matching process in VAC module, we present examples of target classes ``horse,'' ``panda,'' and ``zebra'' from Mini-DomainNet.

As shown in Figure~\ref{attn}, the learned $\mathcal{A}^t$ exhibit two key properties: (1) different attributes reflect distinct semantic concepts within the same image (\emph{e.g.,} $\mathcal{A}^t_2 \rightarrow$ ``Background,'' $\mathcal{A}^t_4 \rightarrow$ ``Head,'' $\mathcal{A}^t_8 \rightarrow$ ``Body''), and (2) the same attribute reflects identical semantic concept across different images. This demonstrates that the learned $\mathcal{A}^t$ are class-agnostic and diverse, effectively retaining knowledge to alleviate catastrophic forgetting. Unlike the learned $\mathcal{A}^t$ corresponding to $\mathcal{D}_t$, the $\mathcal{A}^s$, affected by the distribution shift, fail to learn attributes identical to $\mathcal{A}^t$ and do not exhibit property (2). However, $\mathcal{A}^s$ may still be partially similar to $\mathcal{A}^t$ in semantic concepts. Building on this similarity, VAC module selects cross-domain attributes through a $\rho$-guided matching mechanism to learn domain-invariant attributes that mitigate the distribution shift.

\begin{figure}[t]
  \centering
  \subfloat[Step 1]{
      \includegraphics[width=0.15\textwidth]{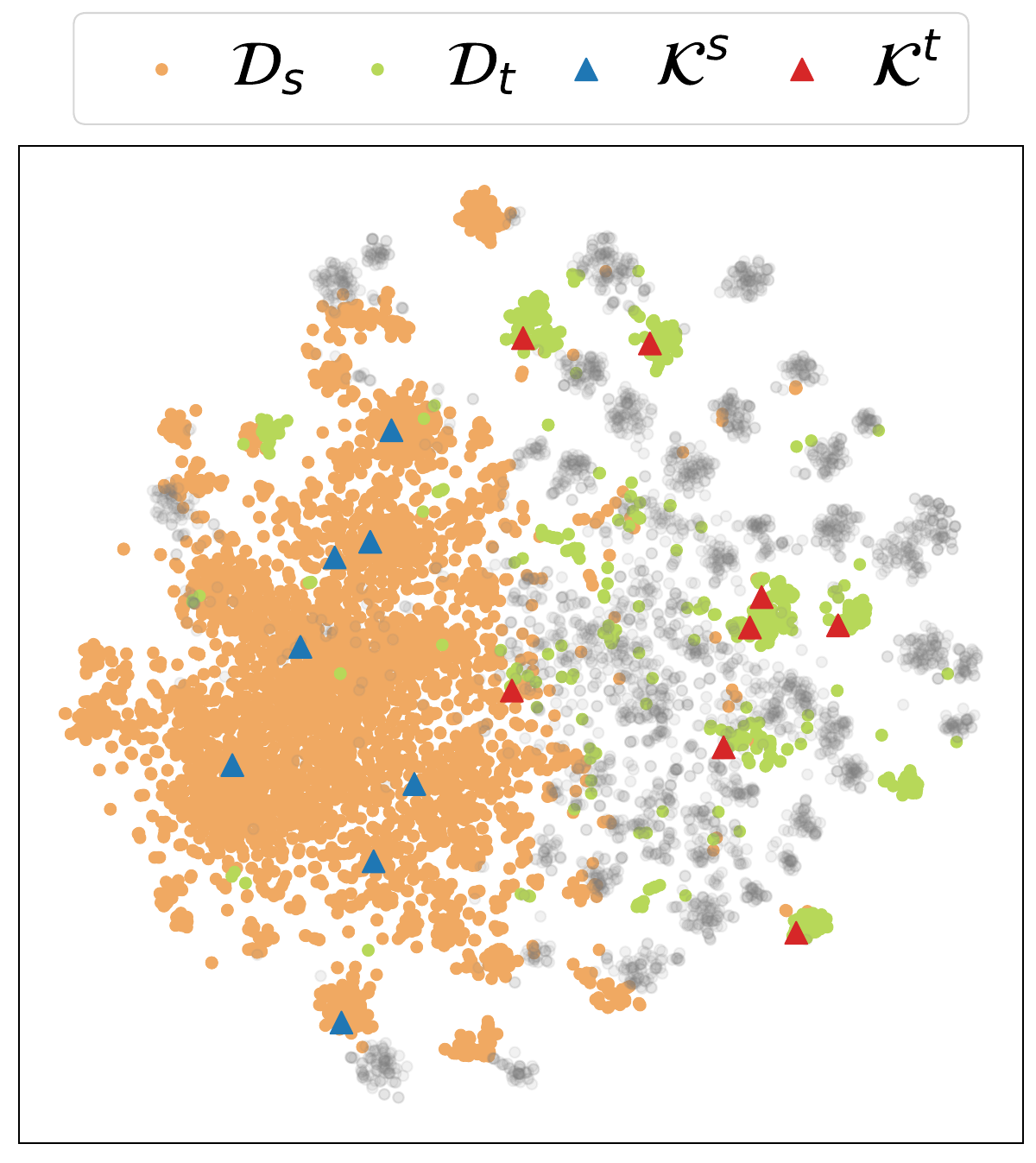}
  }
  \subfloat[Step 4]{
      \includegraphics[width=0.15\textwidth]{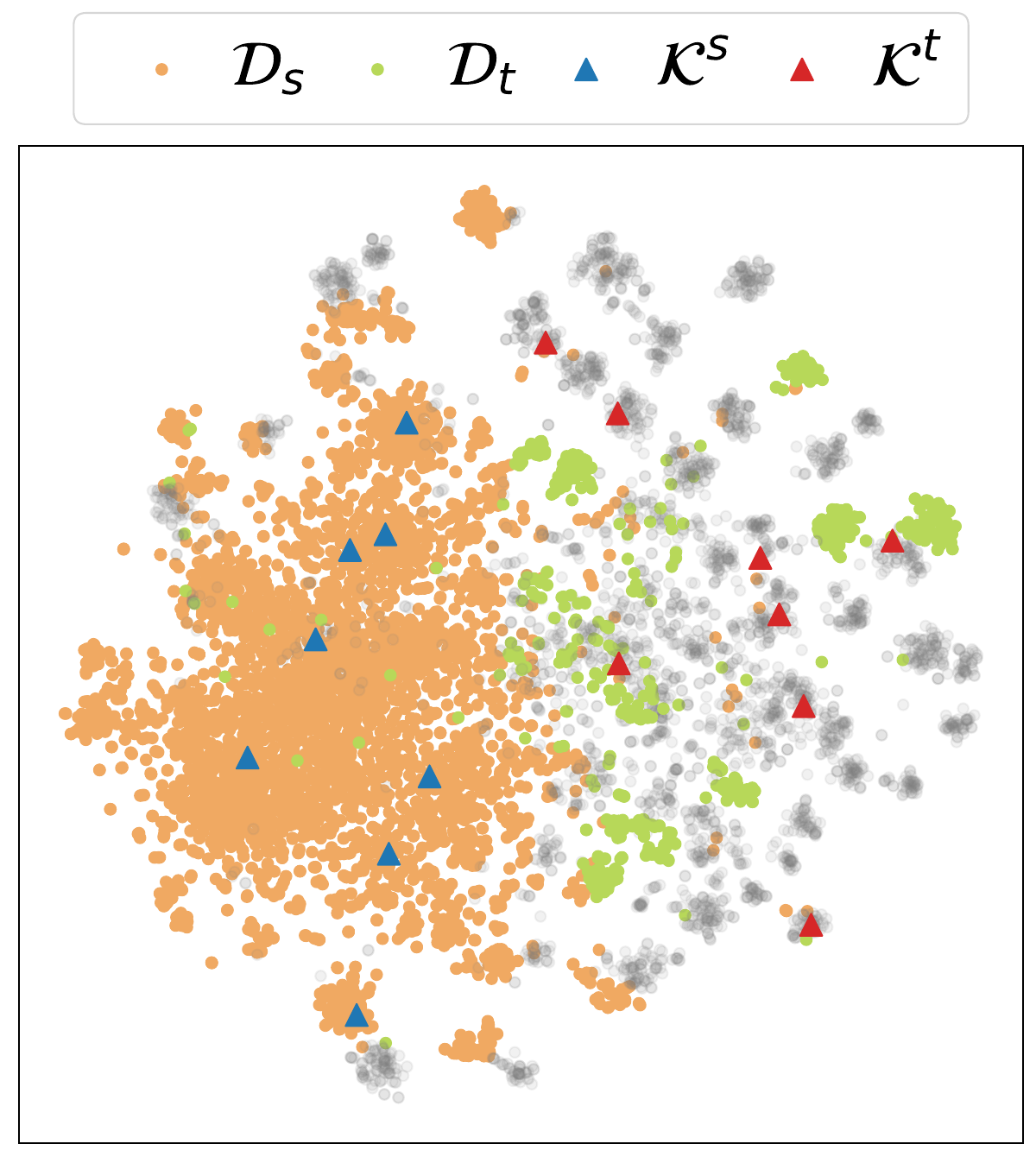}
  }
  \subfloat[Step 6]{
      \includegraphics[width=0.15\textwidth]{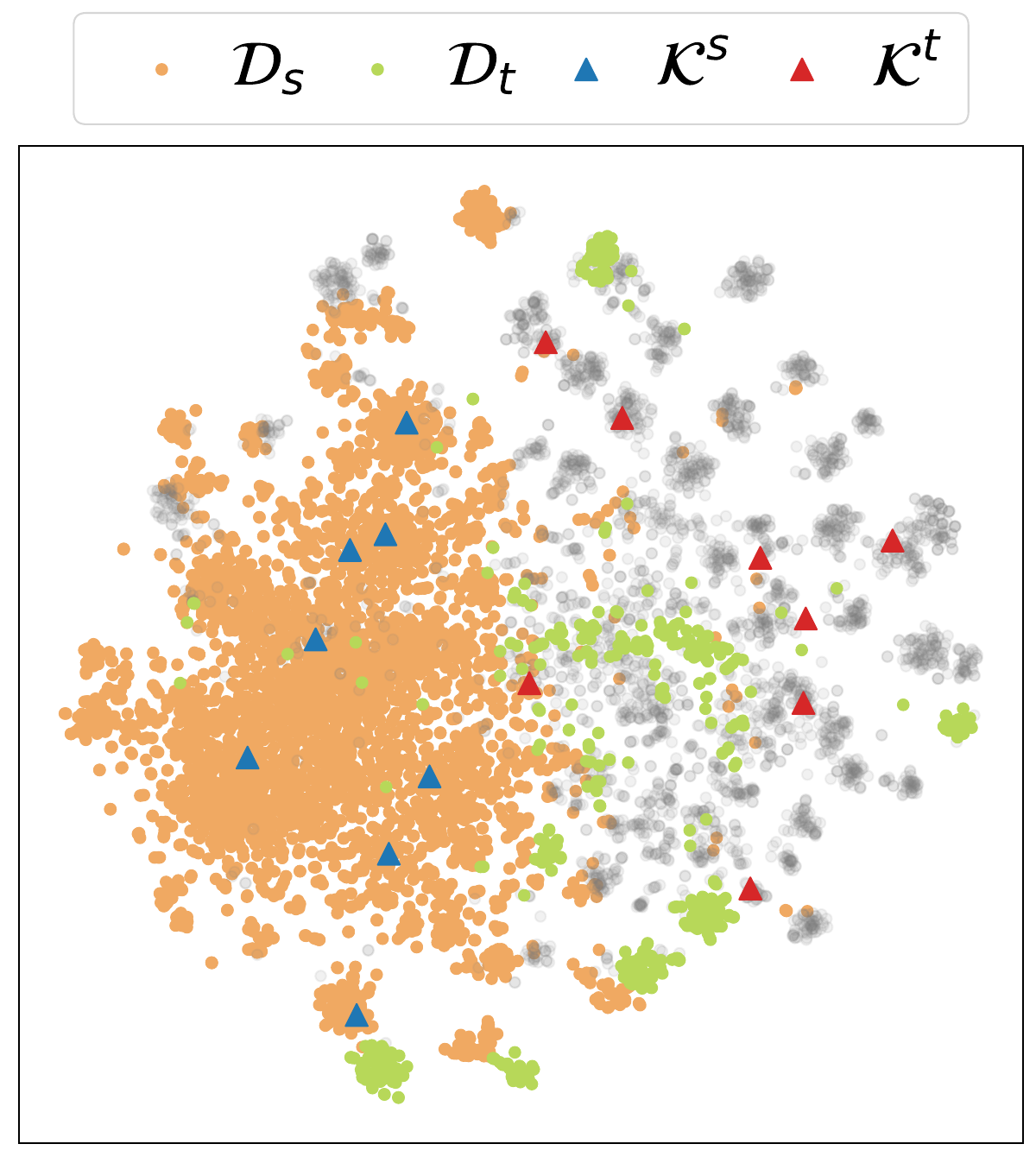}
  }
  \caption{t-SNE visualization of C $\rightarrow$ P task from Office-Home at different time steps.}
  \label{tsne}
  \vspace{-0.6mm}
  \Description{t-SNE visualization of C $\rightarrow$ P task from Office-Home at different time steps.}
\end{figure}

\noindent\textbf{Visualization of visual attributes.} To verify whether $\mathcal{K}^s$ and $\mathcal{K}^t$ can adequately cover the attributes of all examples, we conducted t-SNE visualizations for C $\rightarrow$ P task from Office-Home at steps 1, 4, and 6 as shown in Figure~\ref{tsne}. It displays CLIP-extracted visual features from $\mathcal{D}_s$ (orange) and $\mathcal{D}_t$ (green), along with $\mathcal{K}$ obtained through k-means++ clustering. We observe that the elements within $\mathcal{K}^s$ (blue) and $\mathcal{K}^t$ (red) consistently remain diverse and distinct. Furthermore, $\mathcal{K}^t$ at the final step successfully covers the examples from other time steps (gray), demonstrating that $\mathcal{K}$ effectively captures the overall attributes of the sample from both domains.

\section*{Conclusion}
In this paper, we propose to model and align attributes across domains based on CLIP to deal with the class-incremental unsupervised domain adaptation (CI-UDA), which is a rehearsal-free approach. Specifically, via CLIP, we extract the class-agnostic properties, \emph{i.e.,} attributes. 
Each attribute is represented as a ``key-value'' pair where the key corresponds to visual prototype and the value corresponds to textual prompt.
In our method, we learn to construct two dictionaries, each corresponding to a specific domain. Each dictionary consists of a group of attributes. 
Then we perform attribute alignment to make attribute invariant across domains via utilizing the consistency knowledge including visual attention consistency and prediction consistency. Experiments on three benchmarks verify the effectiveness of our proposed method.

\clearpage
\balance
\begin{acks}
This research is supported by the National Natural Science Foundation of China (NSFC) under Grants Nos. 62336003, 12371510, 92370114, and 62006119; and the National Key Research and Development Program of China (International Collaboration Special Project, No. SQ2023YFE0102775).
\end{acks}

\bibliographystyle{ACM-Reference-Format}
\bibliography{sample-base}

\appendix
\nobalance
\renewcommand{\thefigure}{\Alph{section}-\arabic{figure}}
\renewcommand{\thetable}{\thesection-\arabic{table}}
\makeatletter
\@addtoreset{figure}{section}
\@addtoreset{table}{section}
\makeatother
\section*{Supplementary Materials}
\section{Details of Data Construction}
\label{sec:appendix_Data}
In this section, we show the containing classes in each disjoint subset of all three benchmarks (\emph{i.e.}, Office-31~\cite{office31}, Office-Home~\cite{Venkateswara_2017_CVPR}, and Mini-DomainNet~\cite{Peng_2019_ICCV}) in Table~\ref{tab:data_construction}. For Office-31 and Mini-DomainNet, classes are sorted alphabetically; Office-31 is partitioned into groups of 10 categories per step, while Mini-DomainNet is divided into groups of 20 categories per step. For Office-Home, categories are randomly grouped into steps of 10 classes each.

\begin{table*}[t]
\renewcommand\arraystretch{0.95}
    \begin{center}
    \caption{\label{tab:data_construction}Class names in each time step on Office-31, Office-Home, and Mini-DomainNet.
    }
    \scalebox{0.75}{
         \begin{tabular}{c|cll}
        \toprule
         Dataset & Time Step & Class Index & Class Name \\
         \midrule
         \multirow{6}{*}{Office-31~\cite{office31}} & \multirow{2}{*}{Step 1} & \multirow{2}{*}{[$0\sim9$]} & back pack, bike, bike helmet, bookcase, bottle, calculator, \\ 
         & & & desk chair, desk lamp, desktop computer, file cabinet \\
         \cmidrule(lr){2-4}
         & \multirow{2}{*}{Step 2} & \multirow{2}{*}{[$10\sim19$]} & headphones, keyboard, laptop computer, letter tray, \\ 
         & & & mobile phone, monitor, mouse, mug, paper notebook, pen\\
         \cmidrule(lr){2-4}
         & \multirow{2}{*}{Step 3} & \multirow{2}{*}{[$20\sim29$]} & phone, printer, projector, punchers, ring binder, ruler, \\ 
         & & &  scissors, speaker, stapler, tape dispenser\\
         
         \midrule
         \multirow{12}{*}{Office-Home~\cite{Venkateswara_2017_CVPR}} & \multirow{2}{*}{Step 1} & \multirow{2}{*}{[$0\sim9$]} & Drill, Exit Sign, Bottle, Glasses, Computer, \\ 
         & & &  File Cabinet, Shelf, Toys, Sink, Laptop \\
         \cmidrule(lr){2-4}
         & \multirow{2}{*}{Step 2} & \multirow{2}{*}{[$10\sim19$]} & Kettle, Folder, Keyboard, Flipflops, Pencil, \\ 
         & & & Bed, Hammer, ToothBrush, Couch, Bike\\
         \cmidrule(lr){2-4}
         & \multirow{2}{*}{Step 3} & \multirow{2}{*}{[$20\sim29$]} & Postit Notes, Mug, Webcam, Desk Lamp, Telephone,\\ 
         & & & Helmet, Mouse, Pen, Monitor, Mop \\
          \cmidrule(lr){2-4}
         & \multirow{2}{*}{Step 4} & \multirow{2}{*}{[$30\sim39$]} & Sneakers, Notebook, Backpack, Alarm Clock, Push Pin,  \\ 
         & & & Paper Clip, Batteries, Radio, Fan, Ruler \\
         \cmidrule(lr){2-4}
         & \multirow{2}{*}{Step 5} & \multirow{2}{*}{[$40\sim49$]} & Pan, Screwdriver, Trash Can, Printer, Speaker, \\ 
         & & & Eraser, Bucket, Chair, Calendar, Calculator \\
         \cmidrule(lr){2-4}
         & \multirow{2}{*}{Step 6} & \multirow{2}{*}{[$50\sim59$]} & Flowers, Lamp Shade, Spoon, Candles, Clipboards \\ 
         & & & Scissors, TV, Curtains, Fork, Soda \\
        \bottomrule
         \multirow{18}{*}{Mini-DomainNet~\cite{Peng_2019_ICCV}} & \multirow{3}{*}{Step 1} & \multirow{3}{*}{[$0\sim19$]} & aircraft carrier, alarm clock, ant, anvil, asparagus, axe, banana,\\ 
         & & & basket, bathtub, bear, bee, bird, blackberry, blueberry, \\
         & & & bottlecap, broccoli, bus, butterfly, cactus, cake \\
         \cmidrule(lr){2-4}
         & \multirow{3}{*}{Step 2} & \multirow{3}{*}{[$20\sim39$]} & calculator, camel, camera, candle, cannon, canoe, carrot, \\ 
         & & & castle, cat, ceiling fan, cello, cell phone, chair, chandelier, \\
         & & & coffee cup, compass, computer, cow, crab, crocodile \\
         \cmidrule(lr){2-4}
         & \multirow{3}{*}{Step 3} & \multirow{3}{*}{[$40\sim59$]} & cruise\_ship, dog, dolphin, dragon, drums, duck, dumbbell, \\ 
         & & & elephant, eyeglasses, feather, fence, fish, flamingo, \\
         & & & flower, foot, fork, frog, giraffe, goatee, grapes \\
         \cmidrule(lr){2-4}
         & \multirow{3}{*}{Step 4} & \multirow{3}{*}{[$60\sim79$]} &  guitar, hammer, helicopter, helmet, horse, kangaroo, lantern,\\ 
         & & &   laptop, leaf, lion, lipstick, lobster, microphone, monkey,\\
         & & &  mosquito, mouse, mug, mushroom, onion, panda\\
         \cmidrule(lr){2-4}
         & \multirow{3}{*}{Step 5} & \multirow{3}{*}{[$80\sim99$]} &  peanut, pear, peas, pencil, penguin, pig, pillow, pineapple,\\ 
         & & &  potato, power outlet, purse, rabbit, raccoon, rhinoceros,\\
         & & &  rifle, saxophone, screwdriver, sea turtle, see saw, sheep\\
         \cmidrule(lr){2-4}
         & \multirow{3}{*}{Step 6} & \multirow{3}{*}{[$100\sim119$]} &  shoe, skateboard, snake, speedboat, spider, squirrel, strawberry,\\ 
         & & &  streetlight, string bean, submarine, swan, table, teapot, teddy-bear,\\
         & & &  television, The Eiffel Tower, The Great Wall of China, tiger, toe, train\\
        \bottomrule         
        \end{tabular}
         }  
    \end{center}
\end{table*}

\section{Details of Training Procedures and Hyperparameters}
\label{sec:hyper}

We use ViT-B/16 as the image encoder for VisTA. The model is trained for 10 epochs per incremental step across all datasets. During training, we employ the mini-batch stochastic gradient descent (SGD) optimizer with an initial learning rate of $3\times 10^{-3}$, adjusting it using a cosine decay schedule.  Moreover, the penultimate layer~\cite{zhao2024gradient} is used for gradient computation in the VAC module, as VisTA employs the ViT-B/16 backbone. We also adopt ViT-B/16 as the image encoder for all baseline methods, including replacing the original backbone (ResNet50~\cite{he2016deep}) of ProCA~\cite{lin2022prototype} and PLDCA~\cite{wei2024class}. For AttriCLIP, we learn a shared set of prompts for $\mathcal{D}_s$ and $\mathcal{D}_t$ using Equations~(\ref{ce}) and (\ref{sl}), and also enhance the quality of pseudo-labeling in $\mathcal{D}_t$ via DebiasPL.

The threshold $\gamma$ in Equation~(\ref{sl}) is set to 0.7 for Office-31 and Mini-DomainNet, and 0.6 for Office-Home. For the loss weights, we set $\lambda_1 = 25, \lambda_2 = 20, \lambda_3 = 1$ for Office-Home and Mini-DomainNet, and $\lambda_1=3, \lambda_2=6, \lambda_3=1$ for Office-31. Additionally, for Office-31 we set $M = 5, N = 13, L = 13$; for Office-Home, $M = 10, N = 8, L = 5$, and for Mini-DomainNet, $M = 10, N = 8, L = 6$.

\section{More Results of Class-Incremental Unsupervised Domain Adaptation}
\label{sec:more_results}
\begin{figure}[t]
  \centering
  \subfloat[S-1 Acc. (\%)]{
      \includegraphics[width=0.226\textwidth]{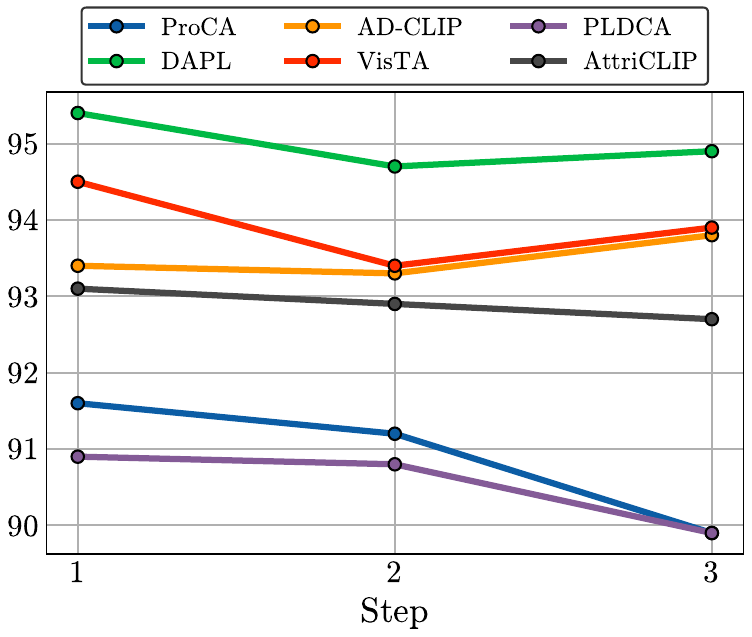}
  }
  \subfloat[$\Delta\%$~(Step 1)]{
      \includegraphics[width=0.236\textwidth]{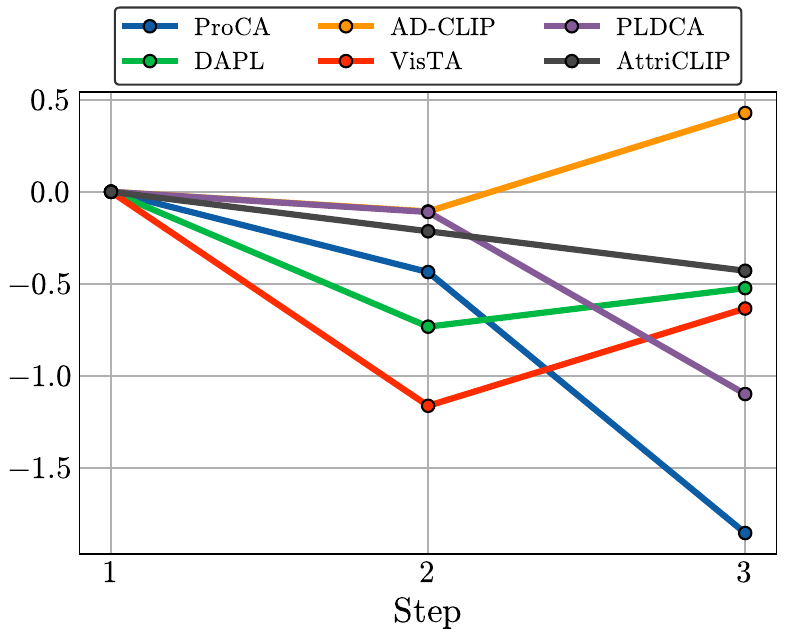}
  }
  \caption{S-1 Accuracy at each step and its percentage change ($\Delta \%$) compared with Step-1 on Office-31.}
  \label{31_S-1}
  \Description{S-1 Accuracy at each step and its percentage change ($\Delta \%$) compared with Step-1 on Office-31.}
\end{figure}

\begin{figure}[t]
  \centering
  \subfloat[S-1 Acc. (\%)]{
      \includegraphics[width=0.226\textwidth]{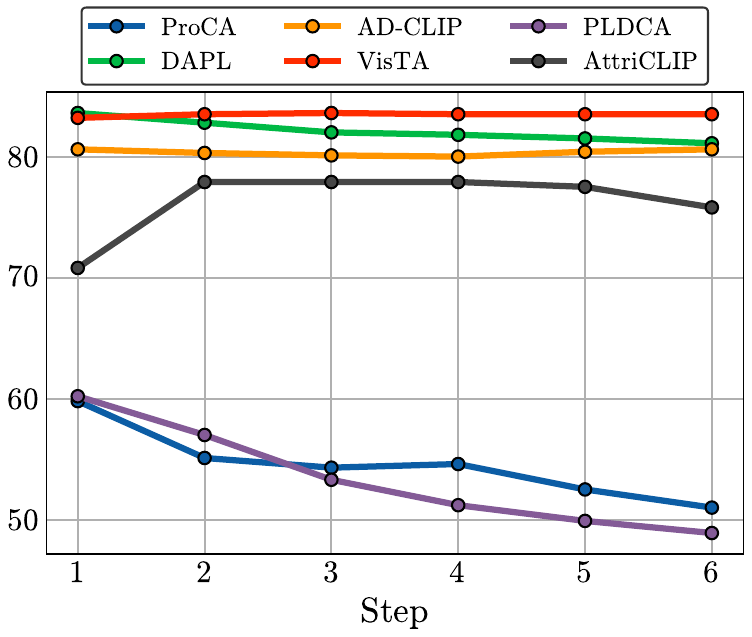}
  }
  \subfloat[$\Delta\%$~(Step 1)]{
      \includegraphics[width=0.234\textwidth]{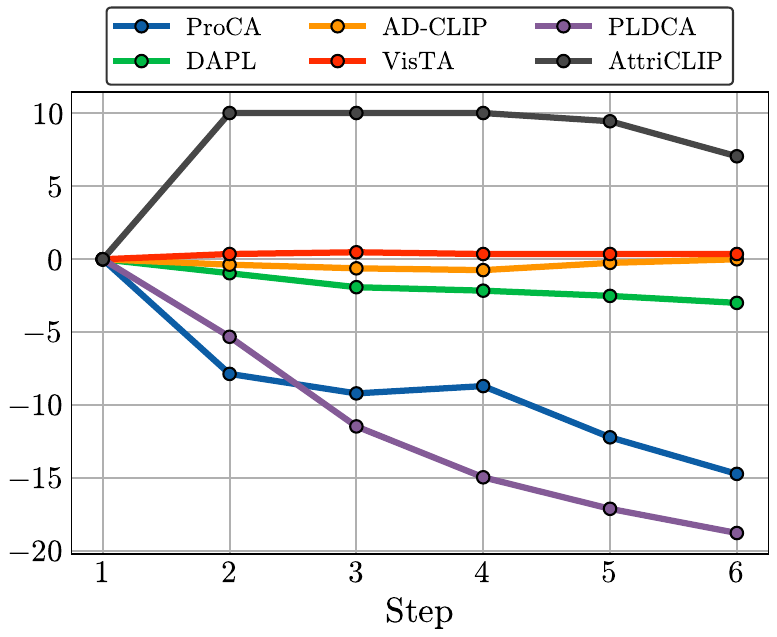}
  }
  \caption{S-1 Accuracy at each step and its percentage change ($\Delta \%$) compared with Step-1 on Mini-DomainNet.}
  \label{Domain_S-1}
  \Description{S-1 Accuracy at each step and its percentage change ($\Delta \%$) compared with Step-1 on Mini-DomainNet.}
\end{figure}

In this section, we report a comprehensive comparison of our VisTA against state-of-the-art baseline methods (e.g., ProCA~\cite{lin2022prototype}, PLDCA~\cite{wei2024class}, CoOp~\cite{zhou2022coop}, AttriCLIP~\cite{Singha_2023_ICCV}, AD-CLIP~\cite{Singha_2023_ICCV}, DAMP~\cite{Du_2024_CVPR}, PGA~\cite{phan2024enhancing}, DAPL~\cite{10313995}), except source-only ViT-B/16~\cite{dosovitskiy2021an} and zero-shot CLIP\cite{pmlr-v139-radford21a}, whose results are directly cited from existing paper. Notably, our evaluation adopts more complete and detailed metrics than those in the main paper:

\begin{itemize}
	\item The Step-level Accuracy on Office-31, Office-Home, and Mini-DomainNet (Tables~\ref{Step-31}, \ref{Step-Home}, and \ref{Step-Domain});
	\item The average S-1 Accuracy of all adaptation tasks at each time step and its percentage change compared with Step-1 on Office-31 and Mini-DomainNet (Figures~\ref{31_S-1} and \ref{Domain_S-1});
	\item The Step-level Accuracy and S-1 Accuracy for \textbf{all adaptation tasks} across Office-31 (Table~\ref{tab:Office31}), Office-Home (Tables~\ref{tab:OH_Ar}, \ref{tab:OH_Cl}, \ref{tab:OH_Pr}, and \ref{tab:OH_Rw}), and Mini-DomainNet (Tables~\ref{tab:MD_Cl}, \ref{tab:MD_Pn}, \ref{tab:MD_Rl}, and \ref{tab:MD_Sk}).
\end{itemize}

\begin{table}[t]
	\centering
	\caption{Step-level Accuracy (\%) on Office-31. DA, CI, and RF respectively represent domain adaptation, class-incremental, and rehearsal-free.}
	\scalebox{0.75}{  
		\begin{tabular}{lccc|cccc}
			\toprule
			Method & DA & CI & RF & Step 1 & Step 2 & Step 3 & Avg.\\
			\midrule
			ProCA~(ECCV22)& \cmark & \cmark & \xmark & 92.9 & 90.9 & 88.0 & 90.6  \\
			PLDCA~(TIP24)& \cmark & \cmark & \xmark & 91.5 & 90.9 & 88.7 & 90.3 \\
			CoOp~(IJCV22)& --  & -- & --  & 93.5 & 93.0 & 87.9 & 91.5 \\
			AttriCLIP~(CVPR23)& \xmark & \cmark & \cmark & 93.1 & 91.1 & 86.3 & 90.1\\ 
			AD-CLIP~(ICCVW23)& \cmark & \xmark & \cmark & 93.3 & 89.2 & 87.0 & 89.8 \\
			DAMP~(CVPR24)& \cmark & \xmark & \cmark & 93.7 & 89.9 & 84.6 & 89.4  \\
			PGA~(NeurIPS24)& \cmark & \xmark & \cmark & 92.2 & 79.4 &  75.1 & 82.2 \\
			DAPL~(TNNLS25)& \cmark & \xmark & \cmark & \textbf{95.4} & 86.6 & 79.2 & 87.0\\
			\midrule
			VisTA~(Ours)& \cmark & \cmark & \cmark & 94.3 & \textbf{93.5} & \textbf{89.3} & \textbf{92.4} \\
			\bottomrule
		\end{tabular}
 	}
	\label{Step-31}
\end{table}

\begin{table*}[t]
	\centering
	\caption{Step-level Accuracy (\%) on Office-Home. DA, CI, and RF respectively represent domain adaptation, class-incremental, and rehearsal-free.}
	\scalebox{0.75}{  
		\begin{tabular}{lccc|ccccccc}
			\toprule
			Method & DA & CI & RF & Step 1 & Step 2 & Step 3 & Step 4 & Step 5 & Step 6 & Avg.\\
			\midrule
			ProCA~(ECCV22)& \cmark  & \cmark & \xmark & 73.2          & 74.0          & 72.3          & 73.3          & 73.8         & 74.1          & 73.5 \\
			PLDCA~(TIP24)& \cmark  & \cmark & \xmark & 70.8          & 74.4          & 73.7          & 74.8          & 76.1        & 76.6          & 74.4  \\
			CoOp~(IJCV22)& --  & -- & -- & 79.2          & 82.1          & 81.6          & 82.7          & 82.8        & 82.7          & 81.9 \\
			AttriCLIP~(CVPR23)& \xmark  & \cmark & \cmark & \textbf{82.7}          & 84.5         & 81.8          & 82.7          & 83.1        & 82.3          & 82.9 \\ 			
			AD-CLIP~(ICCVW23)& \cmark  & \xmark & \cmark & 79.5          & 83.5          & 81.8          & 83.5          & 83.3        & 83.8          & 82.6 \\
			DAMP~(CVPR23)& \cmark  & \xmark & \cmark & 81.0            & 83.3          & 82.1          & 82.4          & 82.0          & 81.9          & 82.1  \\
			PGA~(NeurIPS24)& \cmark  & \xmark & \cmark & 78.7   & 73.9   & 71.1   & 71.4   & 71.0     & 70.5   & 72.8 \\
			DAPL~(TNNLS25)& \cmark  & \xmark & \cmark & 81.6          & 83.7          & 81.9          & 82.9          & 83.4        & 83.4          & 82.8 \\
			\midrule
			VisTA~(Ours)& \cmark  & \cmark & \cmark  & 80.3   & \textbf{84.7}   & \textbf{84.2} & \textbf{85.1} & \textbf{85.1} & \textbf{85.3} & \textbf{84.1} \\
			\bottomrule
		\end{tabular}
	}
	\label{Step-Home}
\end{table*}

\begin{table*}[t]
	\centering
	\caption{Step-level Accuracy (\%) on Mini-DomainNet. DA, CI, and RF respectively represent domain adaptation, class-incremental, and rehearsal-free.}
	\scalebox{0.75}{  
		\begin{tabular}{lccc|ccccccc}
			\toprule
			Method & DA & CI & RF & Step 1 & Step 2 & Step 3 & Step 4 & Step 5 & Step 6 & Avg.\\
			\midrule
			ProCA~(ECCV22)& \cmark & \cmark & \xmark  & 59.8 & 55.3 & 57.8 & 57.1 & 55.1 & 56.3 & 56.9 \\
			PLDCA~(TIP24)& \cmark & \cmark & \xmark  &  61.0 & 60.9 & 63.7 & 61.4 & 60.4 & 61.2 & 61.4\\
			CoOp~(IJCV22)& --  & -- & --  & 80.3 & 81.2 & 83.2 & 82.7 & 81.7 & 81.2 & 81.7\\
			AttriCLIP~(CVPR23)& \xmark & \cmark & \cmark & 74.1 & 77.9 & 80.2 & 79.9 & 80.0 & 78.0 & 78.4\\ 
			AD-CLIP~(ICCVW23)& \cmark & \xmark & \cmark  & 81.5 & 83.2 & 84.4 & 83.9 & 83.6 & 82.1 & 83.1\\
			DAMP~(CVPR24)& \cmark & \xmark & \cmark  &  82.5 & 83.5 & 84.2 & 83.2 & 82.3 & 80.2 & 82.6 \\
			PGA~(NeurIPS24)& \cmark & \xmark & \cmark  & 16.0 & 30.3 & 48.7 & 60.8 & 77.9 & 80.9 & 52.4\\
			DAPL~(TNNLS25)& \cmark & \xmark & \cmark  & \textbf{83.7} & \textbf{85.6} & 86.3 & 85.5 & 84.3 & 83.7 & 84.9 \\
			\midrule
			VisTA~(Ours)& \cmark & \cmark & \cmark  & 83.5 & 85.5 & \textbf{87.2} & \textbf{86.6} & \textbf{86.1} & \textbf{85.4} & \textbf{85.7}\\
			\bottomrule
		\end{tabular}
	}
	\label{Step-Domain}
\end{table*}

\begin{table*}[t]
\renewcommand\arraystretch{1.0}
    \begin{center}
    \caption{\label{tab:Office31} Classification accuracies (\%)  on Office-31. Note that the results outside the brackets are Step-level Accuracy, while the results in brackets represent the S-1 Accuracy in each time step. DA, CI, and RF respectively represent domain adaptation, class-incremental, and rehearsal-free.}
    \scalebox{0.75}{
         \begin{tabular}{llccc|cccccccc}
         \toprule
          Task & Method & DA & CI & RF & Step 1 & Step 2 & Step 3 & Avg.\\
         \midrule
         \multirow{9}{*}{A$\rightarrow$D} &
         ProCA~(ECCV22)& \cmark  & \cmark & \xmark & 97.4 (97.2)  & 90.6 (97.2)  & \textbf{91.5} (96.8)  & 93.2 (97.1)\\
         & PLDCA~(TIP24)& \cmark  & \cmark & \xmark & 90.9 (91.4)   & 90.0 (91.4)  & 91.5 (91.4)  & 90.8 (91.4)  \\
         & CoOp~(IJCV22)& --  & -- & --  & 97.4 (96.8) & \textbf{93.5} (97.4) & 89.0 (97.4) & \textbf{93.3} (97.4)\\
         & AttriCLIP~(CVPR23)& \xmark  & \cmark & \cmark  & 96.1 (96.1) & 71.5 (93.0)   & 81.8 (89.1) & 83.1 (92.7)\\
         & AD-CLIP~(ICCVW23)& \cmark  & \xmark & \cmark & 95.5 (96.1) & 90.3 (96.8) & 82.8 (96.1) & 89.2 (96.3)\\
         & DAMP~(CVPR24)& \cmark  & \xmark & \cmark & 96.1 (95.5) & 89.4 (95.5) & 85.1 (95.5) & 90.2 (95.5)\\
         & PGA~(NeurIPS24)& \cmark  & \xmark & \cmark & 93.0 (93.0)     & 62.9 (67.2) & 49.0 (79.7)   & 68.3 (80.0)\\
         & DAPL~(TNNLS25)& \cmark  & \xmark & \cmark  & 96.8 (96.8) & 87.7 (96.8) & 79.1 (97.4) & 87.9 (97.0)\\
         & VisTA~(Ours)& \cmark  & \cmark & \cmark & \textbf{98.7} (98.7) & 92.3 (96.8) & 87.8 (97.4) & 92.9 (97.6) \\
         \midrule
          \multirow{9}{*}{A$\rightarrow$W} 
          & ProCA~(ECCV22)& \cmark  & \cmark & \xmark & 94.5 (95.8)  & 87.7 (95.8)  & 86.3 (95.3)  & 89.5 (95.6)\\
         & PLDCA~(TIP24)& \cmark  & \cmark & \xmark & 94.0 (95.5)   & 90.0 (95.8)  & 90.7 (95.8)  & 91.6 (95.7)  \\
         & CoOp~(IJCV22)& --  & -- & --  & 96.6 (96.6) & 92.6 (96.6) & 87.0 (96.6) & 92.1 (96.6)\\
         & AttriCLIP~(CVPR23)& \xmark  & \cmark & \cmark  & 98.7 (98.7) & 92.6 (99.1) & 85.7 (97.4) & 92.3 (98.4)\\
         & AD-CLIP~(ICCVW23)& \cmark  & \xmark & \cmark & 99.1 (99.1) & 82.0 (93.6)   & 85.9 (99.6) & 89.0 (97.4)\\
         & DAMP~(CVPR24)& \cmark  & \xmark & \cmark & 99.1 (99.1) & 85.4 (96.6) & 79.3 (94.0)   & 87.9 (96.6)\\
         & PGA~(NeurIPS24)& \cmark  & \xmark & \cmark & 96.9 (96.9) & 67.8 (49.2) & 68.4 (75.0)   & 77.7 (73.7)\\
         & DAPL~(TNNLS25)& \cmark  & \xmark & \cmark  &  98.3 (98.3) & 84.7 (97.0)   & 76.7 (97.9) & 86.6 (97.7) \\
         & VisTA~(Ours)& \cmark  & \cmark & \cmark &  \textbf{100.0} (100.0) & \textbf{94.9} (98.3)  & \textbf{88.2} (98.3) & \textbf{94.4} (98.9)\\

         \midrule
         \multirow{9}{*}{D$\rightarrow$A} &
         ProCA~(ECCV22)& \cmark  & \cmark & \xmark & 81.5 (77.3) & 84.5 (76.0) & 75.6 (71.8) & 80.6 (75.0) \\
         & PLDCA~(TIP24)& \cmark  & \cmark & \xmark & 81.8 (77.5)   & 84.0 (78.2)  & 75.6 (74.9)  & 80.5 (76.9)  \\
         & CoOp~(IJCV22)& --  & -- & --  &  85.1 (85.1) & 90.4 (85.1) & 83.0 (85.1) & 86.2 (86.2) \\
         & AttriCLIP~(CVPR23)& \xmark  & \cmark & \cmark  & 84.0 (84.0)     & 81.3 (84.0)   & 79.5 (83.4) & 81.6 (83.8)\\
         & AD-CLIP~(ICCVW23)& \cmark  & \xmark & \cmark & 85.0 (85.0)     & 84.1 (89.5) & 83.0 (87.1)   & 84.0 (87.2)\\
         & DAMP~(CVPR24)& \cmark  & \xmark & \cmark & 85.3 (85.3) & 89.5 (84.5) & 81.3 (87.1) & 85.4 (85.6)\\
         & PGA~(NeurIPS24)& \cmark  & \xmark & \cmark & 86.5 (86.5) & 88.3 (86.1) & 81.2 (84.1) & 85.3 (85.6)\\
         & DAPL~(TNNLS25)& \cmark  & \xmark & \cmark  & \textbf{90.9} (90.9) & 88.3 (89.6) & 79.5 (90.8) & 86.2 (90.4)\\
         & VisTA~(Ours)& \cmark  & \cmark & \cmark & 85.6 (85.6)   & \textbf{90.9} (86.3)  & \textbf{84.5} (86.9) & \textbf{87.0} (86.3) \\
         \midrule
         
          \multirow{9}{*}{D$\rightarrow$W} 
          & ProCA~(ECCV22)& \cmark  & \cmark & \xmark &100.0 (100.0)  & \textbf{99.8} (100.0)  & \textbf{98.3} (100.0)  & \textbf{99.4} (100.0) \\
         & PLDCA~(TIP24)& \cmark  & \cmark & \xmark & 100.0 (100.0) & 97.3 (100.0) & 96.9 (100.0) & 98.1 (100.0) \\
         & CoOp~(IJCV22)& --  & -- & --  & 98.7 (98.7) & 98.3 (98.7) & 96.1 (98.7) & 97.7 (97.9) \\
         & AttriCLIP~(CVPR23)& \xmark  & \cmark & \cmark  & 98.7 (98.7) & 98.1 (98.7) & 94.4 (99.6) & 97.1 (99.0) \\
         & AD-CLIP~(ICCVW23)& \cmark  & \xmark & \cmark &  97.0 (97.0)     & 97.0 (98.7)   & 95.1 (99.1) & 96.4 (98.3)\\
         & DAMP~(CVPR24)& \cmark  & \xmark & \cmark & 99.6 (99.6) & 90.0 (99.6)   & 90.3 (99.1) & 93.3 (99.4)\\
         & PGA~(NeurIPS24)& \cmark  & \xmark & \cmark & 93.8 (93.8) & 96.5 (100)  & 88.7 (99.2) & 93.0 (97.7)\\
         & DAPL~(TNNLS25)& \cmark  & \xmark & \cmark  & 99.1 (99.1) & 83.0 (98.3)   & 79.2 (97.4) & 87.1 (98.3)\\
         & VisTA~(Ours)& \cmark  & \cmark & \cmark & \textbf{100.0} (100.0) & 97.7 (98.7) & 95.2 (99.6) & 97.6 (99.4)\\

         \midrule
         \multirow{9}{*}{W$\rightarrow$A} &
         ProCA~(ECCV22)& \cmark  & \cmark & \xmark & 83.7 (79.2)  & 83.2 (77.9)  & 77.2 (75.6)  & 81.4 (77.6)\\
         & PLDCA~(TIP24)& \cmark  & \cmark & \xmark & 82.0 (81.2)   & 84.9 (79.4)  & 77.8 (77.5)  & 81.6 (79.4)  \\
         & CoOp~(IJCV22)& --  & -- & --  & 84.7 (84.7) & 87.8 (84.7) & 81.5 (84.7) & 84.7 (84.7) \\
         & AttriCLIP~(CVPR23)& \xmark  & \cmark & \cmark  & 82.9 (82.9) & 87.0 (84.2)   & 79.8 (84.0)   & 83.2 (83.7)\\ 
         & AD-CLIP~(ICCVW23)& \cmark  & \xmark & \cmark & 85.6 (85.6) & 88.4 (84.5) & 81.4 (84.0)   & 85.1 (84.7)\\
         & DAMP~(CVPR24)& \cmark  & \xmark & \cmark & 86.7 (86.7) & \textbf{90.6} (86.2) & 80.1 (88.2) & 85.8 (87.0)\\
         & PGA~(NeurIPS24)& \cmark  & \xmark & \cmark & 86.6 (86.6) & 89.2 (85.8) & 81.3 (85.2) & 85.7 (85.9) \\
         & DAPL~(TNNLS25)& \cmark  & \xmark & \cmark  & \textbf{90.4} (90.4) & 88.8 (89.4) & 81.1 (89.9) & \textbf{86.8} (89.9)\\
         & VisTA~(Ours)& \cmark  & \cmark & \cmark &  85.2 (85.2)   & 90.5 (85.8)  & \textbf{84.0} (85.9) & 86.6 (85.6)\\
         
         \midrule
          \multirow{9}{*}{W$\rightarrow$D} 
          & ProCA~(ECCV22)& \cmark  & \cmark & \xmark & \textbf{100.0} (100.0)  & \textbf{99.7} (100.0)  & \textbf{99.4} (100.0)  & \textbf{99.8} (100.0)\\
         & PLDCA~(TIP24)& \cmark  & \cmark & \xmark & 100.0 (100.0) & 99.4 (100.0) & 99.4 (100.0) & 99.6 (100.0)\\
         & CoOp~(IJCV22)& --  & -- & --  & 98.1 (98.1) & 97.7 (98.1) & 98.1 (98.1) & 98.0 (98.1) \\
         & AttriCLIP~(CVPR23)& \xmark  & \cmark & \cmark  & 97.4 (97.4) & 96.8 (96.1) & 95.4 (96.8) & 96.5 (96.8)  \\
         & AD-CLIP~(ICCVW23)& \cmark  & \xmark & \cmark & 97.4 (97.4) & 94.5 (96.8) & 93.6 (96.8) & 95.2 (97.0) \\
         & DAMP~(CVPR24)& \cmark  & \xmark & \cmark & 95.5 (95.5) & 94.5 (92.9) & 91.7 (96.1) & 93.9 (94.8)\\
         & PGA~(NeurIPS24)& \cmark  & \xmark & \cmark & 96.1 (96.1) & 71.5 (93.0)   & 81.8 (89.1) & 83.1 (92.7)\\
         & DAPL~(TNNLS25)& \cmark  & \xmark & \cmark  &  96.8 (96.8) & 87.1 (96.8) & 79.3 (96.1) & 87.7 (96.6) \\
         & VisTA~(Ours)& \cmark  & \cmark & \cmark & 97.4 (97.4)   & 95.2 (94.2)  & 96.5 (95.5) & 96.4 (95.7)\\
         \bottomrule
         \end{tabular}
         }
    \end{center}
\end{table*}

\begin{table*}[t]
    \begin{center}
    \caption{\label{tab:OH_Ar} Classification accuracies (\%)  on Office-Home with Art as source domain. Note that the results outside the brackets are Step-level Accuracy, while the results in brackets represent the S-1 Accuracy in each time step. DA, CI, and RF respectively represent domain adaptation, class-incremental, and rehearsal-free.}
    \scalebox{0.75}{
         \begin{tabular}{llp{3mm}p{3mm}p{3mm}|ccccccc}
         \toprule
          Task & Method & DA & CI & RF & Step 1 & Step 2 & Step 3 & Step 4 & Step 5 & Step 6 & Avg.\\
         \midrule
         \multirow{9}{*}{A$\rightarrow$C} & ProCA~(ECCV22)& \cmark  & \cmark & \xmark & 58.9 (59.4) & 60.3 (57.5) & 55.7 (56.9) & 56.9 (58.5) & 56.8 (55.8) & 56.8 (55.3) & 57.6 (57.2) \\
         & PLDCA~(TIP24)& \cmark  & \cmark & \xmark & 57.8 (62.3) & 58.1 (57.8) & 54.7 (55.6) & 58.1 (57.0) & 58.7 (55.2) & 58.3 (53.4) & 57.6 (56.9) \\
         & CoOp~(IJCV22)& --  & -- & --   & 65.4 (65.4) & 71.1 (65.4) & 67.4 (65.4) & 69.4 (65.4) & 70.7 (65.4) & 70.6 (65.4) & 69.1 (65.4) \\
         & AttriCLIP~(CVPR23)& \xmark  & \cmark & \cmark  & 63.1 (63.1) & 71.1 (60.2) & 70.1 (61.4) & 68.3 (59.3) & 68.0 (60.1) & 67.4 (56.1) & 68.0 (60.0)\\
         & AD-CLIP~(ICCVW23)& \cmark  & \xmark & \cmark & \textbf{68.5} (68.5) & 71.2 (62.8) & 68.4 (62.0) & 71.1 (63.1) & 70.0 (60.1) & 70.9 (61.1) & 70.0 (62.9)\\
         & DAMP~(CVPR24)& \cmark  & \xmark & \cmark & 67.5 (67.5) & 70.6 (63.5) & 68.4 (67.0) & 69.7 (66.1) & 68.2 (61.3) & 68.3 (64.3) & 68.6 (65.0)\\
         & PGA~(NeurIPS24)& \cmark  & \xmark & \cmark & 55.8 (55.8) & 64.6 (50.5) & 58.5 (51.6) & 58.4 (51.9) & 61.8 (52.3) & 61.6 (45.0) & 60.1 (51.2)\\
         & DAPL~(TNNLS25)& \cmark  & \xmark & \cmark  & 61.0 (61.0) & 72.0 (62.6) & 67.7 (61.9) & 70.0 (62.9) & 70.4 (59.3) & 69.3 (60.5) & 68.4 (61.4)\\
         & VisTA~(Ours)& \cmark  & \cmark & \cmark & 61.9   (61.9) & \textbf{72.2} (63.5) & \textbf{70.4} (64.4) & \textbf{71.9} (64.1) & \textbf{72.4} (63.4) & \textbf{71.8} (64.8) & \textbf{70.1} (63.7) \\
         
         \midrule
          \multirow{9}{*}{A$\rightarrow$P} & ProCA~(ECCV22)& \cmark  & \cmark & \xmark &  89.1 (86.7) & 84.5 (85.2) & 80.3 (84.6) & 81.9 (83.2) & 82.1 (82.5) & 81.9 (83.9) & 83.3 (84.4) \\
         & PLDCA~(TIP24)& \cmark  & \cmark & \xmark & 79.7 (79.1) & 82.0 (77.2) & 80.6 (75.6) & 83.9 (76.6) & 83.5 (75.5) & 83.2 (73.8) & 82.1 (76.3) \\
         & CoOp~(IJCV22)& --  & -- & --  & 86.1 (86.1) & 87.4 (86.1) & 88.2 (86.1) & 89.1 (86.1) & 88.9 (86.1) & 88.9 (86.1) & 88.1 (86.1)\\
         & AttriCLIP~(CVPR23)& \xmark  & \cmark & \cmark  & \textbf{92.1} (92.1) & 90.5 (84.8) & 86.6 (88.6) & 89.4 (88.3) & 89.8 (85.4) & 87.9 (86.7) & 89.4 (87.7)\\
         & AD-CLIP~(ICCVW23)& \cmark  & \xmark & \cmark & 90.1 (90.1) & 88.2 (87.1) & 86.4 (80.3) & 91.3 (86.5) & 90.8 (88.8) & 91.2 (88.6) & 89.7 (86.9)\\
         & DAMP~(CVPR24)& \cmark  & \xmark & \cmark & 92.1 (92.1) & 86.9 (87.9) & 89.1 (86.4) & 89.1 (85.1) & 89.1 (83.6) & 86.7 (80.3) & 88.8 (85.9)\\
         & PGA~(NeurIPS24)& \cmark  & \xmark & \cmark & 87.8 (87.8) & 79.0 (72.3) & 84.2 (73.9) & 77.4 (58.3) & 80.2 (75.0) & 70.9 (66.1) & 79.9 (72.2)\\
         & DAPL~(TNNLS25)& \cmark  & \xmark & \cmark  & 88.5 (88.5) & 85.4 (83.3) & 89.5 (87.7) & 90.8 (89.1) & 91.5 (87.1) & 91.2 (88.8) & 89.5 (87.4)\\
         & VisTA~(Ours)& \cmark  & \cmark & \cmark & 91.8   (91.8) & \textbf{91.7} (91.5) & \textbf{92.1} (91.8) & \textbf{92.9} (91.8) & \textbf{93.1} (91.5) & \textbf{92.8} (92.4) & \textbf{92.4} (91.8) \\
         \midrule
         
          \multirow{9}{*}{A$\rightarrow$R} & ProCA~(ECCV22)& \cmark  & \cmark & \xmark & 86.5 (86.7) & 90.2 (86.1) & 88.2 (85.1) & 89.2 (84.8) & 90.0 (83.6) & 89.9 (83.8) & 89.0 (85.0) \\
         & PLDCA~(TIP24)& \cmark  & \cmark & \xmark & 86.1 (85.8) & \textbf{90.7} (84.4) & 89.0 (82.5) & 88.9 (82.1) & 88.8 (82.2) & 89.5 (81.5) & 88.8 (83.1) \\
         & CoOp~(IJCV22)& --  & -- & --  & 86.2 (86.2) & 89.2 (86.2) & 89.0 (86.2) & 89.8 (86.2) & 89.9 (86.2) & 89.8 (86.2) & 89.0 (86.2)\\
         & AttriCLIP~(CVPR23)& \xmark  & \cmark & \cmark  & 88.3 (88.3) & 90.2 (85.2) & 90.0 (84.3) & 90.3 (87.1) & 90.8 (86.4) & 89.4 (85.8) & 89.8 (86.2)\\
         & AD-CLIP~(ICCVW23)& \cmark  & \xmark & \cmark & 84.8 (84.8) & 89.4 (85.4) & 85.6 (83.3) & 89.7 (84.2) & 89.4 (84.8) & 89.3 (85.7) & 88.0 (84.7)\\
         & DAMP~(CVPR24)& \cmark  & \xmark & \cmark & 87.7 (87.7) & 89.4 (85.5) & 88.5 (85.1) & 89.7 (86.5) & 88.5 (83.3) & 88.7 (84.2) & 88.8 (85.4)\\
         & PGA~(NeurIPS24)& \cmark  & \xmark & \cmark & 89.2 (89.2) & 81.0 (71.4) & 82.5 (70.9) & 78.7 (79.8) & 83.3 (71.4) & 85.3 (74.4) & 83.3 (76.2)\\
         & DAPL~(TNNLS25)& \cmark  & \xmark & \cmark  & \textbf{90.7} (90.7) & 90.3 (86.7) & 89.3 (89.1) & 90.0 (87.1) & 90.3 (85.8) & 90.5 (86.4) & \textbf{90.2} (87.6)\\
         & VisTA~(Ours)& \cmark  & \cmark & \cmark & 86.4   (86.4) & 90.3 (86.8) & \textbf{90.5} (86.8) & \textbf{90.9} (86.5) & \textbf{91.1} (86.4) & \textbf{91.3} (87.0) & 90.1 (86.6) \\         
         \bottomrule
         \end{tabular}
         }
    \end{center}
\end{table*}

\begin{table*}[t]
\renewcommand\arraystretch{1.0}
    \begin{center}
    \caption{\label{tab:OH_Cl} Classification accuracies (\%)  on Office-Home with Clipart as source domain. Note that the results outside the brackets are Step-level Accuracy, while the results in brackets represent the S-1 Accuracy in each time step. DA, CI, and RF respectively represent domain adaptation, class-incremental, and rehearsal-free.}
    \scalebox{0.75}{
         \begin{tabular}{llp{3mm}p{3mm}p{3mm}|cccccccc}
         \toprule
          Task & Method & DA & CI & RF & Step 1 & Step 2 & Step 3 & Step 4 & Step 5 & Step 6 & Avg.\\
         \midrule
         \multirow{9}{*}{C$\rightarrow$A} & ProCA~(ECCV22)& \cmark  & \cmark & \xmark & 67.6 (67.2) & 66.9 (66.2) & 64.7 (67.7) & 65.9 (67.4) & 66.3 (66.2) & 68.4 (66.7) & 66.6 (66.9) \\
         & PLDCA~(TIP24)& \cmark  & \cmark & \xmark & 67.1 (70.6) & 71.5 (66.8) & 70.7 (66.6) & 70.1 (64.0) & 73.8 (66.9) & 76.5 (64.6) & 71.6 (66.6) \\
         & CoOp~(IJCV22)& --  & -- & --  & 81.3 (81.3) & 82.0 (81.3) & 80.5 (81.3) & 80.9 (81.3) & 80.6 (81.3) & 81.4 (81.3) & 81.1 (81.3)\\
         & AttriCLIP~(CVPR23)& \xmark  & \cmark & \cmark  & 83.5 (83.5) & 83.8 (80.0) & 81.1 (78.5) & 82.7 (81.0) & 81.8 (79.7) & 84.0 (79.7) & 82.8 (80.4)\\
         & AD-CLIP~(ICCVW23)& \cmark  & \xmark & \cmark & 78.0 (78.0) & 84.5 (82.3) & 79.0 (72.2) & 82.1 (78.2) & 81.5 (78.0) & 82.2 (81.0) & 81.2 (78.3)\\
         & DAMP~(CVPR24)& \cmark  & \xmark & \cmark & 73.9 (73.9) & 83.9 (80.0) & 81.3 (77.8) & 81.9 (78.5) & 78.6 (70.4) & 81.3 (76.7) & 80.2 (76.2)\\
         & PGA~(NeurIPS24)& \cmark  & \xmark & \cmark & \textbf{84.9} (84.9) & 50.8 (13.0) & 73.8 (64.1) & 62.6 (72.4) & 60.0 (33.1) & 55.7 (51.0) & 64.6 (53.1)\\
         & DAPL~(TNNLS25)& \cmark  & \xmark & \cmark  & 83.0 (83.0) & \textbf{85.8} (80.3) & 82.7 (80.3) & 82.7 (78.2) & 82.8 (78.7) & 83.3 (78.2) & \textbf{83.4} (79.8)\\
         & VisTA~(Ours)& \cmark  & \cmark & \cmark & 78.2   (78.2) & 84.9 (78.7) & \textbf{83.3} (78.7) & \textbf{83.8} (79.5) & \textbf{83.1} (79.2) & \textbf{84.4} (80.8) & 83.0 (79.2) \\         
         \midrule
          \multirow{9}{*}{C$\rightarrow$P} & ProCA~(ECCV22)& \cmark  & \cmark & \xmark &  82.5 (83.5) & 80.0 (82.5) & 79.5 (82.7) & 80.5 (82.1) & 81.1 (82.0) & 80.9 (81.9) & 80.8 (82.4) \\
         & PLDCA~(TIP24)& \cmark  & \cmark & \xmark & 80.0 (82.2) & 79.4 (81.6) & 80.2 (82.0) & 81.2 (81.5) & 81.6 (80.7) & 83.3 (80.4) & 81.0 (81.4) \\
         & CoOp~(IJCV22)& --  & -- & --  & 90.6 (90.6) & 87.6 (90.6) & 88.7 (90.6) & 90.0 (90.6) & 89.9 (90.6) & 88.4 (90.6) & 89.2 (90.6)\\
         & AttriCLIP~(CVPR23)& \xmark  & \cmark & \cmark  & 92.8 (92.8) & \textbf{93.8} (91.8) & 89.3 (87.1) & 90.1 (89.4) & 89.2 (88.6) & 90.2 (89.4) & 90.9 (89.9)\\
         & AD-CLIP~(ICCVW23)& \cmark  & \xmark & \cmark & 92.2 (92.2) & 89.8 (90.4) & 92.3 (90.9) & 91.5 (89.4) & 91.6 (91.8) & 91.3 (91.0) & 91.5 (91.0)\\
         & DAMP~(CVPR24)& \cmark  & \xmark & \cmark & \textbf{93.1} (93.1) & 88.5 (89.1) & 90.4 (90.7) & 91.3 (89.5) & 90.8 (87.1) & 90.1 (89.4) & 90.7 (89.8)\\
         & PGA~(NeurIPS24)& \cmark  & \xmark & \cmark & 87.3 (87.3) & 83.3 (81.7) & 82.3 (63.8) & 83.1 (80.3) & 85.7 (73.1) & 83.7 (83.8) & 84.2 (78.3)\\
         & DAPL~(TNNLS25)& \cmark  & \xmark & \cmark  & 89.4 (89.4) & 85.9 (87.3) & 89.4 (88.5) & 90.5 (90.1) & 91.1 (88.0) & 90.1 (84.6) & 89.4 (88.0)\\
         & VisTA~(Ours)& \cmark  & \cmark & \cmark & 92.1   (92.1) & 91.5 (92.1) & \textbf{92.3} (92.2) & \textbf{93.6} (92.7) & \textbf{93.4} (92.7) & \textbf{92.9} (92.8) & \textbf{92.6} (92.4) \\
         \midrule
         
          \multirow{9}{*}{C$\rightarrow$R} & ProCA~(ECCV22)& \cmark  & \cmark & \xmark & 79.5 (79.7) & 84.3 (77.5) & 84.6 (78.3) & 83.1 (75.8) & 83.7 (76.2) & 83.9 (77.2) & 83.2 (77.5) \\
         & PLDCA~(TIP24)& \cmark  & \cmark & \xmark & 73.7 (74.8) & 80.4 (73.0) & 80.6 (70.8) & 81.5 (70.6) & 84.1 (71.5) & 85.3 (70.9) & 80.9 (71.9) \\
         & CoOp~(IJCV22)& --  & -- & --  & 84.3 (84.3) & 87.6 (84.3) & 86.9 (84.3) & 87.9 (84.3) & 88.1 (84.3) & 87.9 (84.3) & 87.1 (84.3)\\
         & AttriCLIP~(CVPR23)& \xmark  & \cmark & \cmark  & 88.0 (88.0) & 90.9 (86.7) & 90.1 (87.0) & 89.8 (87.1) & 89.0 (82.4) & 87.3 (74.7) & 89.2 (84.3)\\
         & AD-CLIP~(ICCVW23)& \cmark  & \xmark & \cmark & 87.6 (87.6) & 85.1 (78.9) & 88.1 (87.1) & 89.7 (87.6) & 87.8 (81.3) & 90.4 (85.9) & 88.1 (84.7)\\
         & DAMP~(CVPR24)& \cmark  & \xmark & \cmark & 88.3 (88.3) & 89.7 (86.4) & 89.0 (86.5) & 88.6 (86.4) & 87.3 (85.7) & 88.2 (84.8) & 88.5 (86.4)\\
         & PGA~(NeurIPS24)& \cmark  & \xmark & \cmark & 88.8 (88.8) & 81.2 (68.1) & 82.3 (75.5) & 82.0 (72.0) & 79.6 (64.5) & 84.3 (74.1) & 83.0 (73.8)\\
         & DAPL~(TNNLS25)& \cmark  & \xmark & \cmark  & \textbf{90.9 }(90.9) & \textbf{91.1} (88.2) & 89.4 (88.8) & 89.0 (87.4) & 88.8 (80.3) & 90.1 (86.2) & 89.9 (87.0)\\
         & VisTA~(Ours)& \cmark  & \cmark & \cmark & 87.0   (87.0) & 89.9 (87.1) & \textbf{90.3} (87.4) & \textbf{90.8} (87.6) & \textbf{90.8} (87.9) & \textbf{91.1} (87.7) & \textbf{90.0} (87.5) \\
         
         \bottomrule
         \end{tabular}
         }
    \end{center}
\end{table*}

\begin{table*}[t]
\renewcommand\arraystretch{1.0}
    \begin{center}
    \caption{\label{tab:OH_Pr} Classification accuracies (\%)  on Office-Home with Product as source domain. Note that the results outside the brackets are Step-level Accuracy, while the results in brackets represent the S-1 Accuracy in each time step. DA, CI, and RF respectively represent domain adaptation, class-incremental, and rehearsal-free.}
    \scalebox{0.75}{
         \begin{tabular}{llp{3mm}p{3mm}p{3mm}|cccccccc}
         \toprule
          Task & Method & DA & CI & RF & Step 1 & Step 2 & Step 3 & Step 4 & Step 5 & Step 6 & Avg.\\
         \midrule
         \multirow{9}{*}{P$\rightarrow$A} & ProCA~(ECCV22)& \cmark  & \cmark & \xmark & 63.8 (66.0) & 68.1 (65.9) & 66.4 (65.4) & 68.7 (65.7) & 68.8 (66.2) & 70.1 (65.0) & 67.6 (65.7) \\
         & PLDCA~(TIP24)& \cmark  & \cmark & \xmark & 63.3 (64.1) & 68.6 (62.7) & 67.8 (62.3) & 68.6 (63.3) & 72.6 (63.5) & 74.2 (62.6) & 69.2 (63.1) \\
         & CoOp~(IJCV22)& --  & -- & --  & 72.2 (72.2) & 78.3 (72.2) & 78.4 (72.2) & 79.1 (72.2) & 78.8 (72.2) & 79.8 (72.2) & 77.8 (72.2)\\
         & AttriCLIP~(CVPR23)& \xmark  & \cmark & \cmark  & 82.5 (82.5) & \textbf{85.4} (81.5) & 82.8 (79.2) & 82.2 (77.5) & 80.8 (78.7) & 82.3 (74.7) & 82.7 (79.0)\\
         & AD-CLIP~(ICCVW23)& \cmark  & \xmark & \cmark & 73.7 (73.7) & 80.9 (75.4) & 75.2 (78.9) & 79.6 (72.9) & 80.8 (76.7) & 80.4 (77.2) & 78.4 (75.8)\\
         & DAMP~(CVPR24)& \cmark  & \xmark & \cmark & 76.5 (76.5) & 80.2 (71.9) & 80.0 (77.0) & 79.9 (71.1) & 79.1 (72.2) & 81.5 (77.5) & 79.5 (74.4)\\
         & PGA~(NeurIPS24)& \cmark  & \xmark & \cmark & 83.3 (83.3) & 70.8 (54.9) & 64.8 (56.0) & 54.9 (38.5) & 67.2 (62.0) & 58.5 (53.9) & 66.6 (58.1)\\
         & DAPL~(TNNLS25)& \cmark  & \xmark & \cmark  & \textbf{83.8} (83.8) & 84.5 (78.5) & 81.8 (79.2) & 82.8 (82.5) & \textbf{83.3} (79.5) & 83.3 (78.7) & 83.3 (80.4)\\
         & VisTA~(Ours)& \cmark  & \cmark & \cmark & 81.0   (81.0) & 84.7 (79.5) & \textbf{83.9} (80.0) & \textbf{84.5} (80.0) & 83.1 (79.5) & \textbf{84.7} (80.3) & \textbf{83.7} (80.0) \\
         
         \midrule
          \multirow{9}{*}{P$\rightarrow$C} & ProCA~(ECCV22)& \cmark  & \cmark & \xmark & 61.3 (59.5) & 52.1 (59.7) & 48.4 (57.8) & 48.9 (57.4) & 51.5 (58.1) & 51.3 (56.7) & 52.3 (58.2) \\
         & PLDCA~(TIP24)& \cmark  & \cmark & \xmark & 56.3 (54.5) & 50.3 (51.7) & 50.5 (48.9) & 50.8 (46.9) & 54.2 (46.6) & 54.5 (45.0) & 52.8 (48.9) \\
         & CoOp~(IJCV22)& --  & -- & --  & 66.3 (66.3) & 69.4 (66.3) & 66.7 (66.3) & 68.8 (66.3) & 69.6 (66.3) & 69.6 (66.3) & 68.4 (66.3)\\
         & AttriCLIP~(CVPR23)& \xmark  & \cmark & \cmark  & 70.3 (70.3) & 74.2 (70.7) & \textbf{71.9} (66.1) & 69.8 (61.6) & 70.3 (62.6) & 67.3 (55.2) & 70.6 (64.4) \\
         & AD-CLIP~(ICCVW23)& \cmark  & \xmark & \cmark & \textbf{71.9} (71.9) & 73.0 (68.8) & 70.7 (69.9) & 71.0 (67.6) & 72.0 (69.3) & 71.9 (70.2) & \textbf{71.8} (69.6)\\
         & DAMP~(CVPR24)& \cmark  & \xmark & \cmark & 66.6 (66.6) & \textbf{76.4} (72.0) & 69.9 (66.1) & 68.5 (66.9) & 70.2 (64.4) & 68.3 (67.0) & 70.0 (67.2)\\
         & PGA~(NeurIPS24)& \cmark  & \xmark & \cmark & 56.4 (56.4) & 67.3 (58.6) & 62.4 (52.0) & 55.4 (42.0) & 63.6 (45.9) & 59.7 (46.3) & 60.8 (50.2)\\
         & DAPL~(TNNLS25)& \cmark  & \xmark & \cmark  & 63.5 (63.5) & 70.7 (61.3) & 67.0 (59.0) & 68.4 (62.2) & 69.5 (60.5) & 68.9 (59.2) & 68.0 (60.9)\\
         & VisTA~(Ours)& \cmark  & \cmark & \cmark & 65.1   (65.1) & 72.2 (64.1) & 70.7 (65.7) & \textbf{72.6} (65.8) & \textbf{73.3} (66.1) & \textbf{72.8} (66.4) & 71.1 (65.5) \\
         \midrule
         
          \multirow{9}{*}{P$\rightarrow$R} & ProCA~(ECCV22)& \cmark  & \cmark & \xmark & 85.2 (85.3) & 89.5 (83.4) & 86.5 (83.1) & 88.7 (83.2) & 89.1 (81.6) & 89.9 (79.9) & 88.1 (82.8) \\
         & PLDCA~(TIP24)& \cmark  & \cmark & \xmark & 84.8 (84.9) & 87.8 (82.9) & 89.0 (80.5) & 89.1 (80.4) & 89.2 (78.6) & 87.6 (77.9) & 87.9 (80.9) \\
         & CoOp~(IJCV22)& --  & -- & --  & 85.1 (85.1) & 88.2 (85.1) & 87.8 (85.1) & 88.9 (85.1) & 89.2 (85.1) & 89.6 (85.1) & 88.1 (85.1)\\
         & AttriCLIP~(CVPR23)& \xmark  & \cmark & \cmark  & 88.3 (88.3) & 88.9 (82.5) & 90.2 (86.5) & \textbf{90.7} (86.5) & 90.3 (84.8) & 88.7 (83.3) & 89.5 (85.3)\\
         & AD-CLIP~(ICCVW23)& \cmark  & \xmark & \cmark & 84.2 (84.2) & 89.4 (86.5) & 88.8 (84.5) & 89.1 (85.7) & 89.5 (86.2) & 90.5 (87.3) & 88.6 (85.7)\\
         & DAMP~(CVPR24)& \cmark  & \xmark & \cmark & 87.4 (87.4) & 89.3 (85.2) & 88.8 (85.1) & 88.2 (85.1) & 88.8 (83.1) & 89.1 (84.5) & 88.6 (85.1)\\
         & PGA~(NeurIPS24)& \cmark  & \xmark & \cmark & 88.9 (88.9) & 84.9 (77.0) & 81.8 (71.1) & 85.8 (70.9) & 74.5 (59.5) & 81.3 (75.8) & 82.9 (73.9)\\
         & DAPL~(TNNLS25)& \cmark  & \xmark & \cmark  & \textbf{90.3} (90.3) & 89.7 (85.2) & 89.7 (88.6) & 89.7 (88.8) & 89.5 (84.9) & 90.2 (86.7) & 89.9 (87.4)\\
         & VisTA~(Ours)& \cmark  & \cmark & \cmark & 87.3   (87.3) & \textbf{90.4} (87.3) & \textbf{90.6} (87.7) & 90.4 (86.1) & \textbf{91.0} (87.6) & \textbf{91.5} (87.7) & \textbf{90.2} (87.3) \\
         
         \bottomrule
         \end{tabular}
         }
    \end{center}
\end{table*}

\begin{table*}[t]
\renewcommand\arraystretch{1.0}
    \begin{center}
    \caption{\label{tab:OH_Rw} Classification accuracies (\%)  on Office-Home with Real World as source domain. Note that the results outside the brackets are Step-level Accuracy, while the results in brackets represent the S-1 Accuracy in each time step. DA, CI, and RF respectively represent domain adaptation, class-incremental, and rehearsal-free.}
    \scalebox{0.75}{
         \begin{tabular}{llp{3mm}p{3mm}p{3mm}|cccccccc}
         \toprule
          Task & Method & DA & CI & RF & Step 1 & Step 2 & Step 3 & Step 4 & Step 5 & Step 6 & Avg.\\
         \midrule
         \multirow{9}{*}{R$\rightarrow$A} & ProCA~(ECCV22)& \cmark  & \cmark & \xmark & 65.8 (68.8) & 72.1 (70.2) & 72.9 (69.5) & 76.1 (70.4) & 75.6 (70.1) & 77.4 (69.5) & 73.3 (69.8) \\
         & PLDCA~(TIP24)& \cmark  & \cmark & \xmark & 68.9 (74.4) & 76.7 (72.6) & 78.1 (72.4) & 80.1 (71.6) & 80.3 (71.8) & 80.4 (70.4) & 77.4 (72.2) \\
         & CoOp~(IJCV22)& --  & -- & --  & 79.2 (79.2) & 83.1 (79.2) & 82.7 (79.2) & 83.4 (79.2) & 82.5 (79.2) & 83.3 (79.2) & 82.4 (79.2)\\
         & AttriCLIP~(CVPR23)& \xmark  & \cmark & \cmark  & 83.5 (83.5) & 85.2 (79.5) & 80.5 (75.4) & 84.0 (76.7) & 81.9 (72.2) & 83.9 (78.0) & 83.2 (77.5)\\
         & AD-CLIP~(ICCVW23)& \cmark  & \xmark & \cmark & 72.7 (72.7) & 85.3 (80.8) & 82.4 (79.2) & 81.3 (72.4) & 82.0 (79.5) & 83.3 (74.7) & 81.2 (76.6)\\
         & DAMP~(CVPR24)& \cmark  & \xmark & \cmark & 74.7 (74.7) & 81.6 (72.7) & 79.4 (74.7) & 81.2 (74.4) & 81.7 (77.5) & 80.3 (68.1) & 79.8 (73.7)\\
         & PGA~(NeurIPS24)& \cmark  & \xmark & \cmark & 80.5 (80.5) & 76.7 (64.6) & 37.3 (7.8)  & 72.2 (60.2) & 47.2 (28.6) & 59.4 (34.4) & 62.2 (46.0)\\
         & DAPL~(TNNLS25)& \cmark  & \xmark & \cmark  & \textbf{83.5} (83.5) & \textbf{85.3} (80.0) & 81.3 (77.5) & 82.8 (77.5) & 82.3 (73.9) & 82.2 (74.2) & 82.9 (77.8)\\
         & VisTA~(Ours)& \cmark  & \cmark & \cmark & 80.3   (80.3) & 85.2 (79.2) & \textbf{84.4} (80.3) & \textbf{84.8} (80.0) & \textbf{83.5} (78.5) & \textbf{84.8} (80.3) & \textbf{83.8} (79.8) \\
         
         \midrule
          \multirow{9}{*}{R$\rightarrow$C} & ProCA~(ECCV22)& \cmark  & \cmark & \xmark & 50.8 (52.6) & 50.6 (49.6) & 51.0 (50.8) & 52.0 (52.5) & 51.9 (50.9) & 50.3 (50.9) & 51.1 (51.2) \\
         & PLDCA~(TIP24)& \cmark  & \cmark & \xmark & 48.0 (50.2) & 60.5 (49.2) & 56.5 (48.9) & 57.2 (49.6) & 56.7 (48.0) & 55.8 (46.9) & 55.8 (48.8) \\
         & CoOp~(IJCV22)& --  & -- & --  & 63.1 (63.1) & 71.3 (63.1) & 71.0 (63.1) & 73.0 (63.1) & 72.7 (63.1) & 71.6 (63.1) & 70.4 (63.1)\\
         & AttriCLIP~(CVPR23)& \xmark  & \cmark & \cmark  & 66.1 (66.1) & 69.1 (56.1) & 57.0 (47.7) & 62.8 (53.1) & 72.2 (63.1) & 69.0 (62.2) & 66.0 (58.0)\\
         & AD-CLIP~(ICCVW23)& \cmark  & \xmark & \cmark & 62.9 (62.9) & 72.9 (65.7) & \textbf{71.5} (67.5) & \textbf{81.6} (66.6) & 72.3 (62.6) & 71.4 (62.3) & \textbf{72.1} (64.6)\\
         & DAMP~(CVPR24)& \cmark  & \xmark & \cmark & \textbf{72.5} (72.5) & \textbf{74.7} (66.7) & 70.0 (66.1) & 68.9 (60.7) & 70.6 (59.3) & 69.0 (59.8) & 71.0 (64.2)\\
         & PGA~(NeurIPS24)& \cmark  & \xmark & \cmark & 54.7 (54.7) & 64.3 (50.2) & 64.6 (52.8) & 59.2 (42.3) & 61.0 (50.0) & 63.1 (51.9) & 61.1 (50.3)\\
         & DAPL~(TNNLS25)& \cmark  & \xmark & \cmark  & 64.4 (64.4) & 72.4 (62.6) & 66.8 (59.9) & 68.8 (63.7) & 71.0 (61.7) & 70.4 (62.0) & 69.0 (62.4)\\
         & VisTA~(Ours)& \cmark  & \cmark & \cmark & 61.4   (61.4) & 71.8 (62.3) & 70.3 (62.8) & 72.6 (62.2) & \textbf{73.3} (64.1) & \textbf{72.2} (64.3) & 70.3 (62.8) \\
         \midrule
         
          \multirow{9}{*}{R$\rightarrow$P} & ProCA~(ECCV22)& \cmark  & \cmark & \xmark & 87.4 (90.8) & 88.8 (89.0) & 88.8 (89.7) & 87.9 (88.6) & 89.1 (88.0) & 87.9 (88.4) & 88.3 (89.1) \\
         & PLDCA~(TIP24)& \cmark  & \cmark & \xmark & 83.9 (85.8) & 86.9 (84.9) & 86.9 (85.3) & 88.3 (84.6) & 89.4 (84.2) & 90.0 (83.6) & 87.6 (84.7)\\
         & CoOp~(IJCV22)& --  & -- & --  & 90.1 (90.1) & 89.4 (90.1) & 91.6 (90.1) & 92.2 (90.1) & 92.2 (90.1) & 91.9 (90.1) & 91.2 (90.1)\\
         & AttriCLIP~(CVPR23)& \xmark  & \cmark & \cmark  & \textbf{93.7} (93.7) & 91.1 (90.6) & 92.2 (90.3) & 92.4 (90.6) & 92.6 (88.9) & 90.4 (80.7) & 92.1 (89.1)\\
         & AD-CLIP~(ICCVW23)& \cmark  & \xmark & \cmark & 87.0 (87.0) & 91.6 (88.9) & \textbf{93.4} (89.8) & 91.9 (88.8) & 91.7 (88.2) & 92.7 (87.6) & 91.4 (88.4)\\
         & DAMP~(CVPR24)& \cmark  & \xmark & \cmark & 91.2 (91.2) & 88.8 (89.8) & 90.4 (88.2) & 91.3 (88.8) & 91.0 (89.8) & 90.7 (84.0) & 90.6 (88.6)\\
         & PGA~(NeurIPS24)& \cmark  & \xmark & \cmark & 89.2 (89.2) & 82.7 (75.9) & 78.9 (58.6) & 86.8 (83.0) & 87.6 (79.4) & 82.1 (74.1) & 84.5 (76.7)\\
         & DAPL~(TNNLS25)& \cmark  & \xmark & \cmark  & 90.7 (90.7) & 90.7 (87.1) & 88.6 (85.9) & 89.5 (87.6) & 90.2 (84.6) & 90.8 (87.1) & 90.1 (87.2)\\
         & VisTA~(Ours)& \cmark  & \cmark & \cmark & 91.5   (91.5) & \textbf{91.6} (91.9) & 92.0 (91.6) & \textbf{92.8} (91.3) & \textbf{92.8} (90.9) & \textbf{92.8} (91.5) & \textbf{92.2} (91.5)\\
         
         \bottomrule
         \end{tabular}
         }
    \end{center}
\end{table*}

\begin{table*}[t]
\renewcommand\arraystretch{1.0}
    \begin{center}
    \caption{\label{tab:MD_Cl} Classification accuracies (\%)  on Mini-DomianNet with Clipart as source domain. Note that the results outside the brackets are Step-level Accuracy, while the results in brackets represent the S-1 Accuracy in each time step. DA, CI, and RF respectively represent domain adaptation, class-incremental, and rehearsal-free.}
    \scalebox{0.75}{
         \begin{tabular}{llp{3mm}p{3mm}p{3mm}|cccccccc}
         \toprule
          Task & Method & DA & CI & RF & Step 1 & Step 2 & Step 3 & Step 4 & Step 5 & Step 6 & Avg.\\
         \midrule
         \multirow{9}{*}{C$\rightarrow$P} & ProCA~(ECCV22)& \cmark  & \cmark & \xmark & 55.0 (55.0) & 54.0 (49.0) & 52.0 (44.0) & 51.2 (43.0) & 49.8 (43.0) & 50.7 (42.0) & 52.1 (46.0) \\
         & PLDCA~(TIP24)& \cmark  & \cmark & \xmark & 51.3 (56.0) & 57.0 (52.1) & 60.0 (48.8) & 53.2 (43.9) & 55.0 (46.1) & 56.2 (41.2) & 55.4 (48.0) \\
         & CoOp~(IJCV22)& --  & -- & --  & 74.0 (70.4) & 81.0 (70.4) & 80.0 (70.4) & 78.2 (70.4) & 77.0 (70.4) & 78.5 (70.4) & 78.1 (70.4)\\
         & AttriCLIP~(CVPR23)& \xmark  & \cmark & \cmark  & 73.0 (65.2) & 78.5 (69.5) & 76.7 (66.6) & 75.8 (70.3) & 76.2 (68.1) & 73.8 (65.5) & 75.7 (67.5)\\
         & AD-CLIP~(ICCVW23)& \cmark  & \xmark & \cmark & 78.0 (73.9) & 82.5 (72.9) & 79.0 (68.4) & 80.5 (71.1) & 79.4 (71.9) & 77.5 (69.4) & 79.5 (71.3)\\
         & DAMP~(CVPR24)& \cmark  & \xmark & \cmark & 79.0 (71.1) & 79.5 (70.2) & 79.7 (69.1) & 77.8 (69.9) & 78.4 (70.9) & 77.8 (72.4) & 78.7 (70.6)\\
         & PGA~(NeurIPS24)& \cmark  & \xmark & \cmark & 15.6 (15.6) & 31.4 (14.5) & 48.6 (15.0) & 63.7 (16.0) & 77.0 (15.0) & 80.9 (15.8) & 52.9 (15.3)\\
         & DAPL~(TNNLS25)& \cmark  & \xmark & \cmark  & \textbf{82.0} (76.5) & 83.0 (74.9) & 81.7 (73.2) & 81.2 (74.4) & 80.8 (73.3) & 81.5 (72.5) & 81.7 (74.1)\\
         & VisTA~(Ours)& \cmark  & \cmark & \cmark & 81.0 (75.7) & \textbf{85.5} (76.2) & \textbf{85.3} (76.2) & \textbf{84.5} (76.0) & \textbf{84.0} (76.0) & \textbf{84.0} (76.3) & \textbf{84.0} (76.1)\\
         
         \midrule
          \multirow{9}{*}{C$\rightarrow$R} & ProCA~(ECCV22)& \cmark  & \cmark & \xmark & 76.0 (76.0) & 77.5 (73.0) & 77.3 (70.0) & 73.8 (72.0) & 73.0 (68.0) & 73.3 (66.0) & 75.2 (70.8) \\
         & PLDCA~(TIP24)& \cmark  & \cmark & \xmark & 62.5 (65.0) & 68.5 (64.0) & 71.7 (61.2) & 71.0 (59.1) & 69.0 (55.5) & 71.5 (54.8) & 69.0 (59.9) \\
         & CoOp~(IJCV22)& --  & -- & --  & 88.0 (88.1) & 88.5 (88.1) & 90.0 (88.1) & 87.8 (88.1) & 87.6 (88.1) & 88.5 (88.1) & 88.4 (88.1)\\
         & AttriCLIP~(CVPR23)& \xmark  & \cmark & \cmark  & 68.0 (65.2) & 80.0 (84.6) & 81.0 (81.6) & 79.8 (78.9) & 77.2 (80.8) & 74.7 (70.5) & 76.8 (76.9)\\
         & AD-CLIP~(ICCVW23)& \cmark  & \xmark & \cmark & 88.0 (88.0) & 86.2 (88.0) & 89.2 (90.0) & 88.8 (89.5) & 90.0 (89.0) & 89.5 (88.0) & 88.6 (88.8)\\
         & DAMP~(CVPR24)& \cmark  & \xmark & \cmark & 87.0 (85.3) & 84.5 (88.8) & 85.0 (86.1) & 87.2 (84.7) & 86.4 (84.2) & 85.0 (80.9) & 85.9 (85.0)\\
         & PGA~(NeurIPS24)& \cmark  & \xmark & \cmark & 18.8 (18.8) & 35.7 (18.0) & 54.1 (18.0) & 70.7 (17.6) & 89.3 (18.2) & 90.2 (16.8) & 59.8 (17.9)\\
         & DAPL~(TNNLS25)& \cmark  & \xmark & \cmark  & \textbf{98.0} (93.1) & \textbf{92.0} (92.4) & 92.7 (92.1) & \textbf{91.8} (92.1) & 91.2 (91.6) & 90.3 (91.8) & \textbf{92.7} (92.2)\\
         & VisTA~(Ours)& \cmark  & \cmark & \cmark & 93.0 (92.1) & 92.0 (92.1) & \textbf{93.0} (92.2) & 91.2 (92.0) & \textbf{91.4} (92.0) & \textbf{91.5} (92.2) & 92.0 (92.1)\\
         \midrule
         
          \multirow{9}{*}{C$\rightarrow$S} & ProCA~(ECCV22)& \cmark  & \cmark & \xmark & 61.0 (61.0) & 55.0 (58.0) & 51.0 (54.0) & 50.2 (56.0) & 51.6 (55.0) & 51.0 (52.0) & 53.3 (56.0) \\
         & PLDCA~(TIP24)& \cmark  & \cmark & \xmark & 57.2 (53.6) & 57.7 (52.5) & 57.9 (43.3) & 52.2 (42.9) & 50.3 (38.9) & 51.9 (38.5) & 54.5 (44.9) \\
         & CoOp~(IJCV22)& --  & -- & --  & 78.0 (76.2) & 77.0 (76.2) & 79.7 (76.2) & 79.2 (76.2) & 77.4 (76.2) & 77.8 (76.2) & 78.2 (76.2)\\
         & AttriCLIP~(CVPR23)& \xmark  & \cmark & \cmark  & 72.0 (74.0) & 72.5 (76.1) & 75.0 (76.8) & 74.8 (77.0) & 76.4 (77.4) & 76.7 (75.9) & 74.6 (76.2)\\
         & AD-CLIP~(ICCVW23)& \cmark  & \xmark & \cmark & 79.0 (78.9) & 75.0 (77.9) & 82.0 (80.2) & 76.5 (75.8) & 76.0 (72.9) & 75.7 (75.4) & 77.4 (76.9)\\
         & DAMP~(CVPR24)& \cmark  & \xmark & \cmark & 78.0 (81.5) & 79.5 (78.5) & 80.7 (76.9) & 78.5 (77.2) & 76.6 (76.1) & 76.5 (80.1) & 78.3 (78.4)\\
         & PGA~(NeurIPS24)& \cmark  & \xmark & \cmark & 15.6 (15.6) & 29.7 (14.5) & 48.0 (14.6) & 59.6 (13.7) & 74.8 (14.5) & 80.3 (14.6) & 51.3 (14.6)\\
         & DAPL~(TNNLS25)& \cmark  & \xmark & \cmark  & 77.0 (81.6) & \textbf{79.5} (80.9) & 82.3 (80.6) & 80.8 (78.9) & 79.0 (78.7) & 78.3 (78.6) & 79.5 (79.9)\\
         & VisTA~(Ours)& \cmark  & \cmark & \cmark & \textbf{80.0} (81.4) & 79.0 (81.2) & \textbf{83.0} (81.1) & \textbf{83.0} (81.0) & \textbf{82.2} (80.4) & \textbf{81.5} (80.3) & \textbf{81.5} (80.9)\\
         
         \bottomrule
         \end{tabular}
         }
    \end{center}
\end{table*}

\begin{table*}[t]
\renewcommand\arraystretch{1.0}
    \begin{center}
    \caption{\label{tab:MD_Pn} Classification accuracies (\%)  on Mini-DomianNet with Painting as source domain. Note that the results outside the brackets are Step-level Accuracy, while the results in brackets represent the S-1 Accuracy in each time step. DA, CI, and RF respectively represent domain adaptation, class-incremental, and rehearsal-free.}
    \scalebox{0.75}{
         \begin{tabular}{llp{3mm}p{3mm}p{3mm}|cccccccc}
         \toprule
          Task & Method & DA & CI & RF & Step 1 & Step 2 & Step 3 & Step 4 & Step 5 & Step 6 & Avg.\\
         \midrule
         \multirow{9}{*}{P$\rightarrow$C} & ProCA~(ECCV22)& \cmark  & \cmark & \xmark & 63.0 (63.0) & 55.5 (56.0) & 62.0 (58.0) & 63.2 (59.0) & 57.0 (58.0) & 56.2 (56.0) & 59.5 (58.3) \\
         & PLDCA~(TIP24)& \cmark  & \cmark & \xmark & 65.0 (65.3) & 63.6 (59.6) & 66.3 (56.6) & 65.4 (56.7) & 66.0 (56.3) & 61.5 (55.5) & 64.6 (58.3) \\
         & CoOp~(IJCV22)& --  & -- & --  & 81.0 (80.9) & 83.5 (80.9) & 87.0 (80.9) & 87.0 (80.9) & 84.6 (80.9) & 82.5 (80.9) & 84.3 (80.9)\\
         & AttriCLIP~(CVPR23)& \xmark  & \cmark & \cmark  & 72.0 (71.8) & 79.0 (79.4) & 84.3 (80.4) & 87.2 (82.1) & 85.4 (80.1) & 81.8 (80.5) & 81.6 (79.0)\\
         & AD-CLIP~(ICCVW23)& \cmark  & \xmark & \cmark & \textbf{83.4} (81.0) & 84.0 (80.5) & 86.7 (81.6) & 90.0 (80.3) & 84.8 (80.3) & 80.6 (85.7) & 84.9 (81.6)\\
         & DAMP~(CVPR24)& \cmark  & \xmark & \cmark & 83.0 (82.5) & \textbf{89.0} (82.4) & \textbf{89.7} (82.4) & \textbf{90.0} (81.2) & 85.4 (82.2) & 82.7 (81.9) & 86.6 (82.1)\\
         & PGA~(NeurIPS24)& \cmark  & \xmark & \cmark & 16.6 (16.6) & 34.2 (16.0) & 50.2 (15.2) & 43.4 (6.4)  & 82.6 (14.5) & 80.7 (14.1) & 51.3 (13.8)\\
         & DAPL~(TNNLS25)& \cmark  & \xmark & \cmark  & 77.0 (83.4) & 86.5 (83.2) & 87.3 (82.8) & 88.2 (82.4) & 86.4 (82.0) & 85.5 (81.7) & 85.1 (82.6)\\
         & VisTA~(Ours)& \cmark  & \cmark & \cmark & 82.0 (83.1) & 85.5 (84.0) & 88.7 (83.9) & 89.5 (83.9) & \textbf{88.2} (84.1) & \textbf{85.7} (84.1) & \textbf{86.6} (83.9)\\
         
         \midrule
          \multirow{9}{*}{P$\rightarrow$R} & ProCA~(ECCV22)& \cmark  & \cmark & \xmark & 88.0 (88.0) & 85.5 (84.0) & 88.0 (84.0) & 85.2 (84.0) & 79.8 (75.0) & 83.0 (77.0) & 84.9 (82.0) \\
         & PLDCA~(TIP24)& \cmark  & \cmark & \xmark & 73.9 (76.2) & 75.7 (75.5) & 78.3 (74.2) & 78.6 (70.7) & 78.2 (69.8) & 78.6 (67.8) & 77.2 (72.4) \\
         & CoOp~(IJCV22)& --  & -- & --  & 90.0 (89.9) & 87.0 (89.9) & 89.3 (89.9) & 88.5 (89.9) & 88.6 (89.9) & 88.8 (89.9) & 88.7 (89.9)\\
         & AttriCLIP~(CVPR23)& \xmark  & \cmark & \cmark  & 75.0 (70.9) & 82.0 (87.2) & 86.0 (87.9) & 84.2 (85.4) & 83.6 (84.1) & 84.0 (84.1) & 82.5 (83.3)\\
         & AD-CLIP~(ICCVW23)& \cmark  & \xmark & \cmark & 92.1 (94.0) & 91.1 (89.0) & 89.6 (91.7) & 90.8 (91.4) & 89.2 (90.3) & 90.8 (91.2) & 90.6 (91.3)\\
         & DAMP~(CVPR24)& \cmark  & \xmark & \cmark & 88.0 (85.2) & 91.0 (90.4) & 89.3 (89.0) & 87.0 (87.5) & 87.0 (87.7) & 87.3 (86.1) & 88.3 (87.7)\\
         & PGA~(NeurIPS24)& \cmark  & \xmark & \cmark & 18.6 (18.6) & 35.9 (18.2) & 54.3 (17.6) & 70.9 (18.2) & 87.5 (18.0) & 89.3 (17.6) & 59.4 (18.0)\\
         & DAPL~(TNNLS25)& \cmark  & \xmark & \cmark  & \textbf{96.0} (93.1) & 92.5 (92.5) & 92.7 (91.9) & \textbf{92.0} (92.0) & \textbf{91.4} (91.9) & 91.0 (91.2) & 92.6 (92.1)\\
         & VisTA~(Ours)& \cmark  & \cmark & \cmark & 95.0 (92.6) & \textbf{93.0} (92.9) & \textbf{93.3} (93.1) & 91.8 (93.1) & 91.2 (93.3) & \textbf{91.7} (93.3) & \textbf{92.7} (93.1)\\
         \midrule
         
          \multirow{9}{*}{P$\rightarrow$S} & ProCA~(ECCV22)& \cmark  & \cmark & \xmark & 62.0 (62.0) & 31.0 (45.0) & 25.3 (45.0) & 40.2 (52.0) & 37.6 (50.0) & 37.0 (42.0) & 38.9 (49.3) \\
         & PLDCA~(TIP24)& \cmark  & \cmark & \xmark & 64.4 (56.8) & 60.0 (52.1) & 60.0 (45.3) & 55.4 (43.5) & 53.1 (44.1) & 52.5 (43.1) & 57.6 (47.5) \\
         & CoOp~(IJCV22)& --  & -- & --  & 79.0 (75.0) & 76.0 (75.0) & 75.7 (75.0) & 74.2 (75.0) & 74.6 (75.0) & 73.3 (75.0) & 75.5 (75.0)\\
         & AttriCLIP~(CVPR23)& \xmark  & \cmark & \cmark  & 77.0 (78.1) & 75.5 (79.0) & 81.0 (79.8) & 79.5 (80.4) & 80.0 (80.4) & 78.2 (78.7) & 78.5 (79.4) \\
         & AD-CLIP~(ICCVW23)& \cmark  & \xmark & \cmark & 77.0 (79.4) & 75.5 (79.1) & 79.3 (78.7) & 77.4 (78.8) & 80.8 (80.1) & 77.5 (77.8) & 77.9 (79.0)\\
         & DAMP~(CVPR24)& \cmark  & \xmark & \cmark & \textbf{83.0} (83.2) & \textbf{80.5} (80.0) & 81.7 (80.6) & 80.8 (79.2) & 80.8 (81.2) & 79.2 (79.8) & 81.0 (80.7)\\
         & PGA~(NeurIPS24)& \cmark  & \xmark & \cmark & 15.6 (15.6) & 23.6 (9.2)  & 45.5 (15.0) & 59.0 (11.9) & 75.4 (14.6) & 76.2 (14.8) & 49.2 (13.5)\\
         & DAPL~(TNNLS25)& \cmark  & \xmark & \cmark  & 75.0 (81.4) & 80.0 (80.7) & 81.0 (79.4) & 80.0 (78.1) & 78.6 (79.8) & 78.0 (79.7) & 78.8 (79.9)\\
         & VisTA~(Ours)& \cmark  & \cmark & \cmark & 80.0 (82.3) & 80.0 (82.2) & \textbf{83.3} (81.9) & \textbf{83.2} (82.0) & \textbf{82.2} (81.8) & \textbf{81.7} (81.8) & \textbf{81.7} (82.0)\\
         
         \bottomrule
         \end{tabular}
         }
    \end{center}
\end{table*}

\begin{table*}[t]
\renewcommand\arraystretch{1.0}
    \begin{center}
    \caption{\label{tab:MD_Rl} Classification accuracies (\%)  on Mini-DomianNet with Real World as source domain. Note that the results outside the brackets are Step-level Accuracy, while the results in brackets represent the S-1 Accuracy in each time step. DA, CI, and RF respectively represent domain adaptation, class-incremental, and rehearsal-free.}
    \scalebox{0.75}{
         \begin{tabular}{llp{3mm}p{3mm}p{3mm}|cccccccc}
         \toprule
          Task & Method & DA & CI & RF & Step 1 & Step 2 & Step 3 & Step 4 & Step 5 & Step 6 & Avg.\\
         \midrule
         \multirow{9}{*}{R$\rightarrow$C} & ProCA~(ECCV22)& \cmark  & \cmark & \xmark & 57.0 (57.0) & 59.5 (58.0) & 59.7 (53.0) & 58.0 (56.0) & 58.6 (54.0) & 56.8 (47.0) & 58.3 (54.2) \\
         & PLDCA~(TIP24)& \cmark  & \cmark & \xmark & 59.6 (57.7) & 67.2 (54.4) & 66.8 (52.9) & 64.3 (49.9) & 63.7 (50.8) & 60.5 (48.8) & 63.7 (52.4) \\
         & CoOp~(IJCV22)& --  & -- & --  & 75.0 (81.1) & 84.5 (81.1) & 87.3 (81.1) & 88.0 (81.1) & 85.4 (81.1) & 83.0 (81.1) & 83.9 (81.1)\\
         & AttriCLIP~(CVPR23)& \xmark  & \cmark & \cmark  & 79.0 (73.7) & 82.0 (80.2) & 86.7 (80.9) & 88.2 (80.8) & 84.4 (80.7) & 83.0 (79.8) & 83.9 (79.4)\\
         & AD-CLIP~(ICCVW23)& \cmark  & \xmark & \cmark & 80.0 (82.5) & \textbf{88.0} (82.7) & 87.0 (81.1) & 88.0 (82.3) & 86.8 (81.3) & 80.7 (83.8) & 85.1 (82.3)\\
         & DAMP~(CVPR24)& \cmark  & \xmark & \cmark & 81.0 (82.4) & 83.5 (79.3) & \textbf{89.3} (82.2) & 87.5 (78.7) & 84.0 (74.7) & 79.7 (76.8) & 84.2 (79.2)\\
         & PGA~(NeurIPS24)& \cmark  & \xmark & \cmark & 15.0 (15.0) & 24.2 (8.6)  & 44.9 (16.0) & 64.3 (13.3) & 78.7 (13.1) & 65.8 (14.1) & 48.8 (13.3)\\
         & DAPL~(TNNLS25)& \cmark  & \xmark & \cmark  & 81.0 (82.9) & 87.0 (83.4) & 88.7 (82.3) & 88.8 (82.0) & 85.6 (81.1) & 85.0 (80.8) & 86.0 (82.1)\\
         & VisTA~(Ours)& \cmark  & \cmark & \cmark & \textbf{81.0} (83.3) & 86.5 (83.6) & 88.3 (83.3) & \textbf{89.0} (83.6) & \textbf{87.8} (83.4) & \textbf{85.0} (83.3) & \textbf{86.3} (83.4)\\
         
         \midrule
          \multirow{9}{*}{R$\rightarrow$P} & ProCA~(ECCV22)& \cmark  & \cmark & \xmark & 51.0 (51.0) & 51.5 (48.0) & 59.3 (48.0) & 55.8 (46.0) & 50.2 (42.0) & 55.2 (47.0) & 53.8 (47.0) \\
         & PLDCA~(TIP24)& \cmark  & \cmark & \xmark & 57.9 (57.0) & 63.5 (56.2) & 67.7 (53.0) & 63.7 (51.0) & 63.7 (49.9) & 66.2 (51.3) & 63.8 (53.1) \\
         & CoOp~(IJCV22)& --  & -- & --  & 72.0 (71.1) & 79.0 (71.1) & 79.7 (71.1) & 78.8 (71.1) & 78.2 (71.1) & 77.8 (71.1) & 77.6 (71.1)\\
         & AttriCLIP~(CVPR23)& \xmark  & \cmark & \cmark  & 74.0 (64.0) & 75.5 (66.2) & 72.3 (68.2) & 73.8 (72.0) & 77.8 (70.9) & 70.5 (65.2) & 74.0 (67.7)\\
         & AD-CLIP~(ICCVW23)& \cmark  & \xmark & \cmark & 72.0 (69.9) & 82.0 (72.0) & 80.7 (68.2) & 80.2 (70.4) & 77.2 (71.2) & 78.7 (69.5) & 78.5 (70.2)\\
         & DAMP~(CVPR24)& \cmark  & \xmark & \cmark & 77.0 (69.0) & 81.5 (72.7) & 81.0 (71.2) & 78.0 (71.7) & 78.8 (71.1) & 74.0 (71.9) & 78.4 (71.3)\\
         & PGA~(NeurIPS24)& \cmark  & \xmark & \cmark & 14.1 (14.1) & 27.3 (10.5) & 46.1 (13.7) & 61.7 (14.8) & 64.6 (15.0) & 75.6 (14.5) & 48.2 (13.8)\\
         & DAPL~(TNNLS25)& \cmark  & \xmark & \cmark  & \textbf{81.0} (76.5) & 83.0 (74.7) & 82.7 (73.8) & 80.8 (74.1) & 81.0 (73.3) & 81.0 (73.1) & 81.6 (74.2)\\
         & VisTA~(Ours)& \cmark  & \cmark & \cmark & 79.0 (75.5) & \textbf{83.0} (75.9) & \textbf{83.3} (76.4) & \textbf{82.2} (76.4) & \textbf{83.4} (76.2) & \textbf{83.0} (76.3) & \textbf{82.3} (76.1)\\
         \midrule
         
          \multirow{9}{*}{R$\rightarrow$S} & ProCA~(ECCV22)& \cmark  & \cmark & \xmark & 37.0 (37.0) & 38.5 (34.0) & 41.3 (37.0) & 44.0 (40.0) & 41.6 (36.0) & 39.7 (35.0) & 40.4 (36.5) \\
         & PLDCA~(TIP24)& \cmark  & \cmark & \xmark & 47.8 (39.6) & 46.5 (34.8) & 50.0 (31.5) & 48.4 (31.0) & 42.3 (28.5) & 42.9 (27.6) & 46.3 (32.2) \\
         & CoOp~(IJCV22)& --  & -- & --  & 76.0 (76.7) & 78.0 (76.7) & 81.0 (76.7) & 82.0 (76.7) & 81.4 (76.7) & 79.2 (76.7) & 79.6 (76.7)\\
         & AttriCLIP~(CVPR23)& \xmark  & \cmark & \cmark  & 71.0 (73.4) & 76.5 (77.2) & 78.7 (78.0) & 77.0 (77.8) & 78.6 (76.2) & 73.3 (75.6) & 75.8 (76.4)\\
         & AD-CLIP~(ICCVW23)& \cmark  & \xmark & \cmark & \textbf{81.0} (79.1) & 80.5 (80.6) & 81.7 (78.2) & 80.0 (77.5) & \textbf{82.4} (79.2) & 79.8 (80.3) & 80.9 (79.2)\\
         & DAMP~(CVPR24)& \cmark  & \xmark & \cmark & 78.0 (83.1) & 81.0 (79.6) & 80.3 (77.9) & 82.0 (76.0) & 77.0 (74.7) & 78.5 (78.5) & 79.5 (78.3)\\
         & PGA~(NeurIPS24)& \cmark  & \xmark & \cmark & 14.8 (14.8) & 26.2 (11.5) & 40.0 (10.9) & 58.0 (14.6) & 66.2 (8.2)  & 76.0 (13.7) & 46.9 (12.3)\\
         & DAPL~(TNNLS25)& \cmark  & \xmark & \cmark  & 77.0 (81.4) & 78.0 (80.0) & 82.0 (79.6) & 80.0 (78.9) & 79.2 (79.7) & 77.3 (78.9) & 78.9 (79.8)\\
         & VisTA~(Ours)& \cmark  & \cmark & \cmark & 79.0 (81.7) & \textbf{81.0} (81.5) & \textbf{83.7} (81.9) & \textbf{82.5} (81.5) & 82.2 (81.4) & \textbf{81.2} (80.8) & \textbf{81.6} (81.5)\\
         
         \bottomrule
         \end{tabular}
         }
    \end{center}
\end{table*}

\begin{table*}[t]
\renewcommand\arraystretch{1.0}
    \begin{center}
    \caption{\label{tab:MD_Sk} Classification accuracies (\%)  on Mini-DomianNet with Sketch as source domain. Note that the results outside the brackets are Step-level Accuracy, while the results in brackets represent the S-1 Accuracy in each time step. DA, CI, and RF respectively represent domain adaptation, class-incremental, and rehearsal-free.}
    \scalebox{0.75}{
         \begin{tabular}{llp{3mm}p{3mm}p{3mm}|cccccccc}
         \toprule
          Task & Method & DA & CI & RF & Step 1 & Step 2 & Step 3 & Step 4 & Step 5 & Step 6 & Avg.\\
         \midrule
         \multirow{9}{*}{S$\rightarrow$C} & ProCA~(ECCV22)& \cmark  & \cmark & \xmark & 63.0 (63.0) & 57.0 (57.0) & 61.7 (55.0) & 59.8 (53.0) & 56.2 (51.0) & 57.7 (53.0) & 59.2 (55.3) \\
         & PLDCA~(TIP24)& \cmark  & \cmark & \xmark & 65.5 (68.9) & 53.1 (62.5) & 63.5 (61.4) & 60.5 (59.1) & 57.0 (54.1) & 62.1 (57.1) & 60.3 (60.5) \\
         & CoOp~(IJCV22)& --  & -- & --  & 85.0 (82.8) & 80.5 (82.8) & 84.3 (82.8) & 85.5 (82.8) & 84.0 (82.8) & 82.8 (82.8) & 83.7 (82.8)\\
         & AttriCLIP~(CVPR23)& \xmark  & \cmark & \cmark  & 78.0 (78.9) & 80.5 (82.0) & 87.0 (82.5) & 87.0 (82.6) & 86.6 (83.5) & 83.8 (82.5) & 83.8 (82.0)\\
         & AD-CLIP~(ICCVW23)& \cmark  & \xmark & \cmark & 75.0 (77.1) & 84.5 (82.0) & 87.0 (81.5) & 86.2 (81.2) & 86.8 (84.2) & 85.2 (83.1) & 84.1 (81.5)\\
         & DAMP~(CVPR24)& \cmark  & \xmark & \cmark & \textbf{86.0} (84.0) & 87.0 (83.3) & 88.3 (83.3) & 87.8 (79.8) & 86.2 (79.3) & 84.3 (82.6) & 86.6 (82.1)\\
         & PGA~(NeurIPS24)& \cmark  & \xmark & \cmark & 14.1 (14.1) & 27.9 (11.3) & 52.1 (16.6) & 44.9 (10.5) & 79.5 (10.5) & 84.4 (15.2) & 50.5 (13.0)\\
         & DAPL~(TNNLS25)& \cmark  & \xmark & \cmark  & 81.0 (83.6) & \textbf{88.0} (83.8) & \textbf{89.0} (82.8) & \textbf{89.5} (81.9) & 86.4 (80.6) & 85.7 (81.0) & \textbf{86.6} (82.3)\\
         & VisTA~(Ours)& \cmark  & \cmark & \cmark & 78.0 (83.1) & 85.5 (83.9) & 88.0 (84.3) & 88.8 (84.0) & \textbf{87.8} (83.8) & \textbf{86.0} (83.7) & 85.7 (83.8)\\
         
         \midrule
          \multirow{9}{*}{S$\rightarrow$P} & ProCA~(ECCV22)& \cmark  & \cmark & \xmark & 28.0 (28.0) & 28.5 (26.0) & 43.3 (34.0) & 34.5 (27.0) & 38.0 (33.0) & 44.3 (31.0) & 36.1 (29.8) \\
         & PLDCA~(TIP24)& \cmark  & \cmark & \xmark & 53.3 (50.5) & 51.0 (49.5) & 52.2 (41.7) & 54.4 (42.2) & 56.6 (42.7) & 58.8 (39.3) & 54.4 (44.3) \\
         & CoOp~(IJCV22)& --  & -- & --  & 77.0 (71.8) & 76.0 (71.8) & 77.3 (71.8) & 76.0 (71.8) & 76.0 (71.8) & 76.5 (71.8) & 76.5 (71.8)\\
         & AttriCLIP~(CVPR23)& \xmark  & \cmark & \cmark  & 76.0 (66.2) & 74.0 (68.4) & 75.7 (70.3) & 76.0 (70.1) & 77.2 (71.0) & 78.0 (71.8) & 76.1 (69.6)\\
         & AD-CLIP~(ICCVW23)& \cmark  & \xmark & \cmark & 78.0 (71.0) & 80.0 (70.0) & 79.7 (70.6) & 78.5 (71.9) & 80.2 (73.0) & 79.5 (72.5) & 79.3 (71.5)\\
         & DAMP~(CVPR24)& \cmark  & \xmark & \cmark & 79.0 (72.5) & 79.0 (73.8) & 78.7 (73.7) & 76.2 (70.8) & 78.0 (71.2) & 74.7 (72.0) & 77.6 (72.3)\\
         & PGA~(NeurIPS24)& \cmark  & \xmark & \cmark & 14.5 (14.5) & 32.6 (15.0) & 47.3 (15.8) & 61.7 (15.6) & 71.9 (14.3) & 81.8 (14.8) & 51.6 (15.0)\\
         & DAPL~(TNNLS25)& \cmark  & \xmark & \cmark  & \textbf{82.0} (76.9) & \textbf{84.5} (74.4) & 81.3 (73.4) & 81.0 (74.4) & 80.2 (73.6) & 80.2 (72.5) & 81.5 (74.2)\\
         & VisTA~(Ours)& \cmark  & \cmark & \cmark & 80.0 (75.4) & 83.0 (75.9) & \textbf{83.0} (76.2) & \textbf{82.0} (76.3) & \textbf{81.6} (76.3) & \textbf{82.3} (76.4) & \textbf{82.0} (76.1)\\
         \midrule
         
          \multirow{9}{*}{S$\rightarrow$R} & ProCA~(ECCV22)& \cmark  & \cmark & \xmark & 77.0 (77.0) & 70.5 (73.0) & 72.3 (70.0) & 69.2 (67.0) & 68.2 (65.0) & 70.8 (64.0) & 71.3 (69.3)\\
         & PLDCA~(TIP24)& \cmark  & \cmark & \xmark & 73.4 (75.6) & 67.5 (71.2) & 70.2 (69.2) & 69.2 (64.4) & 69.4 (61.7) & 71.9 (61.5) & 70.3 (67.3)\\
         & CoOp~(IJCV22)& --  & -- & --  & 88.0 (88.1) & 83.0 (88.1) & 87.0 (88.1) & 87.0 (88.1) & 86.0 (88.1) & 86.2 (88.1) & 86.2 (88.1)\\
         & AttriCLIP~(CVPR23)& \xmark  & \cmark & \cmark  & 74.0 (68.3) & 79.0 (84.4) & 78.0 (81.7) & 75.5 (77.2) & 77.0 (77.3) & 78.2 (78.9) & 77.0 (78.0)\\
         & AD-CLIP~(ICCVW23)& \cmark  & \xmark & \cmark & 95.0 (92.2) & 89.0 (89.0) & 90.7 (90.8) & 90.0 (89.8) & 90.0 (90.8) & 89.8 (90.1) & 90.8 (90.5)\\
         & DAMP~(CVPR24)& \cmark  & \xmark & \cmark & 91.0 (90.7) & 85.5 (88.5) & 86.3 (86.3) & 86.0 (83.6) & 88.4 (86.9) & 82.3 (81.4) & 86.6 (86.2)\\
         & PGA~(NeurIPS24)& \cmark  & \xmark & \cmark & 18.2 (18.2) & 35.0 (17.6) & 53.3 (17.4) & 71.3 (18.2) & 87.7 (17.8) & 89.6 (18.2) & 59.2 (17.9)\\
         & DAPL~(TNNLS25)& \cmark  & \xmark & \cmark  & \textbf{97.0} (93.2) & \textbf{93.0} (92.4) & \textbf{94.7} (92.4) & 91.8 (92.2) & \textbf{91.8} (92.0) & 90.8 (91.9) & \textbf{93.2} (92.4)\\
         & VisTA~(Ours)& \cmark  & \cmark & \cmark & 94.0 (92.1) & 91.5 (92.3) & 93.3 (92.7) & \textbf{91.8} (92.7) & 91.6 (92.8) & \textbf{91.2} (92.9) & 92.2 (92.6)\\
         
         \bottomrule
         \end{tabular}
         }
    \end{center}
\end{table*}

The experimental results demonstrate that VisTA achieves superior or comparable performance in terms of Step-level Accuracy and S-1 Accuracy across most steps and adaptation tasks. This indicates that VisTA effectively mitigates catastrophic knowledge forgetting while addressing domain shift on Office-31, Office-Home, and Mini-DomainNet.

\section{More Ablation Analysis}
\label{sec:ablation}

We compare the computational overhead (including GPU Memory Usage and Frames Per Second at batch size 1) in deployment between VisTA and some baseline methods. Table~\ref{overhead} summarizes the results. Compared with CLIP-based methods (e.g., CoOp, AttriCLIP, and AD-CLIP), VisTA maintains comparable computational efficiency while achieving significantly higher accuracy. In contrast to the CI-UDA approach PLDCA, VisTA incurs increased time due to the use of CLIP, but substantially outperforms PLDCA in performance.

\begin{table*}[t]
	\caption{Computational overhead on Office-Home.}
	\centering
	\label{overhead}
	\scalebox{0.75}{
		\begin{tabular}{c|ccc}
			\toprule
			Method & GPU Memory Usage~(MB)~$\downarrow$ & Frames Per Second~$\uparrow$ & Final Accuracy~(\%)\\
			\midrule
			PLDCA~(TIP24) & 1435.9 & \textbf{144.8}  & 76.6 \\
			CoOp~(IJCV22) & \textbf{440.7} & 47.5 & 82.7 \\
			AttriCLIP~(CVPR23) & 705.3 & 48.1 & 82.3 \\
			AD-CLIP~(ICCVW23) & 1016.2 & 11.2 & 83.7\\
			VisTA~(Ours) & 885.6 & 43.9 & \textbf{85.3}\\ 
			\bottomrule
	\end{tabular}}
\end{table*}

To analyze the performance of methods across different Vision-Language Models (weaker or stronger), we evaluate VisTA against some CLIP-based baseline methods on weaker models: CLIP ViT-B/32; default models: CLIP ViT-B/16; and stronger models: OpenCLIP~\cite{Cherti_2023_CVPR} ViT-B/16 (LAION400M) and CLIP ViT-L/14. Table~\ref{VLMs} demonstrates that the performance of all methods depends on the selection of Vision-Language Models. However, VisTA consistently achieves the best performance. For the weaker model (CLIP ViT-B/32), though the performance is inferior to the default or stronger model, VisTA still obtains good accuracy and outperforms all baseline methods remarkably.

\begin{table*}[t]
	\caption{Performance across different Vision-Language Models.}
	\centering
	\label{VLMs}
	\scalebox{0.75}{
		\begin{tabular}{c|cccc}
			\toprule
			Method    & CLIP B/32 & CLIP B/16 & OpenCLIP & CLIP L/14 \\
			\midrule
			vanilla~(ICML21)   & 79.1      & 82.4      & 85.5     & 87.0      \\
			CoOp~(IJCV22)      & 79.3      & 82.7      & 83.8     & 87.6      \\
			AttriCLIP~(CVPR23) & 78.8      & 82.3      & 83.1     & 86.8      \\
			AD-CLIP~(ICCVW23)   & 80.1      & 83.7      & 82.0     & 85.2      \\
			VisTA~(Ours)     & \textbf{82.2}      & \textbf{85.3}      & \textbf{86.8}     & \textbf{89.5}     \\ 
			\bottomrule
	\end{tabular}}
\end{table*}

\section{Extension to Class-Incremental Source-Free Unsupervised Domain Adaptation}
\label{sec:sf}
In some practical scenarios, it is infeasible to access source data when training on target domain, while only the pretrained source model are provided~\cite{Litrico_2023_CVPR,tang2024source}. This scenario derives Class-Incremental Source-Free Unsupervised Domain Adaptation (CI-SFUDA)~\cite{CISFUDA}. To address CI-SFUDA, we formalize the training process into two distinct stages: \textbf{Source Pretraining} and \textbf{Target Deployment}. During Source Pretraining, $\mathcal{A}^s$ is optimized using: 
\begin{equation}\label{s_pt}
	\mathcal{L}_{\mathrm{pt}} = \mathcal{L}_{\mathrm{sup}}^s+\lambda_2 \mathcal{L}_{\mathrm{hp}}+\lambda_3 \mathcal{L}_{\mathrm{div}}^s,
\end{equation}
and $\mathcal{K}^s$ is initialized via K-means clustering before training. 

Building on the frozen $[\mathcal{K}^s, \mathcal{A}^s]$, the Target Deployment stage trains $\mathcal{A}^t$ using the following loss where $\mathcal{L}_{\mathrm{con}}$ leverages only $\mathcal{D}^t$: 
\begin{equation}\label{t_dp}
	\mathcal{L}_{\mathrm{dp}} = \mathcal{L}_{\mathrm{sup}}^t+\lambda_1 \mathcal{L}_{\mathrm{con}}+\lambda_3 \mathcal{L}_{\mathrm{div}}^t.
\end{equation}

At each step, VisTA updates $\mathcal{K}^t$ following the protocol in Section~\ref{Learn}. $\mathcal{L}_{\mathrm{div}}$ is applied separately to $\mathcal{D}_s~(\mathcal{L}_{\mathrm{div}}^s)$ and $\mathcal{D}_t~(\mathcal{L}_{\mathrm{div}}^t)$.

\begin{figure}[t]
  \centering
  \subfloat[S-1 Acc. (\%)]{
      \includegraphics[width=0.225\textwidth]{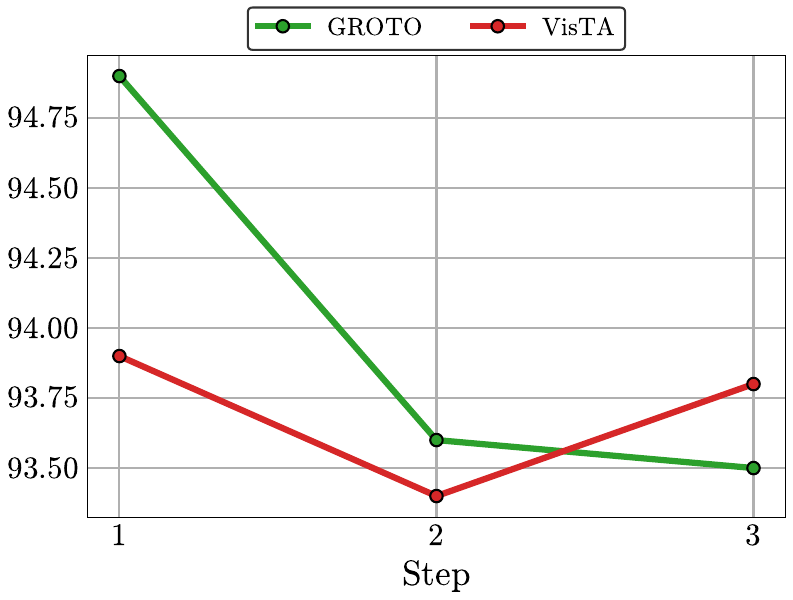}
  }
  \subfloat[$\Delta\%$~(Step 1)]{
      \includegraphics[width=0.2235\textwidth]{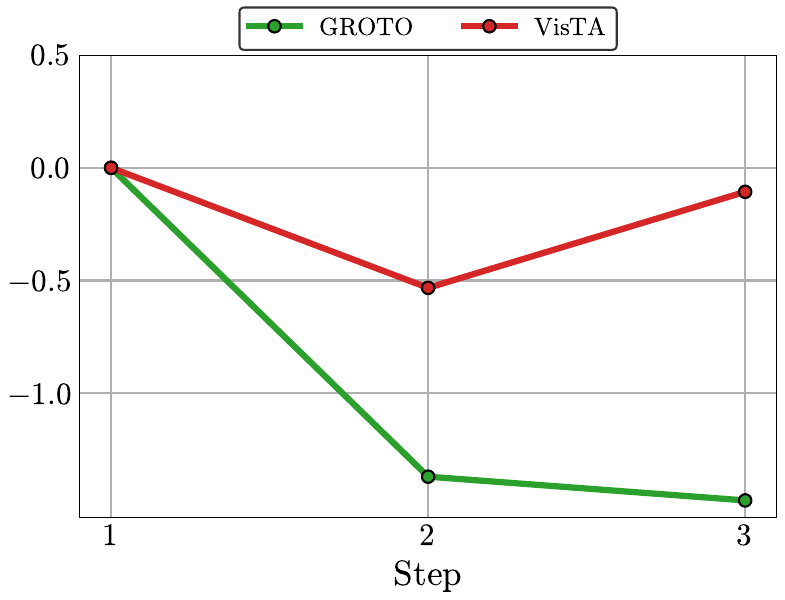}
  }
  \caption{S-1 Accuracy of CI-SFUDA at each step and its percentage change ($\Delta \%$) compared with Step-1 on Office-31.}
  \label{SF31_S-1}
  \Description{S-1 Accuracy of CI-SFUDA at each step and its percentage change ($\Delta \%$) compared with Step-1 on Office-31.}
\end{figure}

\begin{figure}[t]
  \centering
  \subfloat[S-1 Acc. (\%)]{
      \includegraphics[width=0.225\textwidth]{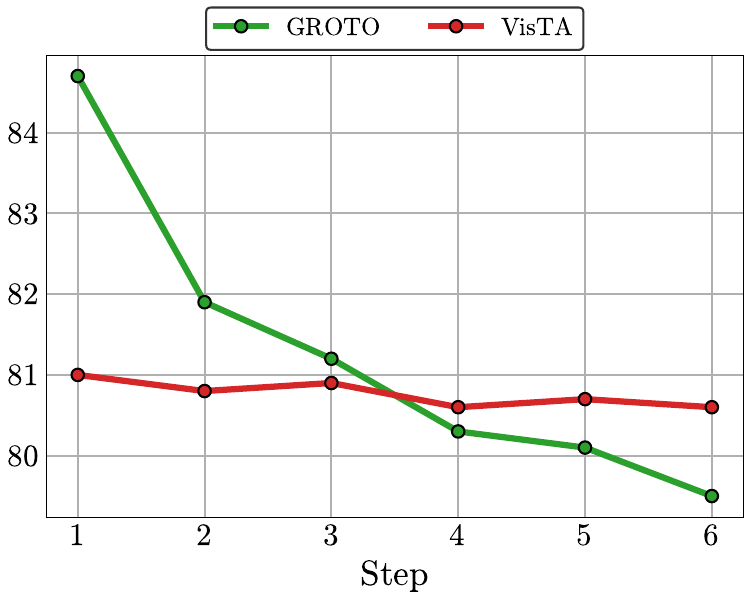}
  }
  \subfloat[$\Delta\%$~(Step 1)]{
      \includegraphics[width=0.228\textwidth]{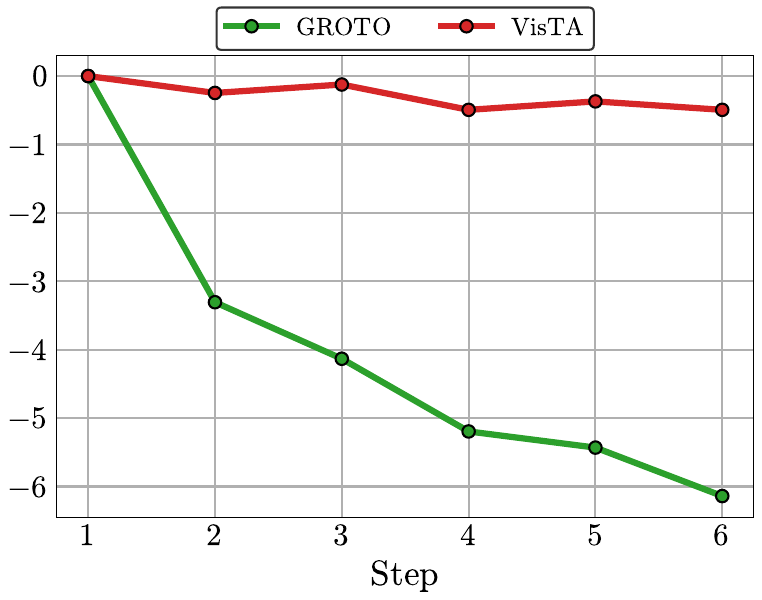}
  }
  \caption{S-1 Accuracy of CI-SFUDA at each step and its percentage change ($\Delta \%$) compared with Step-1 on Office-Home.}
  \label{SFOH_S-1}
  \Description{S-1 Accuracy of CI-SFUDA at each step and its percentage change ($\Delta \%$) compared with Step-1 on Office-Home.}
\end{figure}

In this section, we compare our method with the state-of-the-art CI-SFUDA method GROTO~\cite{CISFUDA}, which outperforms the Source-Free version of ProCA~\cite{lin2022prototype} and some Source-Free (Universal) Domain Adaptation methods~\cite{Litrico_2023_CVPR,tang2024source,Tang_2024_CVPR,Qu_2024_CVPR}. 
We directly cite the reported results of GROTO because: (1) VisTA and GROTO share identical data construction protocols on Office-31, Office-Home, and Mini-DomainNet benchmarks, and (2) GROTO employs the same ViT-B/16 backbone as VisTA.
Following the evaluation protocol of CI-UDA, we measure performance via Final Accuracy (Tables~\ref{SF-31}, \ref{SF-Home}, and \ref{SF-Domain}) across three benchmarks. Additionally, we report Step-level Accuracy (Table~\ref{Step-SFHome}) and S-1 Accuracy (Figures~\ref{SF31_S-1} and \ref{SFOH_S-1}) for Office-31 and Office-Home but not for Mini-DomainNet, since GROTO exclusively reports averaged Final Accuracy across three subtasks per target domain on Mini-DomainNet.

The results demonstrate that the source-free version of VisTA achieves improvements of 2.2\% and 19.1\% over GROTO on Office-Home and Mini-DomainNet, respectively. Notably, it surpasses GROTO in the last S-1 Accuracy on Office-31 and Office-Home, with significantly smaller degradation in S-1 Accuracy than that of GROTO. These findings validate that VisTA—despite not being explicitly designed for CI-SFUDA—either outperforms or achieves performance comparable to GROTO across most metrics. Furthermore, these results reinforce our argument in Section~\ref{soat} that the performance advantages of VisTA scale with benchmark complexity. The findings also confirm the capability of VisTA to mitigate domain shift and catastrophic forgetting when extended to source-free scenarios.

\begin{table*}[t]
	\centering
	\caption{Final Accuracy (\%) of CI-SFUDA on Office-31. DA, CI, Sf, and RF respectively represent domain adaptation, class-incremental, source-free, and rehearsal-free.}
	\scalebox{0.75}{  
		\begin{tabular}{lcccc|cccccccc}
			\toprule
			Method & DA & CI & SF & RF & A$\rightarrow$D & A$\rightarrow$W & D$\rightarrow$A & D$\rightarrow$W & W$\rightarrow$A & W$\rightarrow$D & Avg.\\
			\midrule
			GROTO~(CVPR25)& \cmark  & \cmark & \cmark & \xmark & \textbf{99.4} & \textbf{99.0} & 81.3 & \textbf{99.0} & 81.3 & \textbf{98.1} & \textbf{93.0}  \\
			VisTA~(Ours)& \cmark  & \cmark & \cmark & \cmark & 87.4 & 88.8 & \textbf{85.2} & 93.5 & \textbf{82.5} & 95.7 & 88.9 \\
			\bottomrule
		\end{tabular}
	}
	\label{SF-31}
\end{table*}

\begin{table*}[t]
	\setlength\tabcolsep{2.2pt}
	\centering
	\caption{Final Accuracy (\%) of CI-SFUDA on Office-Home. DA, CI, Sf, and RF respectively represent domain adaptation, class-incremental, source-free, and rehearsal-free.}
	\scalebox{0.75}{  
		\begin{tabular}{lcccc|ccccccccccccc}
			\toprule
			Method & DA & CI & SF & RF &A$\rightarrow$C & A$\rightarrow$P & A$\rightarrow$R & C$\rightarrow$A & C$\rightarrow$P & C$\rightarrow$R & P$\rightarrow$A & P$\rightarrow$C & P$\rightarrow$R & R$\rightarrow$A & R$\rightarrow$C & R$\rightarrow$P & Avg.\\
			\midrule
			GROTO~(CVPR25)& \cmark  & \cmark & \cmark & \xmark & 65.7 & 86.4 & 89.7 & \textbf{85.8} & 86.3 & \textbf{90.9} & \textbf{86.0} & 67.1 & 90.1 & \textbf{86.9} & 66.2 & 89.3 & 82.5 \\

			VisTA~(Ours)& \cmark  & \cmark & \cmark & \cmark & \textbf{71.4} & \textbf{92.3} & \textbf{90.8} & 84.5 & \textbf{92.2} & 90.5 & 83.7 & \textbf{72.5} & \textbf{91.1} & 83.5 & \textbf{71.8} & \textbf{92.6} & \textbf{84.7}\\
			\bottomrule
		\end{tabular}
	}
	\label{SF-Home}
\end{table*}

\begin{table*}[t]
	\setlength\tabcolsep{2.2pt}
	\centering
	\caption{Final Accuracy (\%) of CI-SFUDA on Mini-DomainNet. DA, CI, Sf, and RF respectively represent domain adaptation, class-incremental, source-free, and rehearsal-free.}
	\scalebox{0.75}{  
		\begin{tabular}{lcccc|ccc|ccc|ccc|ccc|c}
			\toprule
			Method & DA & CI & SF & RF & P$\rightarrow$C    & R$\rightarrow$C  & S$\rightarrow$C   &C$\rightarrow$P   & R$\rightarrow$P & S$\rightarrow$P & C$\rightarrow$R  &  P$\rightarrow$R   & S$\rightarrow$R  & C$\rightarrow$S   & P$\rightarrow$S                          & R$\rightarrow$S & Avg.\\
			\midrule
			GROTO~(CVPR25)& \cmark  & \cmark & \cmark & \xmark & \multicolumn{3}{c|}{66.2}& \multicolumn{3}{c|}{61.7} & \multicolumn{3}{c|}{78.0} & \multicolumn{3}{c|}{56.7} & 65.7\\
			VisTA~(Ours)& \cmark  & \cmark & \cmark & \cmark & \textbf{85.7} &\textbf{85.5} & \textbf{86.2} & \textbf{82.7} & \textbf{81.0} & \textbf{81.3} & \textbf{91.5} & \textbf{90.8} & \textbf{91.2} & \textbf{81.7} & \textbf{78.8} & \textbf{80.7} & \textbf{84.8}\\
			\bottomrule
		\end{tabular}
	}
	\label{SF-Domain}
\end{table*}

\begin{table*}[t]
	\setlength\tabcolsep{3pt}
	\centering
	\caption{Step-level Accuracy (\%) of CI-SFUDA on Office-31 and Office-Home. DA, CI, Sf, and RF respectively represent domain adaptation, class-incremental, source-free, and rehearsal-free.}
	\scalebox{0.75}{  
		\begin{tabular}{lcccc|cccc|ccccccc}
			\toprule
			\multirow{2}{*}{Method} & \multirow{2}{*}{DA} & \multirow{2}{*}{CI} & \multirow{2}{*}{SF} & \multirow{2}{*}{RF} & \multicolumn{4}{c|}{Office-31} & \multicolumn{7}{c}{Office-Home} \\
			\cmidrule(lr){6-9} \cmidrule(lr){10-16} & & & & & Step 1 & Step 2 & Step 3 & Avg. & Step 1 & Step 2 & Step 3 & Step 4 & Step 5 & Step 6 & Avg.\\
			\midrule
			GROTO~(CVPR25)& \cmark  & \cmark & \cmark & \xmark & \textbf{96.7}          & \textbf{94.9}          & \textbf{93.0}          & \textbf{94.9}          & \textbf{84.4}        & \textbf{84.8}         & 81.7 & 83.3 & 82.8 & 82.5 & 83.2  \\
			VisTA~(Ours)& \cmark  & \cmark & \cmark & \cmark & 93.9   & 92.7   & 88.9 & 91.8 & 81.0 & 84.3 & \textbf{83.6} & \textbf{84.6} & \textbf{84.7} & \textbf{84.7} & \textbf{83.8} \\
			\bottomrule
		\end{tabular}
	}
	\label{Step-SFHome}
\end{table*}

\end{document}